\documentclass[acmsmall]{acmart}
\RequirePackage{ifluatex}
\RequirePackage{ifvtex}
\let\ifluatex\relax
\usepackage{amsmath}
\usepackage{amsthm}

\usepackage{algorithmic}

\usepackage{amssymb}
\usepackage{multirow}
\usepackage{bm}
\usepackage{algorithm}
\usepackage{bbding}
\newlength\savedwidth

\newlength\savewidth

\newcommand{\tabincell}[2]{\begin{tabular}{@{}#1@{}}#2\end{tabular}}  
\AtBeginDocument{%
  \providecommand\BibTeX{{%
    \normalfont B\kern-0.5em{\scshape i\kern-0.25em b}\kern-0.8em\TeX}}}

\acmJournal{JACM}
\begin{document}

\title{Text Recognition in the Wild: A Survey}


% \author{Xiaoxue~Chen$^\dagger$,~
%         Lianwen~Jin$^\dagger$$^\ast$,~
%         Yuanzhi~Zhu,~
%         Canjie~Luo,~
%         and~Tianwei~Wang% <-this % stops a space
% \thanks{X. Chen, L. Jin, Y. Zhu, C. Luo, and T. Wang are with the College of Electronic and Information Engineering, South China University of Technology, Guangzhou, China.}
% % note need leading \protect in front of \\ to get a newline within \thanks as
% % \\ is fragile and will error, could use \hfil\break instead.
% \thanks{E-mail: \{xxuechen, z.yuanzhi, wangtw\}@foxmail.com, \{$^\ast$lianwen.jin, canjie.luo\}@gmail.com.}
% \thanks{ $^\dagger$These authors contributed equally.}}% <-this % stops an unwanted space
% % \thanks{Manuscript received April 19, 2005; revised August 26, 2015.}}

\author{Xiaoxue~Chen}
\authornote{Both authors contributed equally to this research.}
\email{xxuechen@foxmail.com}
\author{Lianwen~Jin}
\authornotemark[1]
\email{lianwen.jin@gmail.com}
\author{Yuanzhi~Zhu}
\email{z.yuanzhi@foxmail.com}
\author{Canjie~Luo}
\email{canjie.luo@gmail.com}
\author{Tianwei~Wang}
\email{wangtw@foxmail.com}
\affiliation{%
  \institution{the College of Electronic and Information Engineering, South China University of Technology}
  % \streetaddress{381 Wushan Road}
  \city{Guangzhou}
  \country{China}
}

\renewcommand{\shortauthors}{Chen and Jin, et al.}

\begin{abstract}
The history of text can be traced back over thousands of years.
Rich and precise semantic information carried by text is important in a wide range of vision-based application scenarios.
Therefore, text recognition in natural scenes has been an active research field in computer vision and pattern recognition.
In recent years, with the rise and development of deep learning, numerous methods have shown promising in terms of innovation, practicality, and efficiency.
% With the rise and development of deep learning, the community has witnessed substantial advancements in the innovation, practicality and efficiency of algorithms.
This paper aims to (1) summarize the fundamental problems and the state-of-the-art associated with scene text recognition; (2) introduce new insights and ideas; (3) provide a comprehensive review of publicly available resources; (4) point out directions for future work.
In summary, this literature review attempts to present the entire picture of the field of scene text recognition.
It provides a comprehensive reference for people entering this field, and could be helpful to inspire future research.
% It can serve as a reference for researchers and can be helpful in future work.
Related resources are available at our Github repository: \url{https://github.com/HCIILAB/Scene-Text-Recognition}.
\end{abstract}

\begin{CCSXML}
<ccs2012>
 <concept>
  <concept_id>10010520.10010553.10010562</concept_id>
  <concept_desc>Computer systems organization~Embedded systems</concept_desc>
  <concept_significance>500</concept_significance>
 </concept>
 <concept>
  <concept_id>10010520.10010575.10010755</concept_id>
  <concept_desc>Computer systems organization~Redundancy</concept_desc>
  <concept_significance>300</concept_significance>
 </concept>
 <concept>
  <concept_id>10010520.10010553.10010554</concept_id>
  <concept_desc>Computer systems organization~Robotics</concept_desc>
  <concept_significance>100</concept_significance>
 </concept>
 <concept>
  <concept_id>10003033.10003083.10003095</concept_id>
  <concept_desc>Networks~Network reliability</concept_desc>
  <concept_significance>100</concept_significance>
 </concept>
</ccs2012>
\end{CCSXML}

\ccsdesc[500]{Computer systems organization~Embedded systems}
\ccsdesc[300]{Computer systems organization~Redundancy}
\ccsdesc{Computer systems organization~Robotics}
\ccsdesc[100]{Networks~Network reliability}

\keywords{scene text recognition, end-to-end systems, deep learning}

\maketitle

\section{Introduction}
{T}{ext} is a system of symbols used to record, communicate, or inherit culture.
As one of the most influential inventions of humanity, text has played an important role in human life.
Specifically, rich and precise semantic information carried by text is important in a wide range of vision-based application scenarios, such as image search \cite{tsai2011mobile}, intelligent inspection \cite{chen2007cindi}, industrial automation \cite{ham1995recognition}, robot navigation \cite{desouza2002vision}, and instant translation \cite{lu2011foreign}.
Therefore, text recognition in natural scenes has drawn the attention of researchers and practitioners, as indicated by the emergence of recent ``ICDAR Robust Reading Competitions'' \cite{lucas2005icdar}, \cite{lucas2005icdartext}, \cite{shahab2011icdar}, \cite{karatzas2013icdar}, \cite{karatzas2015icdar}, \cite{sanchez2017icdar2017}, \cite{sun2019icdar}. 

\begin{figure}[t]
\centering
\includegraphics[width=0.5\textwidth]{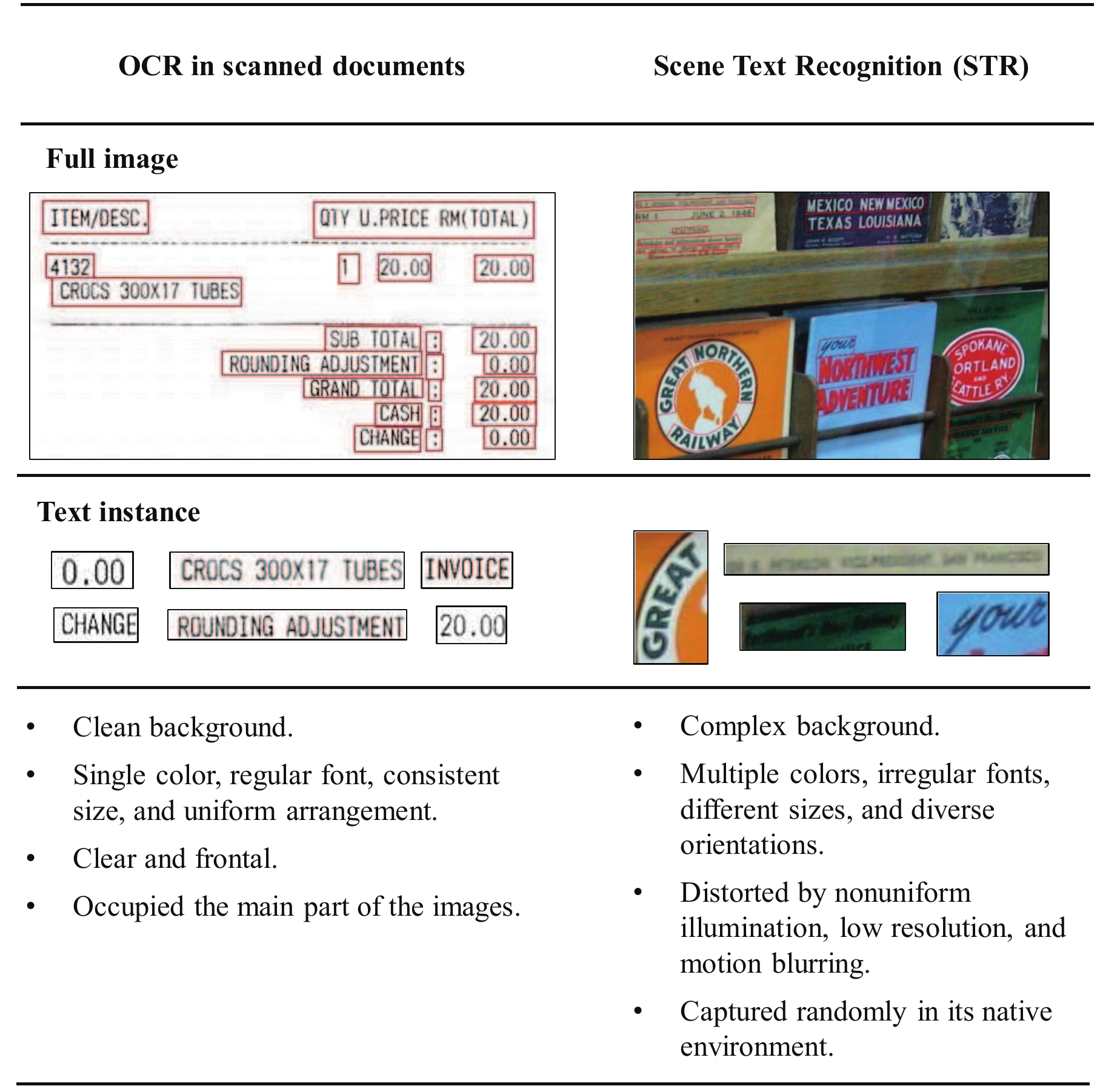}
\caption{Comparison of STR and OCR in scanned documents.}
\label{Figure_OCR_STR}
\Description{Comparison of STR and OCR in scanned documents.}
\end{figure}

Recognizing text in natural scenes, also known as scene text recognition (STR), is usually considered as a special form of optical character recognition (OCR), i.e., camera-based OCR.
Although OCR in scanned documents is well developed \cite{nagy2000twenty}, \cite{zhou2014perspective}, STR remains challenging because of many factors, such as complex backgrounds, various fonts, and imperfect imaging conditions.
Figure~\ref{Figure_OCR_STR} compares the following characteristics of STR and OCR in scanned documents.
% , which are analyzed as follows:
\begin{itemize} 
\item \textbf{Background:}
Unlike OCR in scanned documents, text in natural scenes can appear on anything (e.g., signboards, walls, or product packagings).
Therefore, scene text images may contain very complex backgrounds.
Moreover, the texture of the background can be visually similar to text, which causes additional challenges for recognition.
\item \textbf{Form:}
Text in scanned documents is usually printed in a single color with regular font, consistent size, and uniform arrangement.
In natural scenes, text appears in multiple colors with irregular fonts, different sizes, and diverse orientations.
The diversity of text makes STR more difficult and challenging than OCR in scanned documents.
\item \textbf{Noise:} 
Text in natural scenes is usually distorted by noise interference, such as nonuniform illumination, low resolution, and motion blurring.
% the imperfect imagery conditions, such as nonuniform illumination, low resolution and motion blurring.
Imperfect imaging conditions cause failures in STR.
% Therefore, recognition of text is sensitive to the environmental interference.
\item \textbf{Access:}
Scanned text is usually frontal and occupies the main part of the image.
However, scene text is captured randomly,
% in its native environment
which results in 
%special 
irregular deformations (such as perspective distortion).
%Deformed text only has roughly shape, which limits computers from accurately recognizing text in natural scenes. 
Various shapes of text increase the difficulty of recognizing characters and predicting text strings.
\end{itemize}

% \begin{table}[t]
%   \scriptsize
%   \centering
%   \caption{The comparison of characteristics between STR and OCR based on scanned documents.}
%     \begin{tabular}{c|l}
%     \toprule
%     \toprule
%     \multirow{4}[2]{*}{\textbf{OCR}} & $\bullet$ clean background. \\
%           & $\bullet$ single color, regular font, consistent size and uniform arrangement. \\
%           & $\bullet$ clear and frontal. \\
%           & $\bullet$ occupied the core of the images.  \\
%     \midrule
%     \multirow{4}[2]{*}{\textbf{STR}} & $\bullet$ complex background. \\
%           & $\bullet$ multiple colors, irregular fonts, different sizes and diverse orientations. \\
%           & $\bullet$ distorted by nonuniform illumination, low resolution and motion blurring. \\
%           & $\bullet$ incidentally captured in its native environment. \\
%     \bottomrule
%     \bottomrule
%     \end{tabular}%
%   \label{Table_OCR_STR}%
% \end{table}%
Recognizing text in natural scenes has attracted great interest from academia and industry in recent years because of its importance and challenges.

% Early work {\color{red}{Please give citation}} \cite{} mainly relies on hand-crafted features.
Early research \cite{wang2012end}, \cite{neumann2012real}, \cite{yao2014unified} mainly relied on hand-crafted features.
%which not only requires professional researchers, but also makes recognition performance limited by the low capability of features.
Low capabilities of these features limited the recognition performance. 
%The development of deep learning gives rise to a blossom of works that push the envelope further. 
With the development of deep learning, neural networks significantly boosted the performance of STR.
Several primary factors are driving deep learning-based STR algorithms.
The first factor is the advances in hardware systems. 
%On the one hand, the training of large-scale recognition networks can be possible owing to the emergence of high-performance computing systems, such as GPUs or distributed clusters \cite{dean2012large}.
High-performance computing systems \cite{dean2012large} can train large-scale recognition networks.
%On the other hand, high performance mobile devices \cite{chen2004automatic}, \cite{liu2008camera} create an opportunity for carrying more complex algorithms and recognizing text in various environments at anytime.
Moreover, modern mobile devices \cite{chen2004automatic}, \cite{liu2008camera} are capable of running complex algorithms in real-time.
% meet the requirement of the application of complex algorithms in real time.
The second is 
%excellent characteristics 
automatic feature learning in deep learning-based STR algorithms, which not only frees researchers from the exhausting work of designing and 
%testing 
selecting hand-crafted features, but also significantly improves recognition performance.
% The third factor is the growing demand for applications {\color{red}{Citation required}}.
The third is the growing demand for STR applications \cite{corbelli2016historical}, \cite{Sermanet2012convolutional}, \cite{zhong2000automatic}, \cite{ye2005fast}, \cite{yang2019fully}.
% Text, as one of the most important means of recording and communication, is everywhere in life.
Text in natural scenes can provide rich and precise information, which is beneficial for understanding the scene.
Automatic recognition of text in natural scenes is economically viable in the era of big data, which attracts researchers and practitioners.  
% The second is the development of hardware systems.
% The training of large-scale recognition network can be possible owing to the emergence of high-performance computing systems, such as GPUs or distributed clusters \cite{dean2012large}.
% High performance mobile devices \cite{chen2004automatic}, \cite{liu2008camera} creates an opportunity for carrying more complex algorithms and recognizing text in various environments at anytime.
% The third factor is that numerous potential approaches have been proposed, making it more feasible to address challenging problems, such as the perspective or curved text.

This paper attempts to comprehensively review the field of STR and establish a baseline for a fair comparison of algorithms. 
We present the entire picture of STR by summarizing fundamental problems and the state-of-the-art, introducing new insights and ideas, and looking ahead into future trends.
Hence, this paper aims to serve as a reference for researchers and can be helpful in future work.
% enumerating technical challenges, analyzing, and comparing technical challenges, potential algorithms, and performance of related research in recent years, which can serve as a good reference for researchers and can be used in future research.
Moreover, we provide a comprehensive review of publicly available resources, including the standard benchmark datasets and related code.

% There 
%are already 
% Several
% exists several 
%well-written and 
 % review of STR already exist \cite{zhang2013text}, \cite{uchida2014text}, \cite{ye2014text}, \cite{zhu2016scene}, \cite{yin2016text}, \cite{long2018scene}.
There are several STR reviews in the literature \cite{zhang2013text}, \cite{uchida2014text}, \cite{ye2014text}, \cite{zhu2016scene}, \cite{yin2016text}, \cite{long2018scene}.
However, most of the above-mentioned surveys \cite{zhang2013text}, \cite{uchida2014text}, \cite{ye2014text}, \cite{zhu2016scene}, \cite{yin2016text} are outdated.
Many recent advances, such as the algorithms developed in $2018$ – $2020$, are not included in these surveys.
%$3$ 
We refer the readers to these papers for a more comprehensive historical literature review.
Moreover, Zhu et al. \cite{zhu2016scene} and Long et al. \cite{long2018scene} reviewed methods for both scene text detection and recognition.
Yin et al. \cite{yin2016text} surveyed algorithms for text detection, tracking, and recognition in video.
Unlike these surveys, our paper mainly focuses on STR and aims to provide a more detailed and comprehensive overview of this field.

The remainder of this paper is organized as follows.
Section~\ref{background} presents the background, fundamental problems, and special issues associated with text.
Section~\ref{methodologies} introduces new insights and ideas developed for STR in recent years.
Section~\ref{evaluations_and_protocols} summarizes the standard benchmark datasets and evaluation protocols and compares the performance of recognition algorithms.
Finally, Section~\ref{discussion_and_future_directions} concludes the paper and identifies potential directions for future work in STR.

\section{Background}
\label{background}

To comprehensively understand the field of STR, we will describe the fundamental problems and special issues associated with text.
Moreover, some representative applications of STR will be listed and analyzed in this section.

\subsection{Text in Images}
\label{text_in_imagery}

\begin{figure}[t]
\centering
\includegraphics[width=0.6\textwidth]{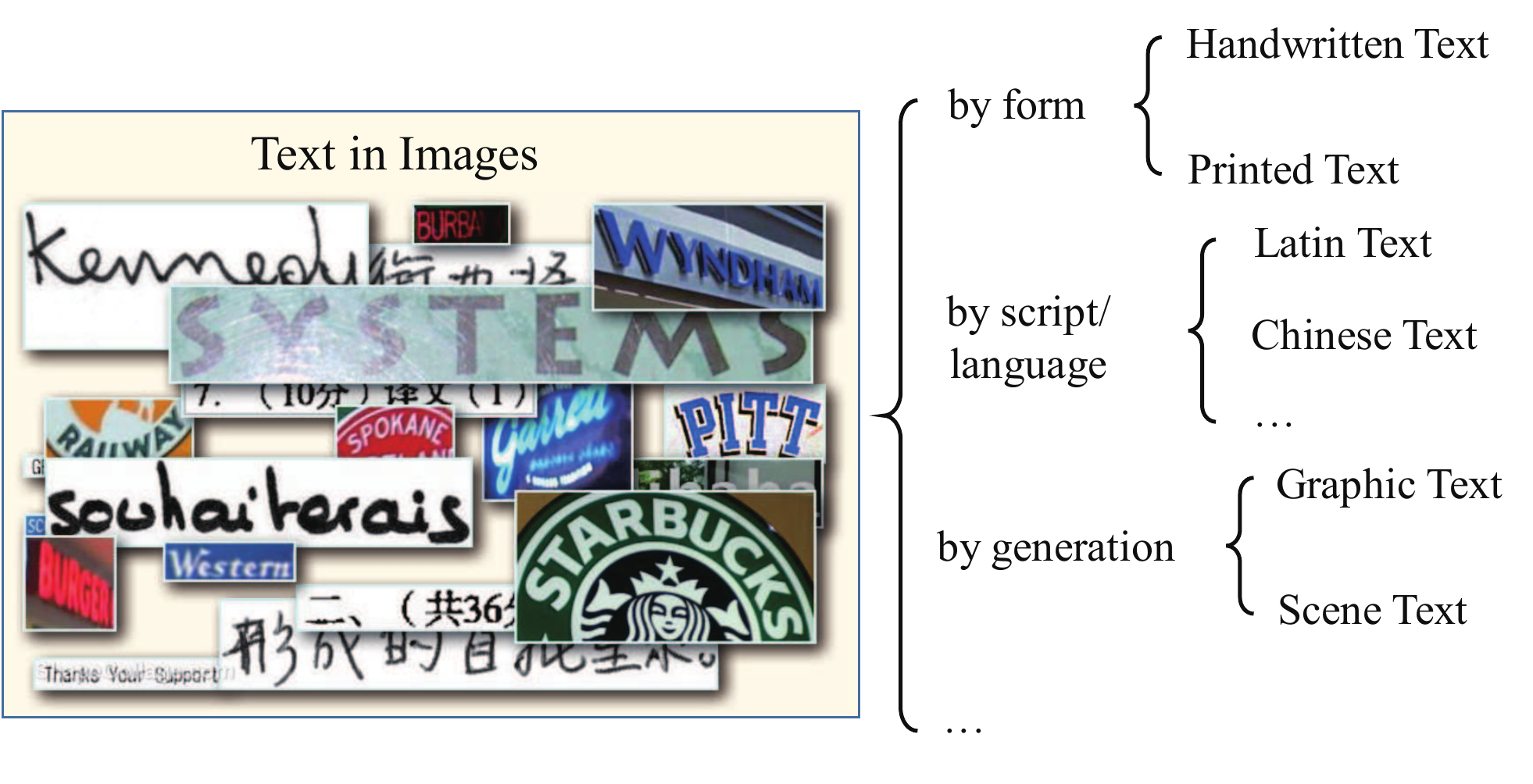}
\caption{Typical classifications of text in images.
}
\label{Figure_text_in_imagery}
\end{figure}

% Understanding its research objects is the first step to having an overview of the field of STR.

Text can appear differently in the images.
Figure~\ref{Figure_text_in_imagery} shows examples and typical classifications.
% It is notable that there are overlaps between different classification methods.
For example, if classified by the text form, handwritten text and printed text are two basic classes.
Notably, classification methods may overlap.
Handwritten text recognition is more challenging than printed text recognition because of various handwriting styles and character-touching problem \cite{xie2017learning}, \cite{zhang2019sequence}.
% Moreover, various handwritten styles and character-touching problem make handwritten text recognition more challenging than printed text recognition \cite{xie2017learning}, \cite{zhang2019sequence}.
Depending on the scripts/languages, text in images may comprise different characters such as Latin, Chinese, or Hindi.
Text characteristics, such as text categories and the reading order, vary greatly in different languages.
% Most recognition algorithms focuses on English text.
Following the definition in \cite{ye2014text}, text in images can also be divided into ``graphic text'' and ``scene text''.
The former refers to text that is digitally added as an overlay on videos or images.
% machine print text overlaid graphically, which can be captured in video or born-digital images on the web.
The latter refers to text on objects, captured in its native environment.
Scene text has diverse styles and can appear on any surface, which makes it difficult to distinguish text from complex backgrounds.
% two categories.
% Figure~\ref{Figure_Compare_Graphic_and_Scene} illustrates sample images of text in imagery.
% The first category is scanned text, which refers to the text in scanned documents or books.
% % Scanned text is usually used in OCR task, which is relatively easy to recognize.
% The first category is graphic text, which refers to machine print text overlaid graphically.
% Graphic text is usually frontal and regular, and can be captured in video or born-digital images on the web. 
% The other category is scene text, which refers to the text on objects in its native environment.
% As discussed in Section~\ref{introduction}, scene text often shows with diverse styles, and can appear on anything, such as signboards or walls, which makes it difficult to distinguish from complex backgrounds.
  
 % owing to its importance and challenge. 
% Most recent research has focused on scene text owing to its importance and challenge.
% The community has seen a surge of research efforts and substantial progresses.
% Typically, STR deals with printed Latin scene text.
% Most approaches summarized in this paper use this type of text.
Typically, most approaches summarized in this paper deals with printed Latin scene text.

\subsection{Fundamental Problems and Special Issues with Text}
\label{text_based_fundamental_problems_and_special_issues}

\begin{figure}[htbp]
\centering
\includegraphics[width=0.4\textwidth]{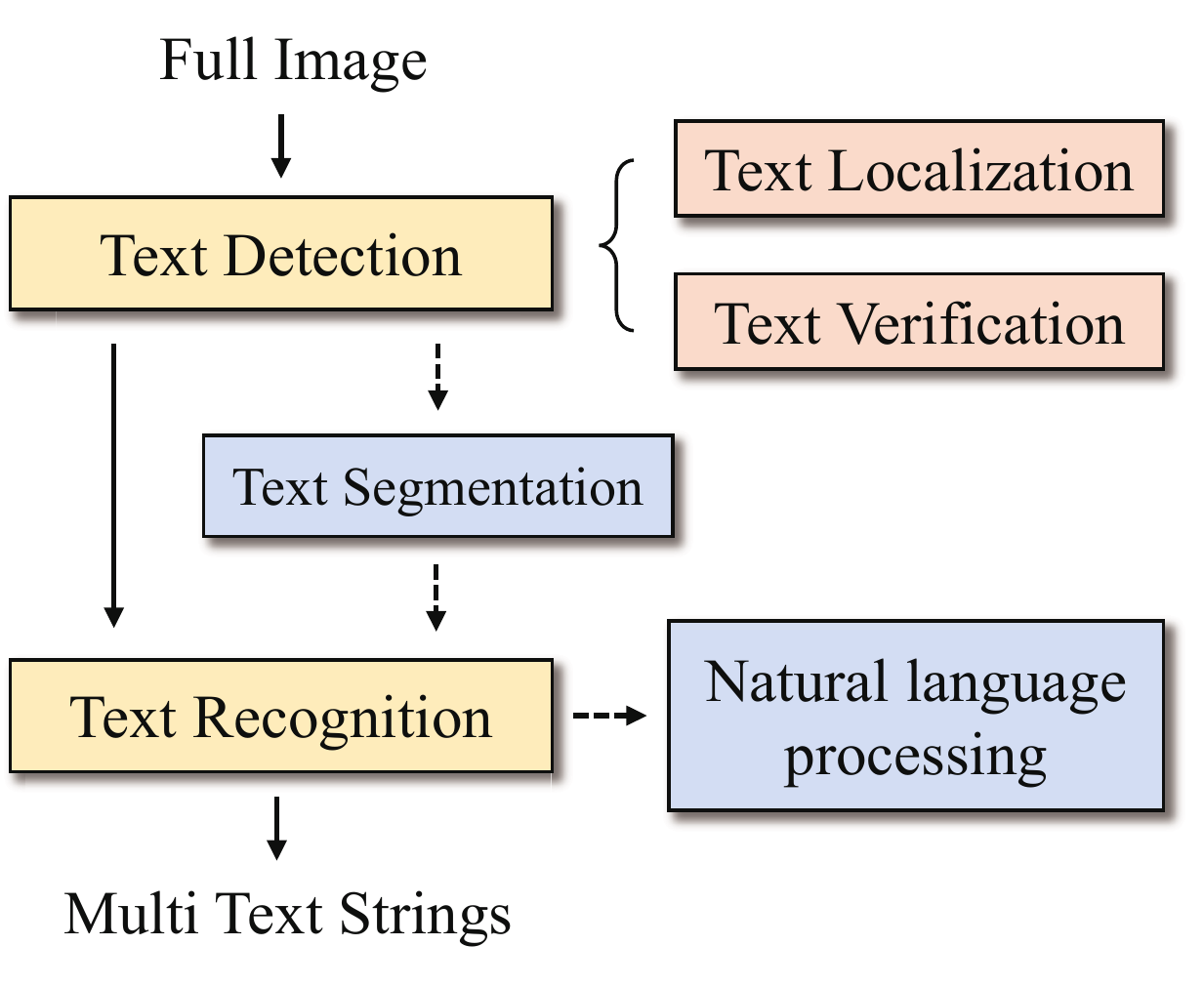}
\caption{Illustration of an end-to-end system, which defines various fundamental problems at various stages: text detection, text localization, text verification, text segmentation, and text recognition.
Some stages are not considered in an end-to-end system.
}
\label{Figure_e2e_system}
\end{figure}

% A clear understanding of common text-based fundamental problems and special issues can be helpful to view the filed of STR.
Rich and precise information carried by text is important in many vision-based application scenarios.
However, extracting text from natural scenes and using it in another application is a complex process.
As illustrated in Figure~\ref{Figure_e2e_system}, various fundamental problems were defined at various stages of this task in the literature: text localization, text verification, text detection, text segmentation, text recognition, and end-to-end systems.
Moreover, special text-related issues exist because of the unique challenges of text.
Text enhancement, text tracking, and natural language processing (NLP) are also briefly introduced.
A clear understanding of these common concepts can help researchers to analyze the differences and connections between different tasks.

\subsubsection{Fundamental Problems}
\label{fundamental_problems}
% \subsubsection{Text Localization}
% \label{text_localization}

\begin{itemize} 
\item \textbf{Text localization:}
The objective of text localization \cite{lienhart2002localizing} is to localize text components precisely and to group them into candidate text regions with as little background as possible \cite{ye2014text}.
% aims at determining the image positions of candidate text.
% It is the basis of text detection, coarsely classifying components and grouping them to precisely provide text candidate regions.
Early text localization methods are based on low-level features, such as color \cite{lee2010scene}, \cite{yi2011text}, gradient \cite{li2008adaptive}, \cite{shivakumara2009gradient}, stroke width transform \cite{epshtein2010detecting}, \cite{mosleh2012image}, maximally stable extremal regions (MSER) \cite{neumann2010method}, \cite{shi2013scene}, canny detector \cite{canny1986computational}, \cite{cho2016canny}, and connected component analysis \cite{Hyung2013text}, \cite{Yi2011AHybrid}.
Most of current methods are based on deep neural networks \cite{he2017single}, \cite{zhan2019esir}, \cite{yin2019video}.

\item \textbf{Text verification:}
Text verification \cite{LaffertyMP01} aims at verifying the text candidate regions as text or non-text.
It is usually used after text localization to filter the candidate regions, because text localization sometimes introduces false positives.
Approaches to text verification include prior knowledge \cite{kim2008new}, \cite{li2008adaptive}, \cite{shivakumara2010accurate}, support vector machine (SVM) classifier \cite{ye2005fast}, and conditional random fields (CRFs) \cite{LaffertyMP01}.
% For example, Lafferty et al. \cite{LaffertyMP01} uses a Conditional Random Fields (CRFs) models to verify candidate regions.
Recent works \cite{wang2012end}, \cite{jaderberg2014deep} used a convolution neural network (CNN) to improve text/non-text discrimination.

\item \textbf{Text detection:}
The function of text detection \cite{yao2012detecting}, \cite{wang2020r} is to determine whether text is present using localization and verification procedures \cite{ye2014text}.
As a basis of an end-to-end system, it provides precise and compact text instance images for text recognition.
Text detection approaches can be roughly categorized as regression-based methods \cite{liu2017deep}, \cite{liu2019curved}, \cite{zhang2018feature}, \cite{baek2019character}, \cite{xie2019derpn} and instance segmentation-based methods \cite{wu2017self}, \cite{he2017multi}, \cite{xu2019textfield}, \cite{liu2019arbitrarily}.

\item \textbf{Text segmentation:}
Text segmentation has been identified as one of the most challenging problems \cite{von2008recaptcha}.
It includes text line segmentation \cite{ye2003robust}, \cite{shivakumara2010accurate} and character segmentation \cite{nomura2005novel}, \cite{shivakumara2011new}.
The former refers to splitting a region of multiple text lines into multiple sub-regions of single text lines.
The latter refers to separating a text instance into multiple regions of single characters.
Character segmentation was typically used in early text recognition approaches \cite{wang2011end}, \cite{mishra2012scene}, \cite{bissacco2013photoocr}.
% It attempts to separate a text instance into multiple regions of single character.
% Adaptive methods are developed for text segmentation, such as the gradient vector flow features \cite{phan2011gradient} , clustering \cite{shivakumara2011new} and morphological operation \cite{nomura2005novel}.

\item \textbf{Text recognition:}
Text recognition \cite{wang2011end} translates a cropped text instance image into a target string sequence.
It is an important component of an end-to-end system, which provides credible recognition results.
Traditional text recognition methods rely on hand-crafted features, such as histogram of oriented gradients descriptors \cite{wang2012end}, connected components \cite{neumann2012real}, and stroke width transform \cite{yao2014unified}.
Most recent studies have used deep learning encoder-decoder frameworks \cite{cheng2017focusing}, \cite{cluo2019moran}, \cite{shi2018aster}.

\item \textbf{End-to-end system:}
% , also known as text spotting system, aims at
Given a scene text image, an end-to-end system \cite{wang2011end} can directly convert all text regions into the target string sequences.
It usually includes text detection \cite{yan2018fast}, text recognition \cite{shi2017end}, and postprocessing.
The construction of a real-time and efficient end-to-end systems \cite{neumann2012real}, \cite{bissacco2013photoocr}, \cite{li2017towards} has become a new trend in recent years.
Some researchers \cite{wang2011end}, \cite{wang2012end}, \cite{Alsharif2014end} interpret text detection and text recognition as two independent subproblems, which are combined to construct an end-to-end system.
Another approach \cite{jaderberg2014deep}, \cite{busta2017deep}, \cite{he2018end}, \cite{Yuliang2020ABCNet} is to jointly optimize text detection and text recognition by sharing information.
% For example, Google PhotoOCR \cite{bissacco2013photoocr} takes advantage of large scale language modeling and careful engineering to construct an effective end-to-end system.  
% He et al. \cite{he2018end} makes up their text spotting system by adapting EAST and a attention-based recognition branch.
\end{itemize}
\subsubsection{Special Issues}
\label{special_issues}
% \subsubsection{Text Enhancement}
% \label{text_enhancement}
\begin{itemize} 
\item \textbf{Script identification:}
Script identification \cite{cheng2019patch} aims to predict the script of a given text image.
It plays a increasingly important role in multilingual systems.
Detecting the script and language helps text recognition to select the correct language model \cite{unnikrishnan2009combined}.
Script identification can be interpreted as an image classification problem, where discriminative representations are usually designed, such as mid-level features \cite{singh2016simple}, \cite{shi2016script}, convolutional features \cite{gomez2016fine}, \cite{shi2015automatic}, \cite{mei2016scene}, and stroke-parts representations \cite{gomez2017improving}. 

\item \textbf{Text enhancement:}
Text enhancement \cite{baker2002limits} can recover degraded text, improve text resolution \cite{yuyang2019text}, remove the distortions of text, or remove the background \cite{Luo2020Separating}, which reduces the difficulty of text recognition.
% It can remove the distortions of text and reduce the difficulty of text recognition.
Many algorithms have been investigated for text enhancement and achieved promising results, such as the deconvolution \cite{xu2010two}, \cite{caner2010shape}, learning-based methods \cite{baker2002limits}, and sparse reconstruction \cite{yao2013rotation}. 

\item \textbf{Text tracking:}
The purpose of text tracking \cite{yin2016text}, \cite{shu2018aunified} is to maintain the integrity of text location and track text across adjacent frames in the video. 
Unlike static text in an image, tracking algorithms for moving text must identify precise text region at pixel level or sub-pixel level, because false tracking may blend text with its background or noise text.
Spatial-temporal analysis \cite{zhao2010text}, \cite{liu2011robustly} is usually used for text tracking in the video.
A recent study \cite{rouh2019method} also predicts movement to track characters.

\item \textbf{Natural language processing:}
Natural language processing (NLP) \cite{chowdhury2003natural} explores how to use computers to understand and manipulate natural language text or speech.
NLP is a bridge for human–computer communication.
Text, as the most important type of unstructured data, is the main object of NLP.
There is a wide range of text-based applications of NLP, including machine translation \cite{bahdanau2014neural}, \cite{cheng2019semi}, automatic summarization \cite{zaeem2018privacycheck}, \cite{liu2018toward}, question answering \cite{anderson2018bottom}, \cite{das2018embodied}, and relationship extraction \cite{xing2018gene}, \cite{li2018employing}.
\end{itemize}

\subsection{Applications}
\label{applications}

Text, as the most important carrier of communication and perception of the world, enriches our lives.
% It is hard to imagine what the world would look like without text.
% , as one of the most influential inventions of humanity, carries the rich and precise semantic information related to scenes, which can be very beneficial to understand the scenes.
Numerous applications of scene text recognition across various industries and in our daily life:
i) Intelligent transportation. 
% With the development of the economy, tourism has become an essential activity in life \cite{aratuo2019industry}.
Constructing automatic geocoding systems \cite{Sermanet2012convolutional}, \cite{xie2018new} is not only convenient to travel, but also enables users to overcome language barriers, e,g., automatically recognizing the road signs \cite{chen2004automatic} and translating text into another language. 
ii) Information extraction.
Although text in the pictures contains precise information, it is almost impossible to type-in massive data by human alone in the era of big data.
For example, the number of China's express delivery business has exceeded $40$ billion \cite{huang2019express} in $2017$.
Automatically recognizing text in natural scenes can save huge resources as well as protect customer privacy.
iii) Visual input and access.
According to the World Health Organization\footnote{\url{https://www.who.int/health-topics/blindness-and-vision-loss}}, at least $2.2$ billion people live in the world with a vision impairment or blindness.
In addition to advanced medical methods, scene text recognition technology can also improve their life, e.g., developing text-to-speech devices to help understand books, ATM instructions and pharmaceutical labels \cite{ezaki2005improved}.
Apart from the applications we have mentioned above, there have been some specific STR application scenarios, such as text visual question answering (text VQA) \cite{singh2019towards}, \cite{biten2019scene}, e-discovery \cite{bai2018integrating}, multimedia retrieval \cite{zhong2000automatic}, \cite{ye2005fast}, automatic identity authentication, which are also quietly changing our life quality.

\section{Methodologies}
\label{methodologies}

% STR has attracted the great interest of numerous researchers and practitioners owing to its unique importance and challenge.

In early research, hand-crafted features were used for text recognition, such as histogram of oriented gradients descriptors \cite{wang2012end}, connected components \cite{neumann2012real}, and stroke width transform \cite{yao2014unified}.
However, the performances of these methods are limited by low-capacity features. 
With the rise and development of deep learning, the community has witnessed substantial advancements in  innovation, practicality, and efficiency of various methods.
Comparing with traditional methods, deep learning methods have the following advantages:
i) Automation: automatic feature representation learning can free researchers from empirically designing the hand-crafted features.
ii) Effectiveness: excellent recognition performance far exceeds traditional algorithms. 
iii) Generalization: algorithms can be easily applied to similar vision-based problems.
% Researchers use the advantages of deep learning to improve the performance of recognition algorithms from different viewpoints.
In this section, we introduce new insights and ideas proposed for STR and end-to-end systems in the era of deep learning.
The primary contribution of each approach is reviewed.
 % with respect to its primary contribution.
In the case of multiple contributions, we analyze them separately.

%\subsection{Scene Text Recognition}
\subsection{Cropped Scene Text Image Recognition}
\label{scene_text_recognition}

The objective of STR is to translate a cropped text instance image into a target string sequence.
There are two types of scene text in nature, i.e., regular and irregular.
% As demonstrated in Figure~\ref{Figure_Methods}, 
Two main STR categories exist: segmentation-based methods and segmentation-free methods.
For segmentation-free methods, they can be roughly classified into CTC-based \cite{graves2006connectionist} methods \cite{shi2017end} and attention-based \cite{bahdanau2014neural} methods \cite{shi2018aster}, \cite{cluo2019moran}.
% We will introduce them in the following subsections accordingly.
Besides, other promising ideas are also introduced in this section, such as label embedding \cite{almazan2014word}, \cite{rodriguez2015label}.
Table~\ref{Table_Recognition_Classify} gives a comprehensive list and categorization of these recognition methods.
% compares the characteristics of these recognition methods.

% \begin{figure*}[htbp]
% \centering
% \includegraphics[width=0.75\textwidth]{Methods.eps}
% \caption{Overview of recent scene text recognition approaches.
% }
% \label{Figure_Methods}
% \end{figure*}

\subsubsection{Segmentation-Based Methods}
\label{segmentation_based_methods}

One category of STR approaches is based on segmentation \cite{bissacco2013photoocr}, \cite{wang2012end}, \cite{jaderberg2014deep}, which usually includes three steps: image preprocessing, character segmentation, and character recognition.
Segmentation-based methods attempt to locate the position of each character from the input text instance image, apply a character classifier to recognize each character, and group characters into text lines to obtain the final recognition results.

An early successful STR system based on deep learning was developed by Wang et al. \cite{wang2011end}, which used a pictorial model that took the scores and locations of characters as input to determine an optimal configuration of a particular word from a small lexicon.
The proposed recognition algorithm outperformed a leading commercial OCR engine ABBYY FineReader\footnote{\url{http://finereader.abbyy.com}}, which is a baseline for STR.
Later, inspired by the success of the deep convolutional neural network in visual understanding \cite{lecun1998gradient}, Wang et al. \cite{wang2012end}, Mishra et al. \cite{mishra2016enhancing}, and Liu et al. \cite{liu2016scene} combined a multilayer neural network with unsupervised feature learning to train a highly-accurate character recognizer module.
For postprocessing, the character responses with character spacings, the beam search algorithm \cite{liu2002lexicon} or the weighted finite state transducer \cite{mohri2002weighted} based representation were applied to recognize target words in a defined lexicon.
To further improve recognition performance, researchers explored robust word image representations, such as scale invariant feature transform (SIFT) descriptors \cite{quy2013recognizing}, Strokelets \cite{yao2014strokelets}, and mid-level features \cite{gordo2015supervised}.

All of the aforementioned methods rely on lexicons to obtain the final recognition results.
However, the query time linearly depends on the size of the lexicon.
With an open lexicon, these strategies are impractical because of the large search space.
% Stronger character representation and larger scale data are required in this case.
To address this issue, lexicon-free attempts had been made for STR.
Some researchers \cite{mishra2012scene} overcame the need for restricted word lists by adopting large dictionaries as higher-order statistical language models.
Others solved STR in a lexicon-free manner by leveraging larger-scale data \cite{bissacco2013photoocr} and more complex neural networks \cite{jaderberg2014deep}, \cite{guo2016convolutional}, e.g., convolutional Maxout network \cite{goodfellow2013maxout}.
Recently, Wan et al. \cite{wan2019textscanner} built a recognition system based on semantic segmentation, which could predict the class and geometry information of characters with two separate branches and further improve recognition performance.

\begin{table*}[htbp]
\tiny
\caption{Summary of the existing recognition approaches. `SK', `ST', `ExPu', `ExPr', and `Un' indicate the approaches that use Synth$90$K dataset, SynthText dataset, extra public data, extra private data, and unknown data, respectively. `Regular' indicates the objective is regular datasets where most text instances are frontal and horizontal. `Irregular' indicates the objective is irregular datasets where most of the text instances are low-resolution, perspective distorted, or curved.
`*' indicates the methods that use the extra datasets other than Synth$90$k and SynthText.}
\label{Table_Recognition_Classify}
\resizebox{\textwidth}{!}{
  \begin{tabular}{ccccccccc}
  \toprule
  \textbf{Method} & \textbf{Year} & \textbf{Data} & \textbf{Regular} & \textbf{Irregular} & \textbf{Segmentation} & \textbf{CTC} & \textbf{Attention} & \textbf{Source Code}  \\
  \midrule %\midrule
  Wang et al. \cite{wang2011end} : ABBYY & 2011  & Un    & {\Checkmark} & {\XSolidBrush} & \Checkmark     & {\XSolidBrush} & \XSolidBrush     & \XSolidBrush \\
  % \midrule
  Wang et al. \cite{wang2011end} : SYNTH+PLEX & 2011  & ExPr  & {\Checkmark} & {\XSolidBrush} & \XSolidBrush     & {\XSolidBrush} & \XSolidBrush     & \XSolidBrush \\
  % \midrule
  Mishra et al. \cite{mishra2012scene} & 2012  & ExPu  & {\Checkmark} & {\XSolidBrush} & {\Checkmark} & {\XSolidBrush} & \XSolidBrush     & \XSolidBrush \\
  % \midrule
  Wang et al. \cite{wang2012end} & 2012  & ExPr  & {\Checkmark} & {\XSolidBrush} & {\Checkmark} & {\XSolidBrush} & \XSolidBrush     & \XSolidBrush \\
  % \midrule
  Goel et al. \cite{goel2013whole} : wDTW & 2013  & Un    & {\Checkmark} & {\XSolidBrush} & {\Checkmark} & {\XSolidBrush} & \XSolidBrush     & \XSolidBrush \\
  % \midrule
  Bissacco et al. \cite{bissacco2013photoocr} : PhotoOCR & 2013  & ExPr  & {\Checkmark} & {\XSolidBrush} & {\Checkmark} & {\XSolidBrush} & \XSolidBrush     & \XSolidBrush \\
  % \midrule
  Phan et al. \cite{quy2013recognizing} & 2013  & ExPu  & \XSolidBrush     & \Checkmark     & {\Checkmark} & {\XSolidBrush} & \XSolidBrush     & \XSolidBrush \\
  % \midrule
  Alsharif et al. \cite{Alsharif2014end} : HMM/Maxout & 2014  & ExPu  & {\Checkmark} & {\XSolidBrush} & {\Checkmark} & {\XSolidBrush} & \XSolidBrush     & \XSolidBrush \\
  % \midrule
  Almazan et al \cite{almazan2014word} : KCSR & 2014  & ExPu  & {\Checkmark} & {\XSolidBrush} & \XSolidBrush     & {\XSolidBrush} & \XSolidBrush     & {https://github.com/almazan/watts} \\
  % \midrule
  Yao et al. \cite{yao2014strokelets} : Strokelets & 2014  & ExPu  & {\Checkmark} & {\XSolidBrush} & {\Checkmark} & {\XSolidBrush} & \XSolidBrush     & \XSolidBrush \\
  % \midrule
  R.-Serrano et al. \cite{rodriguez2015label} : Label embedding & 2015  & ExPu  & {\Checkmark} & {\XSolidBrush} & \XSolidBrush     & {\XSolidBrush} & \XSolidBrush     & \XSolidBrush \\
  % \midrule
  Jaderberg et al. \cite{jaderberg2014deep} & 2014  & ExPu  & {\Checkmark} & {\XSolidBrush} & {\Checkmark} & {\XSolidBrush} & \XSolidBrush     & {https://bitbucket.org/jaderberg/eccv2014\_textspotting/src/master/} \\
  % \midrule
  Su and Lu \cite{su2014accurate} & 2014  & ExPu  & {\Checkmark} & {\XSolidBrush} & \XSolidBrush     & \Checkmark     & \XSolidBrush     & \XSolidBrush \\
  % \midrule
  Gordo \cite{gordo2015supervised} : Mid-features & 2015  & ExPu  & {\Checkmark} & {\XSolidBrush} & {\Checkmark} & {\XSolidBrush} & \XSolidBrush     & \XSolidBrush \\
  % \midrule
  Jaderberg et al. \cite{jaderberg2016reading} & 2015  & ExPr  & {\Checkmark} & {\XSolidBrush} & \XSolidBrush     & {\XSolidBrush} & \XSolidBrush     & {http://www.robots.ox.ac.uk/\~{}vgg/research/text/} \\
  % \midrule
  Jaderberg et al. \cite{jaderberg2015deep} & 2015  & SK + ExPr & {\Checkmark} & {\XSolidBrush} & \XSolidBrush     & {\XSolidBrush} & \XSolidBrush     & \XSolidBrush \\
  % \midrule
  Shi, Bai, and Yao \cite{shi2017end} : CRNN & 2017  & SK    & {\Checkmark} & {\XSolidBrush} & \XSolidBrush     & \Checkmark     & \XSolidBrush     & \tabincell{c}{https://github.com/bgshih/crnn \\https://github.com/meijieru/crnn.pytorch} \\
  % \midrule
  Shi et al. \cite{shi2018aster} : RARE & 2016  & SK    & \XSolidBrush     & \Checkmark     & \XSolidBrush     & {\XSolidBrush} & {\Checkmark} & \XSolidBrush \\
  % \midrule
  Lee and Osindero \cite{lee2016recursive} : R2AM & 2016  & SK    & {\Checkmark} & {\XSolidBrush} & \XSolidBrush     & {\XSolidBrush} & {\Checkmark} & \XSolidBrush \\
  % \midrule
  Liu et al. \cite{liu2016star} : STAR-Net & 2016  & SK + ExPr & \XSolidBrush     & \Checkmark     & \XSolidBrush     & \Checkmark     & \XSolidBrush     & \XSolidBrush \\
  % \midrule
  *Liu et al. \cite{liu2016scene} & 2016  & ExPu  & \Checkmark     & \XSolidBrush     & \Checkmark     & \XSolidBrush     & \XSolidBrush     & \XSolidBrush \\
  % \midrule
  *Mishra et al. \cite{mishra2016enhancing} & 2016  & ExPu  & \Checkmark     & \XSolidBrush     & \Checkmark     & \XSolidBrush     & \XSolidBrush     & \XSolidBrush \\
  % \midrule
  *Su and Lu \cite{su2017accurate} & 2017  & SK + ExPu & \Checkmark     & \XSolidBrush     & \XSolidBrush     & \Checkmark     & \XSolidBrush     & \XSolidBrush \\
  % \midrule
  *Yang et al. \cite{yang2017learning} & 2017  & ExPu  & \XSolidBrush     & \Checkmark     & \XSolidBrush     & {\XSolidBrush} & {\Checkmark} & \XSolidBrush \\
  % \midrule
  Yin et al. \cite{yin2017scene} & 2017  & SK    & {\Checkmark} & {\XSolidBrush} & \XSolidBrush     & \Checkmark     & \XSolidBrush     & \XSolidBrush \\
  % \midrule
  Wang et al. \cite{wang2017gated} : GRCNN & 2017  & SK    & \Checkmark     & \XSolidBrush     & \XSolidBrush     & \Checkmark     & \XSolidBrush     & https://github.com/Jianfeng1991/GRCNN-for-OCR \\
  % \midrule
  *Cheng et al. \cite{cheng2017focusing} : FAN & 2017  & SK + ST (Pixel-level) & {\Checkmark} & {\XSolidBrush} & \XSolidBrush     & {\XSolidBrush} & {\Checkmark} & \XSolidBrush \\
  % \midrule
  Cheng et al. \cite{cheng2018aon} : AON & 2018  & SK + ST  & \XSolidBrush     & \Checkmark     & \XSolidBrush     & {\XSolidBrush} & {\Checkmark} & \XSolidBrush \\
  % \midrule
  Liu et al. \cite{liu2018char} : Char-Net & 2018  & SK  & \XSolidBrush     & \Checkmark     & \XSolidBrush & {\XSolidBrush} & {\Checkmark} & \XSolidBrush \\
  % \midrule
  *Liu et al. \cite{liu2018squeezedtext} : SqueezedText & 2018  & ExPr  & {\Checkmark} & {\XSolidBrush} & \XSolidBrush     & {\XSolidBrush} & \XSolidBrush     & \XSolidBrush \\
  % \midrule
  *Zhan et al. \cite{zhan2018verisimilar} & 2018  & Pr(5 million) & \Checkmark     & \XSolidBrush     & \XSolidBrush     & \Checkmark     & \XSolidBrush     & \tabincell{c}{https://github.com/fnzhan/Verisimilar-Image-Synthesis\\-for-Accurate-Detection-and-Recognition-of-Texts-in-Scenes} \\
  % \midrule
  *Bai et al. \cite{bai2018edit} : EP & 2018  & SK + ST (Pixel-level) & {\Checkmark} & {\XSolidBrush} & \XSolidBrush     & {\XSolidBrush} & {\Checkmark} & \XSolidBrush \\
  % \midrule
  Fang et al. \cite{fang2018attention} & 2018  & SK + ST & \Checkmark     & \XSolidBrush     & \XSolidBrush     & \XSolidBrush     & \Checkmark     & https://github.com/FangShancheng/conv-ensemble-str \\
  % \midrule
  Liu et al. \cite{liu2018connectionist} : EnEsCTC & 2018  & SK    & \Checkmark     & \XSolidBrush     & \XSolidBrush     & \Checkmark     & \XSolidBrush     & https://github.com/liuhu-bigeye/enctc.crnn \\
  % \midrule
  Liu et al. \cite{liu2018synthetically} & 2018  & SK    & {\Checkmark} & {\XSolidBrush} & \XSolidBrush     & \Checkmark     & \XSolidBrush     & \XSolidBrush \\
  % \midrule
  Wang et al. \cite{wang2018memory} : MAAN & 2018  & SK    & \Checkmark     & \XSolidBrush     & \XSolidBrush     & \XSolidBrush     & \Checkmark     & \XSolidBrush \\
  % \midrule
  Sheng et al. \cite{sheng2018nrtr} : NRTR & 2018  & SK    & \Checkmark     & \XSolidBrush     & \XSolidBrush     & \XSolidBrush     & \Checkmark     & \XSolidBrush \\
  Gao et al. \cite{gao2018dense} & 2018  & SK    & \Checkmark     & \XSolidBrush     & \XSolidBrush     & \Checkmark     & \Checkmark     & \XSolidBrush \\
  % \midrule
  Shi et al. \cite{shi2018aster} : ASTER & 2018  & SK + ST & \XSolidBrush     & \Checkmark     & \XSolidBrush     & {\XSolidBrush} & {\Checkmark} & {https://github.com/bgshih/aster} \\
  % \midrule
  Luo et al. \cite{cluo2019moran} : MORAN & 2019  & SK + ST & \XSolidBrush     & \Checkmark     & \XSolidBrush     & {\XSolidBrush} & {\Checkmark} & {https://github.com/Canjie-Luo/MORAN\_v2} \\
  % \midrule
  Luo et al. \footnote{Available at \url{https://github.com/Canjie-Luo/MORAN_v2}}: MORAN-v2 & 2019  & SK + ST & \XSolidBrush     & \Checkmark     & \XSolidBrush     & {\XSolidBrush} & {\Checkmark} & {https://github.com/Canjie-Luo/MORAN\_v2} \\
  % \midrule
  Chen et al. \cite{chen2019adaptive} : AEG & 2019  & SK + ST & \XSolidBrush     & \Checkmark     & \XSolidBrush     & {\XSolidBrush} & {\Checkmark} & \XSolidBrush \\
  % \midrule
  Xie et al. \cite{xie2019convolutional} : CAN & 2019  & SK    & {\Checkmark} & {\XSolidBrush} & \XSolidBrush     & {\XSolidBrush} & {\Checkmark} & \XSolidBrush \\
  % \midrule
  *Liao et al. \cite{liao2019scene} : CA-FCN & 2019  & SK + ST+ ExPr & \XSolidBrush     & \Checkmark     & \XSolidBrush & {\XSolidBrush} & {\Checkmark} & \XSolidBrush \\
  % \midrule
  *Li et al. \cite{li2019show} : SAR & 2019  & SK + ST + ExPr & \XSolidBrush     & \Checkmark     & \XSolidBrush     & {\XSolidBrush} & {\Checkmark} & \tabincell{c}{https://github.com/wangpengnorman/\\SAR-Strong-Baseline-for-Text-Recognition} \\
  % \midrule
  Zhan el at. \cite{zhan2019esir}: ESIR & 2019  & SK + ST & \XSolidBrush     & \Checkmark     & \XSolidBrush     & {\XSolidBrush} & {\Checkmark} & \XSolidBrush \\
  % \midrule
  Zhang et al. \cite{zhang2019sequence}: SSDAN & 2019  & SK    & {\Checkmark} & {\XSolidBrush} & {\XSolidBrush} & {\XSolidBrush} & {\Checkmark} & \XSolidBrush \\
  % \midrule
  *Yang et al. \cite{yang2019symmetry}: ScRN & 2019  & SK + ST(Char-level + Word-level) & \XSolidBrush     & \Checkmark     & \XSolidBrush     & \XSolidBrush     & \Checkmark     & \XSolidBrush \\
  *Yang et al. \cite{wang2019simple} & 2019  & SK + ST + ExPu & \XSolidBrush     & \Checkmark     & \XSolidBrush     & \XSolidBrush     & \Checkmark     & \XSolidBrush \\
  % \midrule
  Wang et al. \cite{wang2019scene}: GCAM & 2019  & SK + ST & \Checkmark     & \XSolidBrush     & \XSolidBrush     & \XSolidBrush     & \Checkmark     & \XSolidBrush \\
  % \midrule
  Jeonghun et al. \cite{baek2019wrong} & 2019  & SK + ST & \XSolidBrush     & \Checkmark     & \XSolidBrush     & \XSolidBrush     & \Checkmark     & {https://github.com/clovaai/deep-text-recognition-benchmark} \\
  % \midrule
  Huang et al. \cite{huang2019epan} : EPAN & 2019  & SK + ST & \XSolidBrush     & \Checkmark     & \XSolidBrush     & \XSolidBrush     & \Checkmark     & \XSolidBrush \\
  % \midrule
  Gao et al. \cite{gao2019reading} & 2019  & SK    & \Checkmark     & \XSolidBrush     & \XSolidBrush     & \Checkmark     & \XSolidBrush     & \XSolidBrush \\
  % \midrule
  *Qi et al. \cite{qi2019novel} : CCL & 2019  & SK + ST(Char-level + Word-level) & \Checkmark     & \XSolidBrush     & \XSolidBrush     & \Checkmark     & \XSolidBrush     & \XSolidBrush \\
  % \midrule
  *Wang et al. \cite{wang2019reelfa} : ReELFA & 2019  & ST(Char-level + Word-level) & \XSolidBrush     & \Checkmark     & \XSolidBrush     & \XSolidBrush     & \Checkmark     & \XSolidBrush \\
  % \midrule
  *Zhu et al. \cite{zhu2019text} : HATN & 2019  & SK + Pu & \XSolidBrush     & \Checkmark     & \XSolidBrush     & \XSolidBrush     & \Checkmark     & \XSolidBrush \\
  % \midrule
  *Zhan et al. \cite{zhan2019spatial} : SF-GAN & 2019  & Pr(1 million) & \Checkmark     & \XSolidBrush     & \XSolidBrush     & \XSolidBrush     & \Checkmark     & \XSolidBrush \\
  % \midrule
  Liao et al. \cite{liao2019mask} : SAM & 2019  & SK + ST & \XSolidBrush     & \Checkmark     & \XSolidBrush     & \XSolidBrush     & \Checkmark     & https://github.com/MhLiao/MaskTextSpotter \\
  % \midrule
  *Liao et al. \cite{liao2019mask} : seg-SAM & 2019  & SK + ST (Char-level)) & \XSolidBrush     & \Checkmark     & \XSolidBrush     & \XSolidBrush     & \Checkmark     & https://github.com/MhLiao/MaskTextSpotter \\
  % \midrule
  Wang et al. \cite{wang2019decoupled} : DAN & 2020  & SK + ST & \XSolidBrush     & \Checkmark     & \XSolidBrush     & \XSolidBrush     & \Checkmark     & https://github.com/Wang-Tianwei/Decoupled-attention-network \\
  Wang et al. \cite{wang2019textsr} & 2020  & SK + ST & \XSolidBrush     & \Checkmark     & \XSolidBrush     & \XSolidBrush     & \Checkmark     & https://github.com/xieenze/TextSR \\
  *Wan et al. \cite{wan2019textscanner} : TextScanner & 2020  & SK + ST (Char-level) & \XSolidBrush     & \Checkmark     & \Checkmark     & \XSolidBrush     & \XSolidBrush     & \XSolidBrush \\
  *Hu et al. \cite{hu2020gtc} : GTC & 2020  & SK + ST + ExPu & \XSolidBrush     & \Checkmark     & \XSolidBrush     & \Checkmark     & \Checkmark     & \XSolidBrush \\
  Luo et al. \cite{Luo2020Separating} & 2020  & SK + ST & \XSolidBrush     & \Checkmark     & \XSolidBrush     & \XSolidBrush     & \Checkmark     & \XSolidBrush \\
  *Litman et al. \cite{Ron2020Scatter} & 2020  & SK + ST + ExPu & \XSolidBrush     & \Checkmark     & \XSolidBrush     & \XSolidBrush     & \Checkmark     & \XSolidBrush \\
  Yu et al. \cite{Deli2020Towards} & 2020  & SK + ST & \XSolidBrush     & \Checkmark     & \XSolidBrush     & \XSolidBrush     & \Checkmark     & \XSolidBrush \\
  Qiao et al. \cite{qiao2020seed} & 2020  & SK + ST & \XSolidBrush     & \Checkmark     & \XSolidBrush     & \XSolidBrush     & \Checkmark     & https://github.com/Pay20Y/SEED \\
  \bottomrule
\end{tabular}}
% \begin{tablenotes}
%  \item[1] Related resources and more information are collected and compiled in our Github repository: \url{https://github.com/HCIILAB/Scene-Text-Recognition}.
% \end{tablenotes}
\end{table*}%

% Mishra et al. \cite{mishra2012scene} presented a framework to recognize scene words, which overcame the need for restricted word lists by adopting large dictionaries as higher order statistical language models.

% Benefit from substantial progress in deep learning and large scale language modeling , Bissacco et al. \cite{bissacco2013photoocr} trained a high-accurate character classifier on two million examples, and a language model on more than a trillion tokens to solve the recognition problem in a lexicon-free manner.

% Jaderberg et al. \cite{jaderberg2014deep} developed a CNN classifier for scene text recognition, which enabled efficient feature sharing for text detection, character case-sensitive and insensitive classification, and bigram classification.

% In \cite{guo2016convolutional}, a novel lexicon-free approach are proposed for character and word recognition problem by leveraging the convolutional Maxout network \cite{goodfellow2013maxout} along with hybrid HMM models.
% \cite{Luo2020Separating}
% Although segmentation-based methods for STR have made significant progress, they have critical shortcomings:
Although significant progress has been made in segmentation-based methods for STR, there are critical shortcomings:
% In contrast to traditional recognizer based on hand-crafted descriptors, some progress has been made by segmentation-based methods.
% However, segmentation-based methods 
% it faces some critical shortcomings: 
i) All these pipelines require accurate detection of individual characters, which has been identified as one of the most challenging problems in the community\cite{von2008recaptcha}. 
Therefore, the quality of character detectors/segmentors usually constrains the recognition performance.
ii) Segmentation-based recognizers fail to model contextual information beyond individual characters, which may result in poor word-level results during the training.

\subsubsection{Segmentation-Free Methods}
\label{segmentation_free_methods}

\begin{figure*}[t]
\centering
\includegraphics[width=1\textwidth]{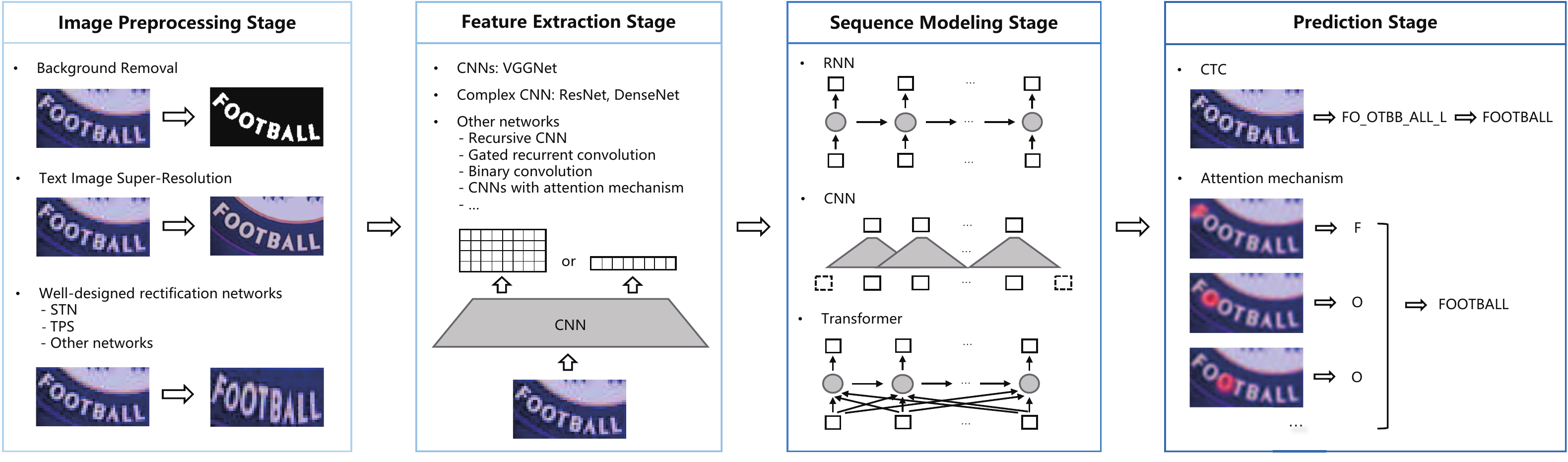}
\caption{Overview of segmentation-free STR approach.
Image preprocessing and sequence modeling stages are not necessarily required.
Moreover, elements of image preprocessing can be combined to further improve recognition performance.
% The stages in dashed boxes are not necessarily required.
}
\label{Figure_recognition}
\end{figure*}

The second category is segmentation-free methods \cite{su2014accurate}, \cite{shi2017end}, \cite{cheng2017focusing}, \cite{shi2018aster}, \cite{cluo2019moran}.
The approach is to recognize the text line as a whole and focus on mapping the entire text instance image into a target string sequence directly by a encoder-decoder framework, thus, avoiding character segmentation.
Figure~\ref{Figure_recognition} shows a typical segmentation-free method, which contains the four stages of image preprocessing, feature representation, sequence modeling, and prediction.
% Following the definition in \cite{baek2019wrong}, segmentation-free methods consist of four stages: transformation, feature extraction, sequence modeling and prediction, as demonstrated in Figure~\ref{Figure_recognition}.
% Under each stage, we review recent methods from different perspectives.

\noindent\textbf{Image Preprocessing Stage}
\label{image_preprocessing_stage}

Image preprocessing aims to improve the image quality by mitigating the interferences from imperfect imaging conditions, which may improve feature representation and recognition.
% have dramatic positive effects on feature extraction and the recognition.

\begin{itemize} 
\item \textbf{Background Removal.}
Text may appear in various scenes with complex backgrounds.
Texture features of backgrounds can be visually similar to the text, which causes additional difficulties in recognition.

% Several researchers try to reduce the impact of complex backgrounds on recognition from different perspectives.
% Liu et al. \cite{liu2018synthetically} argued that the clean images tended to be easier to recognize than the original images, and designed a multi-task network to guide image feature learning toward that of the clean images.
% To better simulate natural scenes and produce realistic text images, Fang et al. \cite{fang2019learning} designed a two-stage synthesis architecture to achieve style consistency of characters. 
% Although the aforementioned methods provided sufficient visualized examples, they do not result in significant improvement in text recognition performance.

% Actually, a brand new perspective of dealing with complex backgrounds is rarely involved in the community
% To deal with complex background, an intuitive solution is to introduce background removal techniques as pre-processing.
Instead of complicated feature representations \cite{liu2018synthetically} and synthesis approaches \cite{fang2019learning}, \cite{wu2019editing}, an intuitive but rarely noticed solution is to separate the text content from complex backgrounds.
Although traditional binarization methods \cite{casey1996survey} work well on document images, they fail to handle substantial variations in text appearance and noise in natural images.
Recently, Luo et al. \cite{Luo2020Separating} used generative adversarial networks (GANs) \cite{goodfellow2014generative} to remove the background while retaining the text contents, which reduced recognition difficulties and dramatically boosted performance.
% Some researchers try to reduce the impact of complex backgrounds on recognition from different perspectives.
% Several attempts \cite{liu2018synthetically} at scene text generation and style transfer have taken a crucial step forward.
% Liu et al. \cite{liu2018synthetically} argued that the clean images tended to be easier to recognize than the original images, and designed a multi-task network to guide image feature learning toward that of the clean images.
% However, the image-level guidance does not result in significant improvement in text recognition performance.

% With the widespread application of GANs, style transfer methods were adapted to 

\item \textbf{Text Image Super-Resolution (TextSR).}
Scene text is usually distorted by various noise interferences, such as low resolution.
Low resolution can lead to misrecognized characters or words.
Text image super-resolution (TextSR) \cite{peyrard2015icdar2015} can output a plausible high-resolution image that is consistent with a given low-resolution image.
This approach can help with text recognition in low-resolution images.

Classical approaches, such as bilinear, bicubic, or designed filtering, aim to reconstruct the detailed texture of natural images, but are not applicable to blurred text \cite{wang2019textsr}.
% Later, super-resolution is simply treated as a regression problem independent of recognition task, where the input is the low-resolution image, and the target output is the high-resolution image \cite{dong2015image}.
% Instead of simply treating super-resolution as a regression problem \cite{dong2015image}, Wang et al. \cite{wang2019textsr} first recognized small text with TextSR methods, which significantly improved the performance of the text recognizer.
Instead of simply treating super-resolution as a regression problem \cite{dong2015image}, Wang et al. \cite{wang2019textsr} first combined TextSR methods with recognition task, which significantly improved the performance of the text recognizer.

\item \textbf{Rectification.}
% \label{rectification}
% Therefore, TextSR is one way to get better recognition performance.
% Different from general object recognition tasks, STR has its unique characteristics.
% Some auxiliary tasks for text are rarely involved in the community, such as text image super-resolution (TextSR) \cite{peyrard2015icdar2015} and separating text content from complex backgrounds \cite{Luo2020Separating}, which can significantly reduce the difficulty of STR and may be potential in future work.
% \paragraph{\textbf{{\color{red}{Rectification Stage}}}}
% \label{rectification_stage}
The function of rectification is to normalize the input text instance image, remove the distortion, and reduce the difficulty of irregular text recognition.
Specifically, irregular text \cite{yang2017learning} refers to text with perspective distortion or arbitrary curving shape, which usually causes additional challenges in recognition. 

The spatial transformer network (STN) \cite{jaderberg2015spatial} was used as an early rectification module to rectify the entire text image \cite{shi2016robust}, \cite{liu2016star} or individual character regions \cite{liu2018char}.
Later, Shi et al. \cite{shi2018aster} and Jeonghun et al. \cite{baek2019wrong} adopted Thin-Plate-Spline (TPS) \cite{warps1989thin} to handle more complex distortions.
Recently, some well-designed rectification networks were proposed.
For example, a multi-object rectification network \cite{cluo2019moran} was developed to rectify irregular text by predicting the offsets of each part of an input image.
Zhan et al. \cite{zhan2019esir} designed a novel line-fitting transformation and an iterative TPS-based rectification framework for optimal scene text rectification.
Based on local attributes, such as center line, scale, and orientation, Yang et al. \cite{yang2019symmetry} proposed a symmetry-constrained rectification network.
 % with negligible extra computations.

To handle a variety of distortions, complex rectification modules are required and become a new trend.
However, these affect the speed and memory consumption of recognition algorithms.
Practitioners should choose the best trade-offs depending on their needs under different application scenarios.
Moreover, with the development of irregular text detection, it is worth reconsidering whether a rectification module is required for a STR system.
\end{itemize}

Image preprocessing includes but is not limited to the aforementioned types.
It can significantly reduce the difficulties of recognition by improving image quality.
Various methods can be used in combination.
Although many recognition algorithms exist, these auxiliary preprocessing approaches for text are not often used in the community, especially for background removal and TextSR.
% Therefore, 
% In addition, different from general object recognition tasks, STR has its unique characteristics.
Moreover, most general off-the-shelf algorithms focus on the style of a single object, whereas scene text images usually contain multiple characters.
Therefore, elaborate and dedicated-design preprocessing algorithms for STR deserve the attention of researchers in future work.
% as scene text image usually contains multiple characters, the general algorithms are 

% Different from general object recognition tasks, STR has its unique characteristics.}}
%  presented the ESIR system to improve the recognition performance of irregular scene text.
% In particular, a novel line-fitting transformation is designed to estimate the pose of text lines in scenes and an iterative TPS-based rectification framework is developed for optimal scene text rectification and recognition.

% MORAN framework to recognize irregular scene text, which consists of a multi-object rectification network and an attention-based sequence recognition network. 
% In particular, the proposed multi-object rectification network rectified irregular scene text images by predicting the offset of each part of the image.

% Transformation Stage is proposed to process text with perspective distortion or curving effects, which we called irregular text.
% Specifically, the various fonts and distorted patterns of irregular text cause additional challenges in recognition.

\noindent\textbf{Feature Representation Stage}
\label{feature_extraction_stage}

Feature representation maps the input text instance image to a representation that reflects the attributes relevant for character recognition, while suppressing irrelevant features such as font, color, size, and background.

Motivated by the successes of \cite{wang2010word}, \cite{mishra2012top}, \cite{wang2011end}, Su et al. \cite{su2014accurate} used the histogram of oriented gradients (HOG) feature \cite{dalal2005histograms} in their STR system to construct sequential features of word images.
Later, CNNs \cite{yin2017scene}, \cite{cheng2018aon}, \cite{liu2018char}, \cite{zhan2018verisimilar}, \cite{cluo2019moran} have been widely used for feature representation stage, such as the VGGNet \cite{simonyan2014very}, \cite{shi2017end}, \cite{shi2016robust}, \cite{jaderberg2015deep}, \cite{cheng2017focusing}, \cite{yang2017learning}, \cite{wang2019reelfa}. 
For more powerful feature representation, some complex neural networks were applied in STR algorithms, such as ResNet \cite{he2016deep} \cite{liu2016star}, \cite{fang2018attention}, \cite{wang2018memory}, \cite{shi2018aster}, \cite{chen2019adaptive}, \cite{wang2019decoupled}, \cite{xie2019convolutional}, \cite{li2019show}, \cite{zhan2019esir}, \cite{yang2019symmetry}, \cite{wang2019scene}, \cite{baek2019wrong}, \cite{qi2019novel}, \cite{zhu2019text}, \cite{zhan2019spatial} and DenseNet \cite{huang2017densely}, \cite{gao2018dense}, \cite{gao2019reading}.
Recently, some attempts have been made to improve the feature representation module from different perspectives.
Recursive CNNs were used by Lee et al. \cite{lee2016recursive} for parametrically efficient and effective image feature representation, which can increase the depth of traditional CNNs under the same parametric capacity and produce much more compact feature response.
Inspired by the recurrent convolution neural network (RCNN) in image classification \cite{liang2015recurrent}, Wang et al. \cite{wang2017gated} designed a gated recurrent convolution layer for feature sequence representation by introducing a gate to control the context modulation in RCNN.
Liu et al. \cite{liu2018squeezedtext} focused on real-time STR and proposed a novel binary convolutional layer.
They claimed that the binary representation remarkably speeds up run-time inference and reduces memory usage.
Some researchers \cite{gao2018dense}, \cite{zhang2019sequence}, \cite{huang2019epan}, \cite{gao2019reading}, \cite{liao2019mask}, \cite{fang2018attention} argued that directly processing the source image by CNNs would introduce extra noise.
Therefore, they combined CNNs with the attention mechanism \cite{bahdanau2014neural} to enhance the representation of foreground text and suppress background noise.

A deeper and more advanced feature extractor usually results in a better representation power, which is suitable for improving STR with complex backgrounds.
However, the performance improvement comes at the cost of memory and computation consumption \cite{baek2019wrong}.
A combination of the background removal technique \cite{elhabian2008moving} with simple feature extractors may be an alternative in future research. 
% , Su et al. \cite{su2014accurate} converted a word image into a sequential column vectors.

\noindent\textbf{Sequence Modeling Stage}
\label{sequence_modeling_stage}

Sequence modeling, as a bridge between visual features and predictions, can capture the contextual information within a sequence of characters for the next stage to predict each character, which is more stable and helpful than treating each symbol independently.

Multiple bidirectional long short term memory (BiLSTM) model was introduced in \cite{graves2008novel} and widely used in \cite{su2014accurate}, \cite{shi2017end}, \cite{shi2016robust}, \cite{liu2016star}, \cite{su2017accurate}, \cite{wang2017gated}, \cite{cheng2018aon}, \cite{liu2018char}, \cite{wang2018memory}, \cite{gao2018dense}, \cite{shi2018aster}, \cite{cluo2019moran}, \cite{chen2019adaptive}, \cite{li2019show}, \cite{zhan2019esir}, \cite{yang2019symmetry}, \cite{wang2019scene}, \cite{baek2019wrong}, \cite{wang2019reelfa}, \cite{wang2019decoupled} as the sequence modeling module because of its ability to capture long-range dependencies.
Litman et al. \cite{Ron2020Scatter} added intermediate supervisions along the network layers and successfully trained a deeper BiLSTM model to improve the encoding of contextual dependencies. 
However, some researchers \cite{yin2017scene}, \cite{fang2018attention}, \cite{xie2019convolutional}, \cite{gao2019reading}, \cite{qi2019novel} considered that BiLSTM was not an essential part of STR algorithms.
They argued that although the BiLSTM was effective to model the context, its structure was computationally intensive and time consuming.
Moreover, it could cause gradient vanishing/exploding during the training.
Therefore, a sliding window \cite{yin2017scene} or deep one-dimensional CNN \cite{fang2018attention}, \cite{xie2019convolutional}, \cite{gao2019reading} was used instead of BiLSTM.
In particular, although CNNs were widely used for feature extraction of individual characters in early research \cite{wang2012end}, \cite{mishra2016enhancing}, \cite{liu2016scene}, the context can also be modeled by CNNs by precisely controlling the receptive field.
Recently, Yu et al. \cite{Deli2020Towards} and Qiao et al. \cite{qiao2020seed} focused on introducing global semantic information to model the context. 
Therefore, Yu et al. \cite{Deli2020Towards} designed a global semantic reasoning module to capture global semantic context through multi-way parallel transmission, while Qiao et al. \cite{qiao2020seed} predicted an additional global semantic information supervised by the word embedding from a pre-trained language model.

Contextual cues are beneficial for image-based sequence recognition.
Although recurrent neural networks (RNNs) \cite{hochreiter1997long} based structures, such as BiLSTM or LSTM, can model character sequences, there are some inherent limitations.
 % hinder their practical applications on STR.
In contrast, CNNs or transformer \cite{vaswani2017attention} can not only effectively deal with long sequences, but also be parallelized efficiently.
Modeling language sequences using CNNs or transformer structure may be a new trend for sequence modeling because of its intrinsic superiority.

\noindent\textbf{Prediction Stage}
\label{prediction_stage}

The objective of the prediction stage is to estimate the target string sequence from the identified features of the input text instance image.
Connectionist temporal classification (CTC) \cite{graves2006connectionist} and the attention mechanism \cite{bahdanau2014neural} are two major techniques.
Moreover, other potential ideas regarding the prediction stage are also introduced in this section.

\subparagraph{\textbf{Connectionist Temporal Classification}}
\label{connectionist_temporal_classification}
% Su et al. \cite{su2014accurate} presented a novel scene text recognition system that made use of the Histogram of Oriented Gradient (HOG) feature \cite{dalal2005histograms} and RNN model.
% In this system, CTC technique was exploited to find out the best match of a list of lexicon words based on the RNN output of the sequential feature using.

% Connectionist Temporal Classification (CTC) \cite{graves2006connectionist} and attention mechanism \cite{bahdanau2014neural} are two widely used methods.
% Moreover, other potential approaches are also introduced in this section, such as label embedding, auxiliary task, etc.

% \paragraph{CTC-based Methods}
% \label{ctc_based_methods}

CTC was proposed by Graves et al. \cite{graves2006connectionist} for training RNNs \cite{cho2014properties}, \cite{hochreiter1997long} to label unsegmented sequences directly.
CTC has achieved significant improvements in many fields, such as speech recognition \cite{graves2013speech}, \cite{graves2014towards} and online handwritten text recognition \cite{graves2008novel}, \cite{graves2012supervised}.
CTC is typically used in STR as a prediction module, i.e., the transcription layer that converts the input features made by CNNs or RNNs into a target string sequence by calculating the conditional probability.
In particular, CTC can maximize the likelihood of an output sequence by efficiently summing over all possible input-output sequence alignments, and allow the classifier to be trained without any prior alignment between the input and target sequences.

The formulation of the conditional probability can be briefly described as follows.
The input features are denoted by $y = (y_{1},y_{2}...y_{T})$, where $T$ is the sequence length.
Each $y_t$ is a probability distribution over $\mathcal{L}$.
Specifically, $\mathcal{L}$ represents a set of all labels, including all characters and an extra blank symbol that represents an invalid output.
A CTC path $\pi$ is a sequence of length $T$, which consists of the blank symbol and label indices.
As there are many possible ways to map these paths to transcription $l$, a CTC mapping function $\mathcal{B}$ is defined to remove repeated labels and delete the blank symbol from each path.
Then, the conditional probability is calculated by summing the probabilities of all paths mapped onto $l$ by $\mathcal{B}$:   
\begin{equation}
p(l|y) = \sum_{\pi:\mathcal{B}(\pi)=l} p(\pi|y),
\end{equation}
where the probability of $\pi$ is defined as $p(\pi|y)=\prod^T_{t=1} y^t_{\pi_t}$, and $y^t_{\pi_t}$ is the probability of having label $\pi_t$ at time step t.
As directly computing the above equation is computationally expensive, most researchers \cite{liu2016star}, \cite{shi2017end}, \cite{yin2017scene} adapt the forward–backward algorithm \cite{graves2008novel} to compute it efficiently. 

Inspired by the success of CTC in speech processing, Su et al. \cite{su2014accurate}, He et al. \cite{He2016reading} and Shi et al. \cite{shi2017end} first applied it to STR.
Since then, numerous CTC-based prediction algorithms \cite{liu2016star}, \cite{su2017accurate}, \cite{yin2017scene}, \cite{wang2017gated}, \cite{gao2018dense}, \cite{gao2019reading}, \cite{qi2019novel} have showed promising transcription performance.
% CTC-based prediction algorithms \cite{su2014accurate}, \cite{shi2017end}, \cite{liu2016star}, \cite{su2017accurate}, \cite{yin2017scene}, \cite{wang2017gated}, \cite{gao2018dense}, \cite{gao2019reading}, \cite{qi2019novel} have showed promising performance for STR. 
However, Liu et al. \cite{liu2018connectionist} argued that CTC tended to produce highly peaky and overconfident distributions, which was a symptom of over-fitting.
To address this issue, they proposed a regularization method based on maximum conditional entropy to enhance generalization and exploration capabilities of CTC.
Feng et al. \cite{feng2019focal} modified the traditional CTC by fusing focal loss to solve the recognition of extremely unbalanced samples.
Recently, Hu et al. \cite{hu2020gtc} improved the accuracy and robustness of CTC by using graph convolutional networks (GCNs) \cite{kipf2016semi} in STR.

CTC enjoys remarkable transcription performance and stability.
However, it faces some inherent limitations:
i) The underlying methodology of CTC is sophisticated, which results in a large computational cost for long text sequences.
ii) CTC suffers from the peaky distribution problems \cite{graves2006connectionist}, \cite{miao2015eesen} and its performance usually degrades for repeated patterns.
% Because of its implementation, 
iii) CTC can hardly be applied to two-dimensional (2D) prediction problems, such as irregular scene text recognition, where characters in the input text instance image are distributed in a spatial structure.
To handle this issue, Wan et al. \cite{wan2019arXiv} extended the vanilla CTC by adding another dimension along the height direction. 
Although the recognition performance is improved to some extent, the proposed 2D-CTC model has not completely solved 2D prediction problems.
Therefore, applying CTC to solve the 2D prediction problem could be a potential direction for future research.

\subparagraph{\textbf{Attention Mechanism}}
\label{attention_mechanism}
The attention mechanism was proposed by Bahdanau et al. \cite{bahdanau2014neural} in the field of neural machine translation, which can automatically search for the predicted word that are relevant to parts of a given source sentence.
Many approaches based on the attention mechanism have achieved significant improvements in various fields, such as image caption \cite{he2019vd}, text recognition \cite{shi2018aster}, and scene classification of remote sensing images \cite{wang2018scene}.
For STR, the attention mechanism is often combined with the RNN structure as a prediction module.
 % to improve the performance of the recognizer.

In particular, the attention mechanism learns the alignment between the input instance image and the output text sequences by referring to the history of the target characters and the encoded feature vectors. 
% The attention-based prediction process can be briefly described as follows.
Let the output prediction sequence be denoted as $o = (o_{1},o_{2}...o_{M})$, where $M$ indicates the maximum decoding step size.
At the $t$-th step, the output prediction $o_t$ is given by:
\begin{equation}
o_t=Softmax(W_{o}s_t+b_{o}),
\label{equation_attn_1}
\end{equation}
where $s_t$ is the hidden state of RNN at time step $t$.
Typically, a gated recurrent unit (GRU) \cite{cho2014properties} is used to update $s_t$ and model the longterm dependencies.
Hence, $s_t$ is computed as
\begin{equation}
\label{cat_equation}
s_t = GRU( o_{prev}, g_t, s_{t-1}),
\end{equation}
where $o_{prev}$ is the embedding vector of the previous output $o_{t-1}$.
Moreover, $g_t$ represents the \textit{glimpse vector}, computing as the weighted sum of features $h=(h_{1},h_{2}...h_{N})$
\begin{equation}
g_t =\sum^N_{j=1}\alpha_{t, j}h_j,
\end{equation}
where $N$ is the feature length.
Here, $\alpha_{t}$ is the vector of attention weights, which is computed as follows:
\begin{equation}
\alpha_{t, j}=\frac{exp(e_{t, j})}{\sum^N_{i=1}exp(e_{t, i})},
\end{equation}
\begin{equation}
e_{t, j}=Tanh(W_ss_{t-1} + W_hh_j + b),
\label{equation_attn_2}
\end{equation}
where $e_{t, j}$ is the alignment score which represents the degree of correlation between the high-level feature representation and the current output.
In the above Equations~\ref{equation_attn_1} – ~\ref{equation_attn_2}, $W_{o}$, $b_{o}$, $W_s$, $W_h$, and $b$ are all trainable parameters.

Inspired by the development of neural machine translation systems, a large number of attention-based methods \cite{lee2016recursive}, \cite{shi2016robust}, \cite{yang2017learning}, \cite{cheng2018aon}, \cite{shi2018aster}, \cite{cluo2019moran}, \cite{li2019show}, \cite{zhan2019esir}, \cite{zhang2019sequence}, \cite{yang2019symmetry}, \cite{baek2019wrong}, \cite{zhan2019spatial} have emerged in STR field.
Moreover, some attempts have been made to improve the vanilla attention from different perspectives:
i) \textbf{Applying to 2D prediction problems.}
For the irregular scene text recognition, the various character placements significantly increase the difficulty of recognition.
The vanilla attention \cite{lee2016recursive} was applied to perform $1$D feature selection and decoding.
There is the significant conflict between $2$D text distribution and $1$D feature representation by applying the vanilla attention directly.
Therefore, Yang et al. \cite{yang2017learning}, Li et al. \cite{li2019show}, and Huang et al. \cite{huang2019epan} proposed $2$D attention mechanism for irregular text recognition.
ii) \textbf{Improving the construction of implicit language model.}
Chen et al. \cite{chen2019adaptive} and Wang et al. \cite{wang2018memory} argued that the generated glimpse vector was not powerful enough to represent the predicted characters.
Therefore, Chen et al. \cite{chen2019adaptive} introduced high-order character language models to the vanilla attention, while Wang et al. \cite{wang2018memory} constructed a memory-augmented attention model by feeding a part of the character sequence already generated and the all attended alignment history.
Shi et al. \cite{shi2018aster} noted that a vanilla attention-based prediction module captured output dependencies in only one direction and missed the other. 
Thus, they proposed a bidirectional attention-based decoder, with two decoders in opposite directions.
iii) \textbf{Improving parallelization and reducing complexity.}
Although the vanilla attention mechanism based on the RNN structure can capture long-range dependencies, it is computationally intensive and time consuming.
A recent attention variant, namely Transformer \cite{vaswani2017attention}, was widely employed in \cite{zhu2019text}, \cite{wang2019simple}, \cite{sheng2018nrtr}, \cite{Deli2020Towards} to improve parallelization and reduce complexity for STR.
iv) \textbf{Addressing attention drift.}
The \textit{attention drift} phenomenon means that attention models cannot accurately associate each feature vector with the corresponding target region in the input image.
Some researchers added extra information to solve this problem by focusing the deviated attention back onto the target areas, such as localization supervision \cite{cheng2017focusing} and encoded coordinates \cite{wang2019reelfa}.
Others \cite{wang2019scene}, \cite{huang2019epan}, \cite{zhu2019text} increased the alignment precision of attention in a cascade way.
Specifically, Wang et al. \cite{wang2019decoupled} argued that a serious alignment problem is caused by its recurrence alignment mechanism.
Therefore, they decoupled the alignment operation from using historical decoding results.

In recent years, the attention-based prediction approaches have become the mainstream method in the field of STR and have outperformed CTC in decoding because of its ability to focus on informative areas.
Moreover, the attentional methods can be easily extended to complex 2D prediction problems.
However, the attention mechanism has some shortcomings:
i) As this method relies on the attention module for label alignment, it requires more storage and computations \cite{bluche2016joint}.
ii) For long text sequences, the attention mechanism is difficult to train from scratch owing to the misalignment between the input instance image and the output text sequences \cite{bahdanau2016end}, i.e., the attention drift phenomenon.
iii) The current research of attention mechanism mainly focuses on languages which involve only a few character categories (e.g., English, French).
% The current attention mechanism may depend on Latin which involves only a few character categories.
To the best of our best knowledge, there is no public report on effectively applying the attention mechanism to deal with the large-scale category text recognition tasks, such as Chinese text recognition.

%   developed the focusing attention network (FAN) to focus the deviated attention back on the target areas.
% Specifically, as illustrated in Figure~\ref{Figure_FAN}, localization supervision is added to the attention module to guide the recognizer automatically learn the alignment. 

\subparagraph{\textbf{Discussion}}
\label{disscussion}

Both CTC and the attention mechanism have their strengths and limitations.
Recently, some researchers \cite{hu2020gtc}, \cite{Ron2020Scatter} applied both CTC and the attention mechanism to achieve accurate prediction and maintain a fast inference speed.
Cong et al. \cite{cong2019acomparative} comprehensively compared these two prediction approaches on large-scale real-world scene text sentence recognition tasks.
Based on extensive experiments, they provided practical advice for researchers and practitioners.
For example, the attention-based approaches can achieve higher recognition accuracy on isolated word recognition tasks but perform worse on sentence recognition tasks compared with CTC-based approaches.
% And it is more effective and efficient to leverage an explicit language model to boost recognition accuracy for CTC based approaches.
Therefore, the right prediction methods should be chosen according to different application scenarios and constraints.
Moreover, it is valuable to explore alternative prediction strategies in future work.
For example, the aggregation cross-entropy function \cite{xie2019aggregation} was designed to replace CTC and the attention mechanism; it achieves competitive performance with a much quicker implementation, reduced storage requirements, and convenient employment.

\subsubsection{Other Potential Approaches}
\label{other_potential_approaches}

Other approaches have been considered and explored with a different view.
Motivated by ``the whole is greater than the sum of parts,'' Goel et al. \cite{goel2013whole} recognized text in natural scenes by matching the scene and synthetic image features with weighted dynamic time warping (wDTW) approach.
Later, Almaz\'{a}n et al. \cite{almazan2014word} and Rodriguez et al. \cite{rodriguez2015label} interpreted the task of recognition and retrieval as a nearest neighbor problem.
They embedded both word images and text strings in a common vectorial subspace or Euclidean space, combining label embedding with attributes learning.
Specifically, images and strings that represent the same word would be close together.
% Similarly, Rodriguez et al. \cite{rodriguez2015label} also regarded STR as a nearest neighbor search, the difference was that they embedded word labels and images into a common Enclidean space, which avoid costly pre-/post-processing operations.
Recently, Jaderberg et al. \cite{jaderberg2016reading} formulated STR as a multi-class classification problem.
They trained a deep CNN-classifier solely on synthetic data: approximately $9$ million images from a $90$k words dictionary. 
As each word corresponds to an output neuron, the proposed text classifier cannot recognize out-of-dictionary words.
Further, they combined CNNs with a CRF graphical model for unconstrained text recognition \cite{jaderberg2015deep}.
% , which incorporated a Conditional Random Field (CRF) graphical model.

% Liu et al. \cite{liu2018squeezedtext} focused on real-time scene text recognition and proposed a novel binary convolutional encoder-decoder network (B-CEDNet).
% They claimed that the binary representation lead to both remarkable inference run-time speedup as well as memory usage reduction for scene text recognition.

% {\color{blue}{Citiation}}
% {\color{blue}{Actually, different from general object recognition tasks, STR has its unique characteristics.
% Some auxiliary tasks for text are rarely involved in the community, such as text image super-resolution (TextSR) \cite{peyrard2015icdar2015} and separating text content from complex backgrounds \cite{Luo2020Separating}, which can significantly reduce the difficulty of STR and may be potential in future work.}}

\begin{table*}[htbp]
\caption{Summary of the existing end-to-end system approaches.}
\label{Table_end_to_end_classify}%
\tiny
\resizebox{\textwidth}{!}{%
  \begin{tabular}{ccccc}
  \toprule
  \textbf{Method} & \textbf{Year} & \textbf{Detection} & \textbf{Recognition} & \textbf{Source Code} \\
  \midrule%\midrule
  Wang et al. \cite{wang2011end} & 2011  & Sliding windows and Random Ferns & Pictorial Structures & \XSolidBrush       \\
  % \midrule
  Wang et al. \cite{wang2012end} & 2012 & CNN-based & Sliding windows for classification  & \XSolidBrush       \\
  % \midrule
  Jaderberg et al. \cite{jaderberg2014deep} & 2014  & CNN-based and saliency maps & CNN classifier & \XSolidBrush       \\
  % \midrule
  Alsharif et al. \cite{Alsharif2014end} & 2014  & CNN and hybrid HMM Maxout models & Segmentation-based & \XSolidBrush      \\
  % \midrule
  Yao et al. \cite{yao2014unified} & 2014  & Random Forest & Component Linking and Word Partition & \XSolidBrush       \\
  % \midrule
  Neumann et al. \cite{neumann2015real} & 2015  & Extremal Regions & Clustering algorithm to group characters & \XSolidBrush      \\
  % \midrule
  Jaderberg et al. \cite{jaderberg2016reading} & 2016  & Region proposal mechanism & Word-level classification & \XSolidBrush      \\
  % \midrule
  Liao et al. \cite{liao2017textboxes} : TextBoxes & 2017  & SSD-based framework & CRNN  & https://github.com/MhLiao/TextBoxes     \\
  % \midrule
  Bŭsta et al. \cite{busta2017deep} : Deep TextSpotter & 2017  & Yolo-v2 & CTC  & \XSolidBrush      \\
  % \midrule
  Li et al. \cite{li2017towards} & 2017  & Text Proposal Network & Attention & \XSolidBrush      \\
  % \midrule
  Lyu et al. \cite{lyu2018mask} : Mask TextSpotter & 2018  & Fast R-CNN with mask branch & Character segmentation & https://github.com/lvpengyuan/masktextspotter.caffe2     \\
  % \midrule
  He et al. \cite{he2018end} & 2018  & EAST framework & Attention & https://github.com/tonghe90/textspotter     \\
  % \midrule
  Liu et al. \cite{liu2018fots} : FOTS & 2018  & EAST framework & CTC   & https://github.com/jiangxiluning/FOTS.PyTorch     \\
  % \midrule
  Liao et al. \cite{liao2018textboxes++} : TextBoxes++ & 2018  & SSD-based framework & CRNN  & https://github.com/MhLiao/TextBoxes\_plusplus     \\
  % \midrule
  Liao et al. \cite{liao2019mask} : Mask TextSpotter & 2019  & Mask RCNN & Character segmentation, Spatial Attention Module & https://github.com/MhLiao/MaskTextSpotter     \\
  % \midrule
  % Neumann et al. \cite{neumann2015efficient} & 2015  & - & - & -     \\
  % \midrule
  Xing et al. \cite{xing2019convo} : CharNet & 2019  & EAST framework & CNN classifier & https://github.com/MalongTech/research-charnet     \\
  Feng et al. \cite{feng2019textdragon} : TextDragon & 2019  & TextSnake & Sliding convolution character models with CTC & \XSolidBrush     \\
  % \midrule
  Qin et al. \cite{qin2019towards} & 2019 & Mask RCNN & Attention & \XSolidBrush      \\
  Wang et al. \cite{wang2019all} : Boundary & 2020 & RPN-based framework & Attention & \XSolidBrush      \\
  Qiao et al. \cite{qiao2019text} : Text Perceptron & 2020 & ResNet and Feature Pyramid Network & Attention & \XSolidBrush      \\
  Liu et al. \cite{Yuliang2020ABCNet} : ABCNet & 2020 & Bezier curve detection & CTC & https://github.com/Yuliang-Liu/bezier\_curve\_text\_spotting      \\
  \bottomrule
  \end{tabular}}
\end{table*}%

\subsection{End-to-End Systems}
\label{end_to_end_system}

Given a text image with a complex background as input, an end-to-end system aims to directly convert all text regions into string sequences.
Typically, it includes text detection, text recognition, and postprocessing.
In the past, text detection and recognition have been interpreted as two independent subproblems that are combined to retrieve text from images \cite{wang2011end}, \cite{wang2012end}, \cite{jaderberg2014deep}, \cite{neumann2015efficient}, \cite{gupta2016synthetic},  \cite{Alsharif2014end}, \cite{jaderberg2016reading}, \cite{neumann2015real}, \cite{liao2017textboxes}.
 % interpreted text detection and recognition as two independent subproblems and combined them to retrieve text from images.
Recently, the construction of real-time and efficient end-to-end systems has become a new trend in the community.
Table~\ref{Table_end_to_end_classify} compares the characteristics of these end-to-end methods.

Several factors promote the emergence of end-to-end systems:
i) Errors can accumulate in a cascade way of text detection and recognition, which may lead to a large fraction of garbage predictions, while an end-to-end system can prevent errors from being accumulated during the training.
ii) In an end-to-end system, text detection and recognition can share information and can be jointly optimized to improve overall performance.
iii) An end-to-end system is easier to maintain and adapt to new domains, whereas maintaining a cascaded pipeline with data and model dependencies requires substantial engineering efforts.
iv) An end-to-end system exhibits competitive performance with faster inference and smaller storage requirements.
% .
% Therefore, they can not complement each other to further improve text spotting performance.

% Recent years, many attempts \cite{busta2017deep} have proven the effectiveness of a joint end-to-end model.
Many recent studies \cite{busta2017deep} have shown the effectiveness of a joint optimized end-to-end model, which usually includes a detection branch and a recognition branch.
% In early work \cite{jaderberg2014deep}, a multi-layer CNN was designed for both branches, which enabled feature sharing by using a number of layers in common.
Bartz et al. \cite{bartz2018see} integrated and jointly learned a STN \cite{jaderberg2015spatial} to detect text regions of an image.
Corresponding image regions were directly cropped and fed into a simple neural network to recognize text content.
% applied STN \cite{jaderberg2015spatial} to detect text regions of an image: they directly cropped corresponding image regions and fed them into a simple neural network to recognize their text content.
Advanced detection \cite{dai2019deep}, \cite{tang2018scene} and recognition algorithms \cite{shi2017end} were then used to build joint end-to-end systems.
Both branches were bridged by cropping region of interests (RoIs) features of the detection branch and feeding them to the recognition branch. 
Typically, RoIPool was proposed by Girshick \cite{girshick2015fast} to convert RoIs of different scales and aspect ratios into fixed-size feature maps for object detection. 
However, this approach may lead to significant distortion because of the large variation of text length.
To address this issue, Li et al. \cite{li2017towards} proposed varying-size RoIPool to accommodate the original aspect ratios. 
As quantizations performed by RoIPool would introduce misalignments between the RoIs and the extracted features, many methods used bilinear interpolation to extract text instance features, such as bilinear sampling \cite{busta2017deep}, RoIRotate \cite{liu2018fots}, and the text alignment layer \cite{he2018end}.
Recent end-to-end systems \cite{lyu2018mask}, \cite{liao2019mask}, \cite{feng2019textdragon}, \cite{wang2019all}, \cite{qiao2019text} have focused on curved text of arbitrary shapes. 
For example, Liao et al. \cite{lyu2018mask} and their extended work \cite{liao2019mask} used RoIAlign \cite{he2017mask} to preserve more accurate location information, retrieved each character as a generic object, and composed the final text with character-level annotations. 
Feng et al. \cite{feng2019textdragon} generated dense detection quadrangles and used the proposed RoISlide to transform features cropped from each quadrangle into rectified features. 
All text features were then fed into a CTC-based recognizer, making the framework free from character-level annotations. 
Instead of formulating the text detection branch as a bounding box extraction or instance segmentation task, Wang et al. \cite{wang2019all} localized a set of points on the boundary and adopted TPS \cite{warps1989thin} transformation to flatten features of each text. 
Qiao et al. \cite{qiao2019text} proposed the shape transform module, which iteratively generated potential fiducial points and used TPS to transform the detected text regions into regular morphologies without extra parameters. 
Liu et al. \cite{Yuliang2020ABCNet} introduced parameterized Bezier curve to adaptively fit arbitrarily-shaped text and designed a novel BezierAlign layer to precisely calculate convolutional features of text instances in curved shapes. 
% for iteratively generating potential fiducial points
The purpose of the aforementioned bilinear interpolation methods is to rectify the features of irregular shapes into axis-aligned features for text recognizer, where the difference is the way of generating the sampling grid. 
However, Qin et al. \cite{qin2019towards} argued that the feature rectification was a key bottleneck in generalizing to irregular shaped text. 
They introduced RoI masking to filter out the neighboring text and the background, which made rectification unnecessary for the recognizer. 
Xing et al. \cite{xing2019convo} directly performed character detection and recognition on the full features without any RoI operations.

Although the current end-to-end systems work fairly well in many real-world scenarios, they contain limitations. 
The following difficulties should be considered:
i) How to efficiently bridge and share information between text detection and recognition? 
ii) How to balance the significant differences in learning difficulty and convergence speed between text detection and recognition?
iii) How to improve joint optimization?
Moreover, a simple, compact, and powerful end-to-end system is yet to be developed.

\section{Evaluations and Protocols}
\label{evaluations_and_protocols}

Diverse datasets and unified evaluation protocols bring new challenges and fair comparison to the community, respectively, but both are necessary to advance the field of STR.
In this section, we examine the standard benchmark datasets and evaluation protocols.
Table~\ref{Table_recognition_performance} and Table~\ref{Table_end_to_end_performance} compare the performance of the current advanced algorithms in STR and end-to-end systems.
% Table~\ref{Table_recognition_performance} and Table~\ref{Table_end_to_end_performance} report the approaches performance of scene text recognition and end-to-end system performance, respectively.
% Moreover, the details of these approaches have been introduced in the Section~\ref{methodologies}.

\subsection{Datasets}
\label{datasets}

Several primary reasons justify the need for additional datasets:
i) Most deep learning approaches are data-driven.
% , which rely on enough data.
Large-scale datasets are important and crucial to train a good text recognizer.
ii) Advanced STR algorithms have been overused on previous datasets, indicating that more challenging aspects could be investigated.
% Advanced STR algorithms have saturated on the previous datasets, indicating that researchers can tackle more challenging aspects.
% For example, the recognition algorithms proposed by Cheng et al. \cite{cheng2018aon} and Zhan et al. \cite{zhan2019esir} have achieved $99.6$\% accuracy rate on IIIT$5$K datasets in a lexicon-constrained manner.
iii) New datasets usually represent potential directions for future work, such as lexicon-free text recognition, irregular text recognition, unsupervised or weakly supervised text recognition, and large-scale category text recognition.
% New datasets not only contain numerous novel challenges, but also point out potential directions on scene text recognition in the future, such as lexicon-free text recognition, irregular text recognition and large-scale category text recognition.

\begin{figure}[t]
\centering
\includegraphics[width=0.45\textwidth]{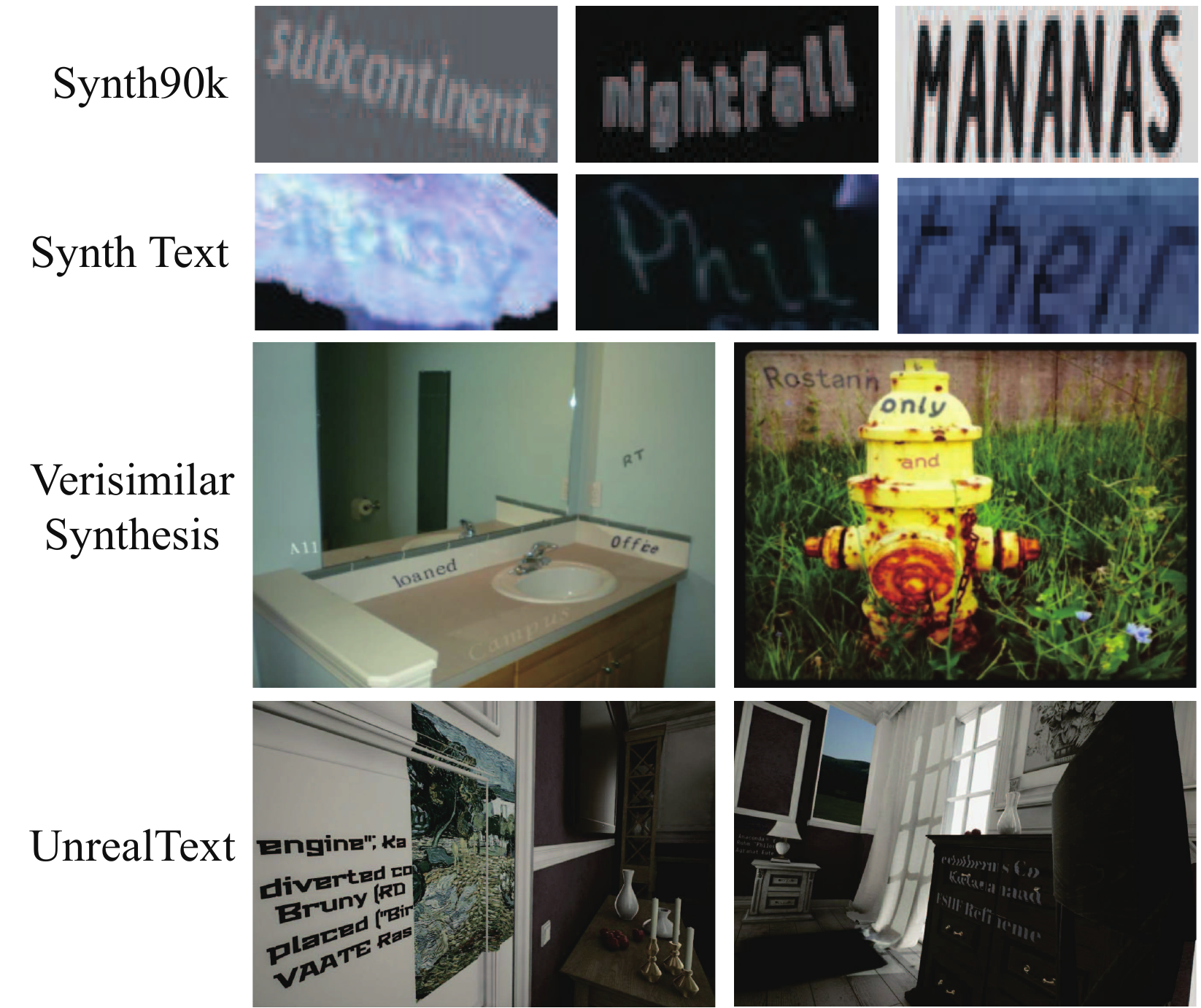}
\caption{Synthetic sample images of text from Synth$90$k, SynthText, Verisimilar Synthesis, and UnrealText datasets.
}
\label{Figure_Dataset_Synth}
\end{figure}

\begin{figure}[b]
\centering
\includegraphics[width=0.45\textwidth]{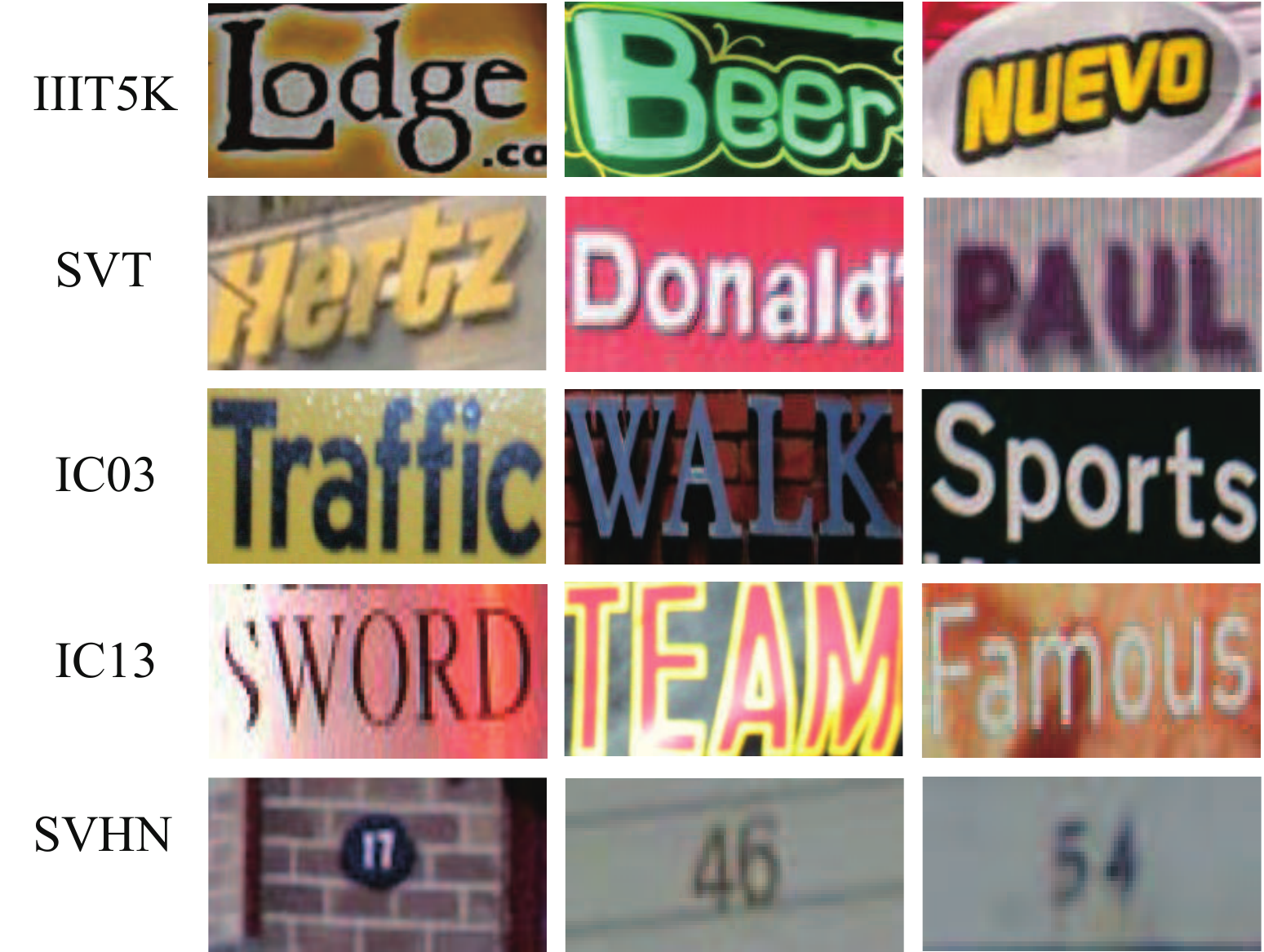}
\caption{Realistic sample images of regular Latin text from IIIT$5$K, SVT, IC$03$, IC$11$, IC$13$, and SVHN datasets.
}
\label{Figure_Dataset_Regular}
\end{figure}

Depending on the type of dataset collection, we divide the standard benchmark datasets into two categories: synthetic datasets and realistic datasets.
In particular, realistic datasets include regular Latin datasets, irregular Latin datasets and multilingual datasets.
Table~\ref{Table_dataset} describes the panorama of these datasets, and Figures~\ref{Figure_Dataset_Synth} – ~\ref{Figure_Dataset_multiLing} show representative samples.
% Moreover, some representative samples are demonstrated in Figure~\ref{Figure_Dataset_Synth}, Figure~\ref{Figure_Dataset_Regular}, Figure~\ref{Figure_Dataset_irRegular} and Figure~\ref{Figure_Dataset_multiLing}, respectively.

\subsubsection{Synthetic Datasets}
\label{synthetic_datasets}

Most deep learning algorithms rely on sufficient data.
However, the existing realistic datasets are relatively small for training a highly accurate scene text recognizer, because they only contain thousands of data samples.
Moreover, manually collecting and annotating large amount of real-world data will involve huge efforts and resources.
Therefore, synthetic and artificial data generation has been a popular research topic \cite{jaderberg2014synthetic}, \cite{gupta2016synthetic}, \cite{zhan2018verisimilar}, \cite{Long2020UnrealText}.

\begin{table*}[htbp]
\tiny
\caption{Comparison of the benchmark datasets.
`50', `1k', and `full' are the lexicon sizes.}
\label{Table_dataset}
\resizebox{\textwidth}{!}{
\begin{tabular}{cccccccccccc}
  \toprule
  \multirow{2}[1]{*}{\textbf{Datasets}} & \multirow{2}[1]{*}{\textbf{Language}} & \multicolumn{3}{c}{\textbf{Images}} & \multicolumn{3}{c}{\textbf{Instances}} & \multirow{2}[1]{*}{\textbf{Lexicon}} & \multirow{2}[1]{*}{\textbf{Char-Level Label}} & \multirow{2}[1]{*}{\textbf{Type}} & \multirow{2}[1]{*}{\textbf{Source Code}} \\
  \cmidrule(lr){3-5} \cmidrule(lr){6-8} &       & Total & Train & Test  & Total & Train & Test  &      &     &       &  \\
  \midrule
  % Synth90k \cite{jaderberg2014synthetic}\tablefootnote{\url{http://www.robots.ox.ac.uk/~vgg/data/text/}}  & English & $\sim$9000000 & -     & -     &  $\sim$9000000     & -     & -     &  \XSolidBrush    & {\XSolidBrush} & Regular & http://www.robots.ox.ac.uk/~vgg/data/text/ \\
  Synth90k \cite{jaderberg2014synthetic}  & English & $\sim$9000000 & -     & -     &  $\sim$9000000     & -     & -     &  \XSolidBrush    & {\XSolidBrush} & Regular & http://www.robots.ox.ac.uk/~vgg/data/text/ \\
  SynthText \cite{gupta2016synthetic} & English & $\sim$6000000 & -     & -     & $\sim$6000000     & -     & -     &  \XSolidBrush    & \Checkmark & Regular & https://github.com/ankush-me/SynthText \\
  Verisimilar Synthesis \cite{zhan2018verisimilar} & English & - & -     & -     & $\sim$5000000     & -     & -     &  \XSolidBrush    & {\XSolidBrush} & Regular &\tabincell{c}{https://github.com/fnzhan/Verisimilar-Image-Synthesis\\-for-Accurate-Detection-and-Recognition-of-Texts-in-Scenes} \\
  UnrealText \cite{Long2020UnrealText} & English & $\sim$600000 & -     & -     & $\sim$12000000     & -     & -     &  \XSolidBrush    & \Checkmark & Regular & https://jyouhou.github.io/UnrealText/ \\
  IIIT5K \cite{mishra2012scene} & English & 1120  & 380   & 740   & 5000  & 2000  & 3000  & {50 and 1k} & \Checkmark     & Regular & \tabincell{c}{http://cvit.iiit.ac.in/research/projects\\/cvit-projects/the-iiit-5k-word-dataset} \\
  SVT \cite{wang2011end}, \cite{wang2010word}  & English & 350   & 100   & 250   & 725   & 211   & 514   & 50    & {\XSolidBrush} & {Regular} & {http://vision.ucsd.edu/~kai/svt/} \\
  IC03 \cite{lucas2005icdar} & English & 509   & 258   & 251   & 2268  & 1157  & 1111  & {50, 1k and full} & \Checkmark     & Regular & \tabincell{c}{http://www.iapr-tc11.org/mediawiki/index.php?\\title=ICDAR\_2003\_Robust\_Reading\_Competitions} \\
  IC11 \cite{shahab2011icdar} & English & 522   & 420   & 102   & 4501  & 3583     & 918     &  \XSolidBrush    & \Checkmark     & Regular & http://www.cvc.uab.es/icdar2011competition/?com=downloads \\
  IC13 \cite{karatzas2013icdar} & English & 561   & 420   & 141   & 5003  & 3564  & 1439  &  \XSolidBrush    & \Checkmark     & Regular & \tabincell{c}{http://dagdata.cvc.uab.es/\\icdar2013competition/?ch=2\&com=downloads} \\
  SVHN \cite{netzer2011reading} & Digits & 600000 & 573968 & 26032 & 600000 & 573968 & 26032 &  \XSolidBrush    & \Checkmark     & Regular & http://ufldl.stanford.edu/housenumbers/ \\
  SVT-P \cite{quy2013recognizing} & English & 238   & 0     & 238   & 639   & 0     & 639   & {50 and full} & {\XSolidBrush} & {Irregular} & \tabincell{c}{https://pan.baidu.com/s/\\1rhYUn1mIo8OZQEGUZ9Nmrg (pw: vnis)} \\
  CUTE80 \cite{risnumawan2014robust} & English & 80    & 0     & 80    & 288   & 0     & 288   &  \XSolidBrush    & {\XSolidBrush} & Irregular & http://cs-chan.com/downloads\_CUTE80\_dataset.html \\
  IC15 \cite{karatzas2015icdar} & English & 1500  & 1000  & 500   & 6545     & 4468     & 2077  &  \XSolidBrush    & {\XSolidBrush} & Irregular & http://rrc.cvc.uab.es/?ch=4\&com=downloads \\
  COCO-Text \cite{veit2016coco} & English & 63686 & 43686 & 10000 & 145859 & 118309 & 27550 &  \XSolidBrush    & {\XSolidBrush} & Irregular & https://vision.cornell.edu/se3/coco-text-2/ \\
  Total-Text \cite{ch2017total} & English & 1555  & 1255  & 300   & 11459 & 11166     & 293     &  \XSolidBrush    & {\XSolidBrush} & Irregular & https://github.com/cs-chan/Total-Text-Dataset \\
  RCTW-17 \cite{shi2017icdar2017} & Chinese/English & 12514 & 11514 & 1000  & -     & -     & -     &  \XSolidBrush    & {\XSolidBrush} & Regular & http://rctw.vlrlab.net/dataset/ \\
  MTWI \cite{he2018icpr2018} & Chinese/English & 20000 & 10000 & 10000 & 290206     & 141476     & 148730     &  \XSolidBrush    & {\XSolidBrush} & {Regular} & \tabincell{c}{https://pan.baidu.com/s/1SUODaOzV7YOPkrun0xSz6A\\\#list/path=\%2F (pw:gox9)} \\
  CTW \cite{yuan2018chinese} & Chinese/English & 32285 & 25887 & 3269  & 1018402 & 812872 & 103519 &  \XSolidBrush    & \Checkmark     & Regular & https://ctwdataset.github.io/ \\
  SCUT-CTW1500 \cite{yuliang2017detecting} & Chinese/English & 1500  & 1000  & 500   & 10751 & 7683     & 3068     &  \XSolidBrush    & {\XSolidBrush} & Irregular & https://github.com/Yuliang-Liu/Curve-Text-Detector \\
  LSVT \cite{sun2019icdar}, \cite{sun2019chinese} & Chinese/English & 450000 & 30000 & 20000 & -     & -     & -     &  \XSolidBrush    & {\XSolidBrush} & {Irregular} & {https://rrc.cvc.uab.es/?ch=16\&com=downloads} \\
  ArT \cite{chng2019icdar2019} & Chinese/English & 10166 & 5603  & 4563  & 98455     & 50029     & 48426     &  \XSolidBrush    & {\XSolidBrush} & Irregular & https://rrc.cvc.uab.es/?ch=14\&com=downloads \\
  ReCTS-25k \cite{chng2019icdar2019} & Chinese/English & 25000 & 20000     & 5000     & 119713     & 108924     & 10789     &  \XSolidBrush    & \Checkmark     & Irregular & https://rrc.cvc.uab.es/?ch=12\&com=downloads \\
  MLT \cite{nayef2019icdar2019} & Multilingual & 20000 & 10000  & 10000  & 191639   & 89177     & 102462     &  \XSolidBrush    &  \XSolidBrush     & Irregular  & https://rrc.cvc.uab.es/?ch=15\&com=downloads \\
\bottomrule
\end{tabular}}
\end{table*}

\begin{itemize} 
\item \textbf{Synth90k.}
The Synth$90$k dataset \cite{jaderberg2014synthetic} contains $9$ million synthetic text instance images from a set of $90$k common English words.
Words are rendered onto natural images with random transformations and effects, such as random fonts, colors, blur, and noises.
Synth$90$k dataset can emulate the distribution of scene text images and can be used instead of real-world data to train data-hungry deep learning algorithms.
Besides, every image is annotated with a ground-truth word.

\item \textbf{SynthText.}
The SynthText dataset \cite{gupta2016synthetic} contains $800,000$ images with $6$ million synthetic text instances.
As in the generation of Synth$90$k dataset, the text sample is rendered using a randomly selected font and transformed according to the local surface orientation.
Moreover, each image is annotated with a ground-truth word.

\item \textbf{Verisimilar Synthesis.}
The Verisimilar Synthesis dataset \cite{zhan2018verisimilar} contains $5$ million synthetic text instance images.
% contains $10,000$ synthetic images with $5$ million cropped text instances.
% contains $5$ million synthetic text instance images.
Given background images and source texts, a semantic map and a saliency map are first determined which are then combined to identify semantically sensible and apt locations for text embedding.
The color, brightness, and orientation of the source texts are further determined adaptively according to the color, brightness, and contextual structures around the embedding locations within the background image. 

\item \textbf{UnrealText.}
The UnrealText dataset \cite{Long2020UnrealText} contains $600$K synthetic images with $12$ million cropped text instances.
It is developed upon Unreal Engine $4$ and the UnrealCV plugin \cite{Qiu2016UnrealCV}.
Text instances are regarded as planar polygon meshes with text foregrounds loaded as texture.
These meshes are placed in suitable positions in $3$D world, and rendered together with the scene as a whole.
The same font set from Google Fonts\footnote{\url{https://fonts.google.com/}} and the same text corpus, i.e., Newsgroup$20$, are used as SynthText does.
% The intensity, color and direction of all light sources in the scene are randomly changed.

\end{itemize}

\subsubsection{Realistic Datasets}
\label{realistic_datasets}

Most of current realistic datasets contain only thousands of text instance images.
Therefore, for STR, realistic datasets are typically used to evaluate recognition algorithms under real-world conditions.
Subsequently, we will list and briefly describe the existing realistic datasets: regular Latin datasets, irregular Latin datasets, and multilingual datasets.

\noindent\textbf{Regular Latin Datasets}
\label{regular_latin_datasets}

For the regular Latin datasets, most text instances are frontal and horizontal, whereas a small part of them is distorted.

\begin{itemize} 
\item \textbf{IIIT5K-Words (IIIT5K).}
The IIIT$5$K dataset \cite{mishra2012scene} contains $5,000$ text instance images: $2,000$ for training and $3,000$ for testing.
It contains words from street scenes and from originally-digital images.
Every image is associated with a $50$-word lexicon and a $1,000$-word lexicon.
Specifically, the lexicon consists of a ground-truth word and some randomly picked words.

\item \textbf{Street View Text (SVT).}
The SVT dataset \cite{wang2011end}, \cite{wang2010word} contains $350$ images: $100$ for training and $250$ for testing.
 % with $647$ cropped text instances collected from Google Street View.
Some images are severely corrupted by noise, blur, and low resolution.
Each image is associated with a $50$-word lexicon.

\item \textbf{ICDAR 2003 (IC03).}
The IC$03$ dataset \cite{lucas2005icdar} contains $509$ images: $258$ for training and $251$ for testing.
Specifically, it contains $867$ cropped text instances after discarding images that contain non-alphanumeric characters or less than three characters.
Every image is associated with a $50$-word lexicon and a full-word lexicon.
Moreover, the full lexicon combines all lexicon words.

\item \textbf{ICDAR 2011 (IC11).}
The IC$11$ dataset \cite{shahab2011icdar} contains $485$ images.
This is an extension of the dataset used for the text locating competitions of ICDAR $2003$.

\item \textbf{ICDAR 2013 (IC13).}
The IC$13$ dataset \cite{karatzas2013icdar} contains $561$ images: $420$ for training and $141$ for testing.
% $1,015$ cropped text instance images.
It inherits data from the IC$03$ dataset and extends it with new images.
Similar to IC$03$ dataset, the IC$13$ dataset contains $1,015$ cropped text instance images after removing the words with non-alphanumeric characters. 
No lexicon is associated with IC$13$. 
Notably, $215$ duplicate text instance images \cite{baek2019wrong} exist between the IC$03$ training dataset and the IC$13$ testing dataset.
Therefore, care should be taken regarding the overlapping data when evaluating a model on the IC$13$ testing data.

\item \textbf{Street View House Number (SVHN).}
The SVHN dataset \cite{netzer2011reading} contains more than $600,000$ digits of house numbers in natural scenes.
It is obtained from a large number of street view images using a combination of automated algorithms and the Amazon Mechanical Turk (AMT) framework\footnote{\url{https://www.mturk.com/mturk/welcome}}. 
The SVHN dataset was typically used for scene digit recognition.
\end{itemize} 

\noindent\textbf{Irregular Latin Datasets}
\label{irregular_latin_datasets}

For the irregular benchmark datasets, most of the text instances are low-resolution, perspective distorted, or curved.
Various fonts and distorted patterns of irregular text bring additional challenges in STR.

\begin{itemize} 
\item \textbf{StreetViewText-Perspective (SVT-P).}
The SVT-P dataset \cite{quy2013recognizing} contains $238$ images with $639$ cropped text instances.
It is specifically designed to evaluate perspective distorted text recognition.
It is built based on the original SVT dataset by selecting the images at the same address on Google Street View but with different view angles.   
Therefore, most text instances are heavily distorted by the non-frontal view angle.
Moreover, each image is associated with a $50$-word lexicon and a full-word lexicon.

\item \textbf{CUTE80 (CUTE).}
The CUTE dataset \cite{risnumawan2014robust} contains $80$ high-resolution images with $288$ cropped text instances.
It focuses on curved text recognition.
Most images in CUTE have a complex background, perspective distortion, and poor resolution.
No lexicon is associated with CUTE.

\begin{figure}[t]
\centering
\includegraphics[width=0.45\textwidth]{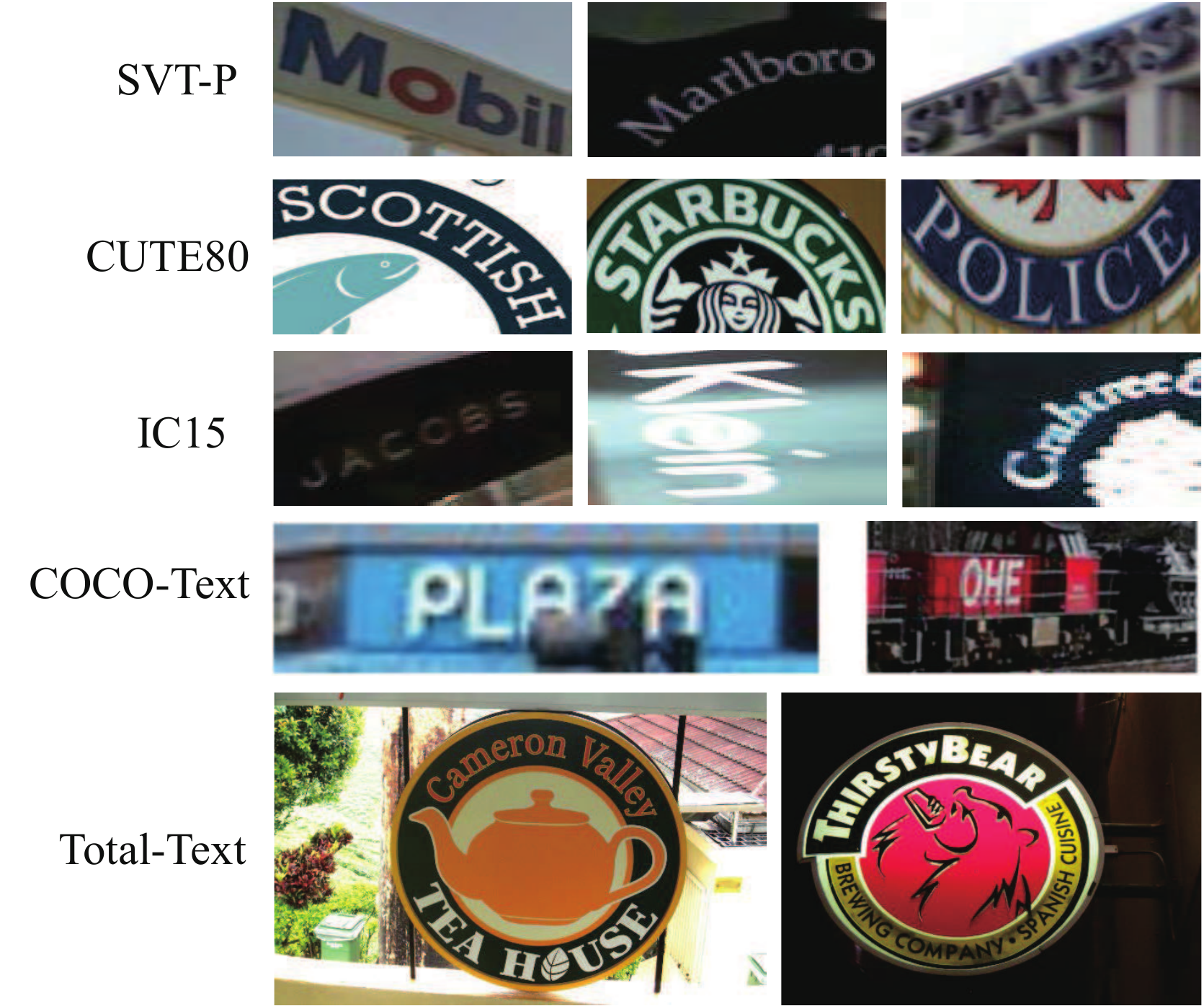}
\caption{Realistic sample images of irregular Latin text from SVT-P, CUTE$80$, IC$15$, COCO-Text and Total-Text datasets.
}
\label{Figure_Dataset_irRegular}
\end{figure}

\item \textbf{ICDAR 2015 (IC15).}
The IC$15$ dataset \cite{karatzas2015icdar} contains $1,500$ images: $1,000$ for training and $500$ for testing.
Specifically, it contains $2,077$ cropped text instances, including more than $200$ irregular text samples.
As text images were taken by Google Glasses without ensuring the image quality, most of the text is very small, blurred, and multi-oriented.
No lexicon is provided.

\item \textbf{COCO-Text.}
The COCO-Text dataset \cite{veit2016coco} contains $63,686$ images with $145,859$ cropped text instances.
It is the first large-scale dataset for text in natural images and also the first dataset to annotate scene text with attributes such as legibility and type of text.
However, no lexicon is associated with COCO-Text. 

\item \textbf{Total-Text.}
The Total-Text \cite{ch2017total} contains $1,555$ images with $11,459$ cropped text instance images.
It focuses on curved scene text recognition.
Images in Total-Text have more than three different orientations, including horizontal, multi-oriented, and curved.
No lexicon is associated with Total-Text.
\end{itemize}

\noindent\textbf{Multilingual Datasets}
\label{Multilingual_datasets}

Multilingual text can be found in modern cities, where representatives of multiple cultures live and communicate.
Bilingual datasets are the simplest form.
Subsequently, some bilingual or multilingual scene text datasets are introduced below.
The bilingual datasets introduced in this paper are mainly composed of Latin and Chinese.

The reason for choosing Chinese as the second language of bilingual scene text datasets are three-fold.
First, Chinese is one of the most widely used languages in the world.
% Therefore, Chinese text reading has great potential practical value.
Second, although many STR algorithms exist, most of them focus on Latin characters.
The problem of recognition of Chinese scene text has not been solved well.
Third, Chinese text has unique characteristics compared with Latin text:
i) Chinese is a large-scale category text, with a much larger character set than in Latin text.
ii) The imbalanced class problem of Chinese characters is more obvious owing to the larger character set.
iii) Many confusing characters with similar structures exist in Chinese, which makes them hard to distinguish.
Therefore, reading Chinese in the wild is an important and challenging problem.

\begin{itemize} 
\item \textbf{Reading Chinese Text in the Wild (RCTW-17).}
The RCTW-$17$ dataset \cite{shi2017icdar2017} contains $12,514$ images: $11,514$ for training and $1,000$ for testing.
Most are natural images collected by cameras or mobile phones, whereas others are digital-born.
Text instances are annotated with labels, fonts, languages, etc.

\item \textbf{Multi-Type Web Images (MTWI).}
The MTWI dataset \cite{he2018icpr2018} contains $20,000$ images.
This is the first dataset constructed by Chinese and Latin web text.
Most images in MTWI have a relatively high resolution and cover diverse types of web text, including multi-oriented text, tightly-stacked text, and complex-shaped text.

\item \textbf{Chinese Text in the Wild (CTW).}
The CTW dataset \cite{yuan2018chinese} includes $32,285$ high-resolution street view images with $1,018,402$ character instances.
% , which is the largest publicly available dataset for Chinese text in natural images.
All images have character-level annotations: the underlying character, the bounding box, and six other attributes.

\item \textbf{SCUT-CTW1500.}
The SCUT-CTW$1500$ dataset \cite{yuliang2017detecting} contains $1,500$ images: $1,000$ for training and $500$ for testing.
In particular, it provides $10,751$ cropped text instance images, including $3,530$ with curved text.
The images are manually harvested from the Internet, image libraries such as Google Open-Image \cite{krasin2017openimages}, or phone cameras.
The dataset contains a lot of horizontal and multi-oriented text.

% \item \textbf{Chinese Street View Text (C-SVT).}

% The C-SVT dataset \cite{sun2019chinese} contains more than $430,000$ street view images in total, including $30,000$ fully annotated images with locations and text labels for the regions and $400,000$ more images in which only the annotations of text-of-interest are given.
% It is the largest one compared with existing Chinese text reading datasets. 

\begin{figure}[t]
\centering
\includegraphics[width=0.4\textwidth]{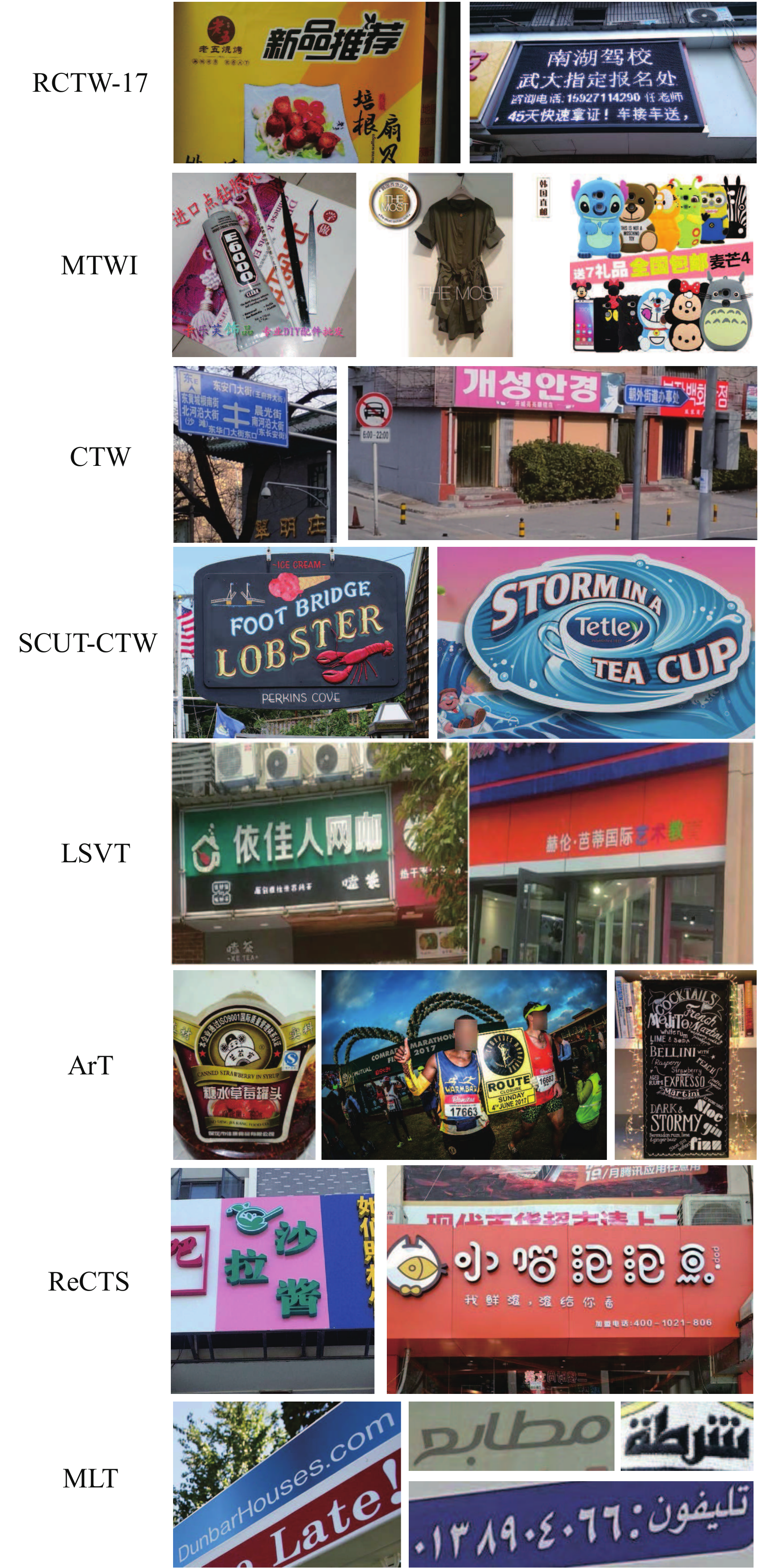}
\caption{Realistic sample images of multilingual scene text from RCTW-$17$, MTWI, CTW, SCUT-CTW$1500$, LSVT, ArT, ReCTS-$25$k, and MLT datasets.
}
\label{Figure_Dataset_multiLing}
\end{figure}

\item \textbf{Large-Scale Street View Text (LSVT).}
The LSVT dataset \cite{sun2019icdar}, \cite{sun2019chinese} contains $20,000$ testing samples, $30,000$ fully annotated training samples, and $400,000$ training samples with weak annotations (i.e., with partial labels).
All images are captured from streets and reflect a large variety of complicated real-world scenarios, e.g., store fronts and landmarks.

\item \textbf{Arbitrary-Shaped Text (ArT).}
The ArT dataset \cite{chng2019icdar2019} contains $10,166$ images: $5,603$ for training and $4,563$ for testing.
ArT is a combination of Total-Text, SCUT-CTW$1500$, and Baidu Curved Scene Text\footnote{A subset of LSVT}, which was collected to introduce the arbitrary-shaped text problem.
Moreover, all existing text shapes (i.e., horizontal, multi-oriented, and curved) have multiple occurrences in the ArT dataset.

\item \textbf{Reading Chinese Text on Signboard (ReCTS-25k).}
The ReCTS-$25$k dataset\cite{liu2019icdar} contains $25,000$ images: $20,000$ for training and $5,000$ for testing.
All the text lines and characters are annotated with locations and transcriptions.
All the images are from the Meituan-Dianping Group, collected by Meituan business merchants, using phone cameras under uncontrolled conditions.
Specifically, ReCTS-$25$k dataset mainly contains images of Chinese text on signboards.

\item \textbf{Multi-lingual Text (MLT-2019).}
The MLT-$2019$ dataset \cite{nayef2019icdar2019} contains $20,000$ images: $10,000$ for training ($1,000$ per language) and $10,000$ for testing.
The dataset includes ten languages, representing seven different scripts: Arabic, Bangla, Chinese, Devanagari, English, French, German, Italian, Japanese, and Korean.
The number of images per script is equal.

\end{itemize} 

\subsection{Evaluation Protocols}
\label{evaluation_protocols}

In this section, we summarize the evaluation protocols for Latin text and multilingual text.

\subsubsection{Evaluation Protocols for Latin Text}
\label{evaluation_protocols_for_latin_text}

\textbf{Recognition Protocols}

% $WRA$ and $WER$ are two widely used recognition evaluation protocols for Latin text.
The word recognition accuracy ($WRA$) and word error rate ($WER$) are two widely used recognition evaluation protocols for Latin text.
% the character-level recognition rate ($CR$ and $AR$) and word-level recognition rate ($WRA$ and $WER$).
% The evaluation for STR consists of the character-level recognition rate ($CR$ and $AR$) and word-level recognition rate ($WRA$ and $WER$).

\begin{itemize} 
% \item \textbf{CR.}
% $CR$ means the character correct rate, computed by,
% \begin{equation}
% CR =\frac{N-D_e-S_e}{N},
% \end{equation}
% where $N$ is the total number of characters in the ground-truth text lines.
% $D_e$ and $S_e$ represent deletion errors and substitution errors, respectively.

% \item \textbf{AR.}
% $AR$ indicates the accurate rate, expressed as follows,
% \begin{equation}
% AR =\frac{N-D_e-S_e-I_e}{N},
% \end{equation}
% where $N$, $D_e$ and $S_e$ have the same meanings as in $CR$, and $I_e$ represents insertion errors.

\item \textbf{WRA.}
$WRA$ is defined by
\begin{equation}
WRA =\frac{W_r}{W},
\end{equation}
where $W$ is the total number of words, and $W_r$ represents the number of correctly recognized words. 

\item \textbf{WER.}
$WER$ is defined by
\begin{equation}
WER = 1 - WRA = 1 -\frac{W_r}{W}.
\end{equation}

\end{itemize}
% \subsubsection{Evaluation Protocols for An End-to-End System}
% \label{evaluation_protocols_for_an_end_to_end_system}

\noindent\textbf{End-to-End Protocols}

Widely used evaluation protocols for Latin end-to-end systems\footnote{\url{https://rrc.cvc.uab.es/files/Robust_Reading_2015_v02.pdf}} are defined in \cite{karatzas2013icdar}, \cite{karatzas2015icdar}, where the recognition algorithms are evaluated in two modalities: \textit{end-to-end recognition} and \textit{word spotting}.
In particular, all words in the scene text images should be detected and recognized under \textit{end-to-end recognition}.
Under \textit{word spotting}, only words provided in the vocabulary should be detected and recognized.
Moreover, three different vocabularies are provided for candidate transcriptions: \textit{strongly contextualised}, \textit{weakly contextualised} and \textit{generic} (denoted as \textbf{S}, \textbf{W}, and \textbf{G} in short, respectively).

\begin{itemize} 
\item \textbf{Strongly Contextualised (S).}
The per-image vocabulary consists of $100$ words, including all words in the corresponding image as well as distractors selected from the rest of the training/testing set, which follows the setup of \cite{wang2011end}.

\item \textbf{Weakly Contextualised (W).}
The vocabulary includes all words in the training/testing set.

\item \textbf{Generic (G).}
The generic vocabulary contains approximately $90$K words derived from the dataset\footnote{Available at:\url{http://www.robots.ox.ac.uk/~vggldataltext/ }} of Jaderberg et al. \cite{jaderberg2016reading}.
\end{itemize}

\begin{table*}[th]
\tiny
\centering
\caption{Performance comparison of recognition algorithms on benchmark datasets.
`50', `1k', and `Full' are lexicon sizes. `None' means lexicon-free.
`*' indicates the methods that use the extra datasets other than Synth$90$k and SynthText. The \textbf{bold} represents the best recognition results. `$^\dagger$' denotes the best recognition performance of using extra datasets.}
\label{Table_recognition_performance}%
\resizebox{\textwidth}{!}{%
\begin{tabular}{ccccccccccccccccc}
  \toprule
  \multirow{2}[2]{*}{\textbf{Method}} & \multicolumn{3}{c}{\textbf{IIIT5K}} & \multicolumn{2}{c}{\textbf{SVT}} & \multicolumn{4}{c}{\textbf{IC03}}     & \textbf{IC13}  & \multicolumn{3}{c}{\textbf{SVT-P}} & \textbf{CUTE80} & \textbf{IC15}  & \textbf{COCO-TEXT} \\
  \cmidrule(lr){2-4} \cmidrule(lr){5-6} \cmidrule(lr){7-10} \cmidrule(lr){11-11} \cmidrule(lr){12-14} \cmidrule(lr){15-15} \cmidrule(lr){16-16} \cmidrule(lr){17-17}         & 50    & 1K    & None  & 50    & None  & 50    & Full  & 50k   & None  & None  & 50    & Full  & None  & None  & None  & None \\
% \cmidrule{2-17}          & 50    & 1K    & None  & 50    & None  & 50    & Full  & 50k   & None  & None  & 50    & Full  & None  & None  & None  & None \\
  \midrule
  Wang et al. \cite{wang2011end} : ABBYY & 24.3  & -     & -     & 35.0  & -     & 56.0  & 55.0  & -     & -     & -     & 40.5  & 26.1  & -     & -     & -     & - \\
  Wang et al. \cite{wang2011end} : SYNTH+PLEX & -     & -     & -     & 57.0  & -     & 76.0  & 62.0  & -     & -     & -     & -     & -     & -     & -     & -     & - \\
  Mishra et al. \cite{mishra2012scene} & 64.1  & 57.5  & -     & 73.2  & -     & 81.8  & 67.8  & -     & -     & -     & 45.7  & 24.7  & -     & -     & -     & - \\
  Wang et al. \cite{wang2012end} & -     & -     & -     & 70.0  & -     & 90.0  & 84.0  & -     & -     & -     & 40.2  & 32.4  & -     & -     & -     & - \\
  Goel et al. \cite{goel2013whole} : wDTW & -     & -     & -     & 77.3  & -     & 89.7  & -     & -     & -     & -     & -     & -     & -     & -     & -     & - \\
  Bissacco et al. \cite{bissacco2013photoocr} : PhotoOCR & -     & -     & -     & 90.4  & 78.0  & -     & -     & -     & -     & 87.6  & -     & -     & -     & -     & -     & - \\
  Phan et al. \cite{quy2013recognizing} & -     & -     & -     & 73.7  & -     & 82.2  & -     & -     & -     & -     & 62.3  & 42.2  & -     & -     & -     & - \\
  Alsharif et al. \cite{Alsharif2014end} : HMM/Maxout & -     & -     & -     & 74.3  & -     & 93.1  & 88.6  & 85.1  & -     & -     & -     & -     & -     & -     & -     & - \\
  Almaz\'{a}n et al \cite{almazan2014word} : KCSR & 88.6  & 75.6  & -     & 87.0  & -     & -     & -     & -     & -     & -     & -     & -     & -     & -     & -     & - \\
  Yao et al. \cite{yao2014strokelets} : Strokelets & 80.2  & 69.3  & -     & 75.9  & -     & 88.5  & 80.3  & -     & -     & -     & -     & -     & -     & -     & -     & - \\
  R.-Serrano et al.\cite{rodriguez2015label} : Label embedding & 76.1  & 57.4  & -     & 70.0  & -     & -     & -     & -     & -     & -     & -     & -     & -     & -     & -     & - \\
  Jaderberg et al. \cite{jaderberg2014deep} & -     & -     & -     & 86.1  & -     & 96.2  & 91.5  & -     & -     & -     & -     & -     & -     & -     & -     & - \\
  Su and Lu \cite{su2014accurate} & -     & -     & -     & 83.0  & -     & 92.0  & 82.0  & -     & -     & -     & -     & -     & -     & -     & -     & - \\
  Gordo \cite{gordo2015supervised} : Mid-features & 93.3  & 86.6  & -     & 91.8  & -     & -     & -     & -     & -     & -     & -     & -     & -     & -     & -     & - \\
  Jaderberg et al. \cite{jaderberg2016reading} & 97.1  & 92.7  & -     & 95.4  & 80.7  & 98.7  & 98.6 & 93.3  & 93.1  & 90.8  & -     & -     & -     & -     & -     & - \\
  Jaderberg et al. \cite{jaderberg2015deep} & 95.5  & 89.6  & -     & 93.2  & 71.7  & 97.8  & 97.0  & 93.4  & 89.6  & 81.8  & -     & -     & -     & -     & -     & - \\
  Shi, Bai, and Yao \cite{shi2017end} : CRNN & 97.8  & 95.0  & 81.2  & 97.5  & 82.7  & 98.7  & 98.0  & \textbf{95.7} & 91.9  & 89.6  & -     & -     & -     & -     & -     & - \\
  Shi et al. \cite{shi2016robust} : RARE & 96.2  & 93.8  & 81.9  & 95.5  & 81.9  & 98.3  & 96.2  & 94.8  & 90.1  & 88.6  & 91.2  & 77.4  & 71.8  & 59.2  & -     & - \\
  Lee and Osindero \cite{lee2016recursive} : R2AM & 96.8  & 94.4  & 78.4  & 96.3  & 80.7  & 97.9  & 97.0  & -     & 88.7  & 90.0  & -     & -     & -     & -     & -     & - \\
  Liu et al.  \cite{liu2016star} : STAR-Net & 97.7  & 94.5  & 83.3  & 95.5  & 83.6  & 96.9  & 95.3  & -     & 89.9  & 89.1  & 94.3 & 83.6  & 73.5  & -     & -     & - \\
  *Liu et al. \cite{liu2016scene} & 94.1  & 84.7  & -     & 92.5  & -     & 96.8  & 92.2  & -     & -     & -     & - & -     & -     & -     & -     & - \\
  *Mishra et al. \cite{mishra2016enhancing} & 78.1  & -     & 46.7  & 78.2  & -     & 88.0  & -     & -     & 67.7  & 60.2  & - & -     & -     & -     & -     & - \\
  *Su and Lu \cite{su2017accurate} & -     & -     & -     & 91.0  & -     & 95.0  & 89.0  & -     & -     & 76.0  & - & -     & -     & -     & -     & - \\
  *Yang et al. \cite{yang2017learning} & 97.8  & 96.1  & -     & 95.2  & -     & 97.7  & -     & -     & -     & -     & 93.0  & 80.2  & 75.8  & 69.3  & -     & - \\
  Yin et al. \cite{yin2017scene} & 98.7  & 96.1  & 78.2  & 95.1  & 72.5  & 97.6  & 96.5  & -     & 81.1  & 81.4  & -     & -     & -     & -     & -     & - \\
  *Cheng et al. \cite{cheng2017focusing} : FAN & 99.3  & 97.5  & 87.4  & 97.1  & 85.9  & \textbf{$^\dagger$99.2} & 97.3  & -     & 94.2  & 93.3  & -     & -     & -     & -     & \textbf{$^\dagger$85.3} & - \\
  Cheng et al. \cite{cheng2018aon} : AON & \textbf{99.6} & 98.1  & 87.0  & 96.0  & 82.8  & 98.5  & 97.1  & -     & 91.5  & -     & 94.0  & 83.7  & 73.0  & 76.8  & 68.2  & - \\
  Liu et al.  \cite{liu2018char} : Char-Net & -     & -     & 83.6  & -     & 84.4  & -     & 93.3  & -     & 91.5  & 90.8  & -     & -     & 73.5  & -     & 60.0  & - \\
  *Liu et al.  \cite{liu2018squeezedtext} : SqueezedText & 97.0  & 94.1  & 87.0  & 95.2  & -     & 98.8 & 97.9  & 93.8  & 93.1  & 92.9  & -     & -     & -     & -     & -     & - \\
  *Zhan et al. \cite{zhan2018verisimilar} & 98.1  & 95.3  & 79.3  & 96.7  & 81.5  & - & -     & -     & -     & 87.1  & -     & -     & -     & -     & -     & - \\
  *Bai et al. \cite{bai2018edit} : EP & 99.5  & 97.9  & 88.3  & 96.6  & 87.5  & 98.7  & 97.9  & -     & 94.6  & 94.4 & -     & -     & -     & -     & 73.9  & - \\
  Fang et al. \cite{fang2018attention} & 98.5  & 96.8  & 86.7  & 97.8  & 86.7  & \textbf{99.3}  & 98.4  & -     & 94.8  & 93.5 & -     & -     & -     & -     & 71.2  & - \\
  Liu et al. \cite{liu2018connectionist} : EnEsCTC & -     & -     & 82.0  & -     & 80.6  & -     & -     & -     & 92.0  & 90.6 & -     & -     & -     & -     & -     & - \\
  Liu et al. \cite{liu2018synthetically} & 97.3  & 96.1  & 89.4  & 96.8  & 87.1  & 98.1  & 97.5  & -     & 94.7  & 94.0 & -     & -     & 73.9  & 62.5  & -     & - \\
  Wang et al. \cite{wang2018memory} : MAAN & 98.3  & 96.4  & 84.1  & 96.4  & 83.5  & 97.4  & 96.4  & -     & 92.2  & 91.1 & -     & -     & -     & -     & -     & - \\
  Sheng et al. \cite{sheng2018nrtr} : NRTR  & 99.2  & 98.8  & 86.5  & 98.0  & 88.3  & 98.9  & 97.9  & -     & 95.4  & 94.7  & -     & -     & -     & -     & -     & - \\
  Gao et al. \cite{gao2018dense} & 99.1  & 97.2  & 83.6  & 97.7  & 83.9  & 98.6  & 96.6  & -     & 91.4  & 89.5  & -     & -     & -     & -     & -     & - \\
  Shi et al. \cite{shi2018aster} : ASTER & \textbf{99.6} & \textbf{98.8} & 93.4 & 97.4  & 89.5  & {98.8} & 98.0  & -     & 94.5  & 91.8  & -     & -     & 78.5  & 79.5  & 76.1  & - \\
  Luo et al. \cite{cluo2019moran} : MORAN & 97.9  & 96.2  & 91.2  & 96.6  & 88.3  & 98.7  & 97.8  & -     & 95.0 & 92.4  & 94.3 & 86.7 & 76.1  & 77.4  & 68.8  & - \\
  Luo et al. : MORAN-v2 & -     & -     & 93.4 & -     & 88.3  & -     & -     & -     & 94.2  & 93.2  & -     & -     & {79.7} & 81.9  & 73.9  & - \\
  % Luo et al.\tablefootnote{Available at \url{https://github.com/Canjie-Luo/MORAN_v2}} : MORAN-v2 & -     & -     & 93.4 & -     & 88.3  & -     & -     & -     & 94.2  & 93.2  & -     & -     & {79.7} & 81.9  & 73.9  & - \\
  Chen et al. \cite{chen2019adaptive} & 99.5  & 98.7  & {94.6} & 97.4  & 90.4  & 98.8  & 98.3  & -     & {95.3}  & 95.3  & {94.7}  & {89.6}  & {82.8} & 81.3  & 77.4  & - \\
  Xie et al. \cite{xie2019convolutional} : CAN & 97.0  & 94.2  & 80.5  & 96.9  & 83.4  & 98.4  & 97.8  & -     & 91.0  & 90.5  & -     & -     & -     & -     & -     & - \\
  *Liao et al. \cite{liao2019scene} : CA-FCN & \textbf{$^\dagger$99.8} & {98.9} & 92.0  & {98.8} & 82.1  & -     & -     & -     & -     & 91.4  & -     & -     & -     & 78.1  & -     & - \\
  *Li et al. \cite{li2019show} : SAR & 99.4  & 98.2  & {95.0} & {98.5} & {91.2} & -     & -     & -     & -     & {94.0} & \textbf{$^\dagger$95.8} & {91.2} & 86.4 & 89.6 & {78.8} & \textbf{$^\dagger$66.8} \\
  Zhan el at. \cite{zhan2019esir} : ESIR & \textbf{99.6} & \textbf{98.8} & 93.3  & 97.4  & {90.2} & -     & -     & -     & -     & 91.3  & -     & -     & 79.6  & {83.3} & {76.9} & - \\
  Zhang et al. \cite{zhang2019sequence} : SSDAN & -     & -     & 83.8  & -     & 84.5  & -     & -     & -     & 92.1  & 91.8  & -     & -     & -     & -     & -     & - \\
  *Yang et al. \cite{yang2019symmetry}: ScRN & 99.5  & 98.8  & 94.4  & 97.2  & 88.9  & 99.0  & 98.3  & -     & 95.0  & 93.9  & -     & -     & 80.8  & 87.5  & 78.7  & - \\
  *Yang et al. \cite{wang2019simple} & -  & -  & 94.2  & -  & 89.0  & -  & -  & -     & -  & 92.0  & 95.7     & 90.1     & 81.7  & 83.7  & 74.8  & - \\
  Wang et al. \cite{wang2019scene}: GCAM & -     & -     & 93.9  & -     & {91.3} & -     & -     & -     & {95.3}  & \textbf{95.7}  & -     & -     & \textbf{85.7} & 83.3  & {83.5} & - \\
  Jeonghun et al. \cite{baek2019wrong} & -     & -     & 87.9  & -     & 87.5  & -     & -     & -     & 94.4  & 92.3  & -     & -     & 79.2  & 74.0  & 71.8  & - \\
  Huang et al. \cite{huang2019epan}: EPAN & 98.9  & 97.8  & 94.0  & 96.6  & 88.9  & 98.7  & 98.0  & -     & 95.0  & 94.5  & 91.2  & 86.4  & 79.4  & 82.6  & 73.9  & - \\
  Gao et al. \cite{gao2019reading} & 99.1  & 97.9  & 81.8  & 97.4  & 82.7  & 98.7  & 96.7  & -     & 89.2  & 88.0  & -     & -     & -     & -     & 62.3  & \textbf{40.0} \\
  *Qi et al. \cite{qi2019novel} : CCL & 99.6  & 99.1  & 91.1  & 98.0  & 85.9  & 99.2  & \textbf{$^\dagger$98.8} & -     & 93.5  & 92.8  & -     & -     & -     & -     & 72.9  & - \\
  *Wang et al. \cite{wang2019reelfa} : ReELFA & 99.2  & 98.1  & 90.9  & -     & 82.7  & -     & -     & -     & -     & -     & -     & -     & -     & 82.3  & 68.5  & - \\
  *Zhu et al. \cite{zhu2019text} : HATN & -     & -     & 88.6  & -     & 82.2  & -     & -     & -     & 91.3  & 91.1  & -     & -     & 73.5  & 75.7  & 70.1  & - \\
  *Zhan et al. \cite{zhan2019spatial} : SF-GAN & -     & -     & 63.0  & -     & 69.3  & -     & -     & -     & -     & 61.8  & -     & -     & 48.6  & 40.6  & 39.0  & - \\
  Liao et al. \cite{liao2019mask} : SAM & 99.4  & 98.6  & 93.9  & {98.6} & 90.6  & 98.8  & 98.0  & -     & 95.2  & 95.3  & -     & -     & 82.2  & {87.8} & 77.3  & - \\
  *Liao et al. \cite{liao2019mask} : seg-SAM & \textbf{$^\dagger$99.8} & \textbf{$^\dagger$99.3} & 95.3 & \textbf{$^\dagger$99.1} & 91.8 & 99.0  & 97.9  & -     & 95.0  & 95.3  & -     & -     & 83.6  & 88.5  & 78.2  & - \\
  Wang et al. \cite{wang2019decoupled} : DAN & -     & -     & 94.3  & -     & 89.2  & -     & -     & -     & 95.0  & 93.9  & -     & -     & 80.0  & 84.4  & 74.5  & - \\
  Wang et al. \cite{wang2019textsr} & -     & -     & 92.5  & 98.0     & 87.2  & -     & -     & -     & 93.2  & 91.3  & -     & -     & 77.4  & 78.9  & 75.6  & - \\
  *Wan et al. \cite{wan2019textscanner} : TextScanner & 99.7     & 99.1     & 93.9  & 98.5     & 90.1  & -     & -     & -     & -  & 92.9  & -     & -     & 84.3  & 83.3  & 79.4  & - \\
  *Hu et al. \cite{hu2020gtc} : GTC & -     & -     & \textbf{$^\dagger$95.8}  & -     & \textbf{$^\dagger$92.9}  & -     & -     & -     & 95.5  & 94.4  & -     & -     & 85.7  & \textbf{$^\dagger$92.2}  & 79.5  & - \\
  Luo et al. \cite{Luo2020Separating} & \textbf{99.6}     & 98.7     & \textbf{95.4}  & \textbf{98.9}     & \textbf{92.7}  & 99.1     & \textbf{98.8}     & -     & \textbf{96.3}  & 94.8  & \textbf{95.5}     & \textbf{92.2}     & 85.4  & \textbf{89.6}  & \textbf{83.7}  & - \\
  *Litman et al. \cite{Ron2020Scatter} & -     & -     & 93.7  & -     & 92.7  & -     & -     & -     & \textbf{$^\dagger$96.3}  & 93.9  & -    & -     & \textbf{$^\dagger$86.9}  & 87.5  & 82.2  & - \\
  Yu et al. \cite{Deli2020Towards} & -     & -     & 94.8  & -     & 91.5  & -     & -     & -     & -  & 95.5  & -    & -     & 85.1  & 87.8  & 82.7  & - \\
  Qiao et al. \cite{qiao2020seed} & -     & -     & 93.8  & -     & 89.6  & -     & -     & -     & -  & 92.8  & -    & -     & 81.4  & 83.6  & 80.0  & - \\
  \bottomrule
\end{tabular}}
\end{table*}%

\begin{table*}[htbp]
\tiny
\centering
\caption{Performance comparison of end-to-end system algorithms on benchmark datasets.
`50' and `Full' are lexicon sizes. `None' means lexicon-free. 
`S', `W', and `G' stand for three different vocabularies, i.e., strongly contextualised, weakly contextualised, and generic. `*' represents testing with multiple scales. 
The \textbf{bold} represents the best results.
}
\label{Table_end_to_end_performance}%
\resizebox{\textwidth}{!}{%
\begin{tabular}{ccccccccccccccccccccc}
  \toprule
  \multirow{3}[3]{*}{\textbf{Method}} & \multicolumn{2}{c}{\multirow{2}[-2]{*}{\textbf{SVT}}} & \multicolumn{3}{c}{\multirow{2}[-2]{*}{\textbf{IC03}}} & \multirow{3}[-6]{*}{\textbf{IC11}} & \multicolumn{6}{c}{\textbf{IC13}}                     & \multicolumn{6}{c}{\textbf{IC15}}                     & \multicolumn{2}{c}{\textbf{Total-text}} \\
  \cmidrule(lr){8-13} \cmidrule(lr){14-19} & \multicolumn{2}{c}{} & \multicolumn{3}{c}{} &       & \multicolumn{3}{c}{End-to-end} & \multicolumn{3}{c}{Spotting} & \multicolumn{3}{c}{End-to-end} & \multicolumn{3}{c}{Spotting} &  & \\
  \cmidrule(lr){2-3} \cmidrule(lr){4-6} \cmidrule(lr){8-10} \cmidrule(lr){11-13} \cmidrule(lr){14-16} \cmidrule(lr){17-19} \cmidrule(lr){20-21}        & 50    & None  & 50    & Full  & None  &       & S     & W     & \multicolumn{1}{c}{G} & S     & W     & G     & S     & W     & \multicolumn{1}{c}{G} & S     & W     & G     & Full  & None  \\
  \midrule
  Wang et al. \cite{wang2011end} & 38.0     & -     & 68.0      & 51.0    & -     & -     & -     & -     & -     & -     & -     & -     & -     & -     & -     & -     & -     & -     & -     & - \\
  Wang et al. \cite{wang2012end} & 46.0    & -     & 72.0    & 67.0    & -     & -     & -     & -     & -     & -     & -     & -     & -     & -     & -     & -     & -     & -     & -     & - \\
  Jaderberg et al. \cite{jaderberg2014deep} & 56.0    & -     & 80.0    & 75.0    & -     & -     & -     & -     & -     & -     & -     & -     & -     & -     & -     & -     & -     & -     & -     & - \\
  Alsharif et al. \cite{Alsharif2014end} & 48.0    & -     & 77.0    & 70.0    & -     & -     & -     & -     & -     & -     & -     & -     & -     & -     & -     & -     & -     & -     & -     & - \\
  Yao et al. \cite{yao2014unified} & -     & -     & -     & -     & -     & 48.6  & -     & -     & -     & -     & -     & -     & -     & -     & -     & -     & -     & -     & -     & - \\
  Neumann et al. \cite{neumann2015real} & 68.1  &       & -     & -     & -     & -     & -     & -     & 45.2  & -     & -     & -     & 35.0    & 19.9  & 15.6  & 35.0    & 19.9  & 15.6  & -     & - \\
  Jaderberg et al. \cite{jaderberg2016reading} & 76.0    & 53.0    & \textbf{90.0}    & \textbf{86.0}    & \textbf{78.0}    & 76.0    & -     & -     & -     & -     & -     & 76.0    & -     & -     & -     & -     & -     & -     & -     & - \\
  *Liao et al. \cite{liao2017textboxes} : TextBoxes & 84.0    & 64.0    & -     & -     & -     & 87.0    & 91.0    & 89.0    & 84.0    & 94.0    & 92.0    & 87.0    & -     & -     & -     & -     & -     & -     & -  & - \\
  Bŭsta et al. \cite{busta2017deep} : Deep TextSpotter & -     & -     & -     & -     & -     & -     & 89.0    & 86.0    & 77.0    & 92.0    & 89.0    & 81.0    & 54.0    & 51.0    & 47.0    & 58.0    & 53.0    & 51.0    & -     & - \\
  Li et al. \cite{li2017towards} & \textbf{84.9} & \textbf{66.2} & -     & -     & -     & \textbf{87.7}  & {91.1} & {89.8}  & {84.6}  & {94.2}  & {92.4}  & {88.2}  & -     & -     & -     & -     & -     & -     & -     & - \\
  Lyu et al. \cite{lyu2018mask} : Mask TextSpotter & -     & -     & -     & -     & -     & -     & 92.2  & 91.1  & 86.5  & 92.5  & 92.0    & 88.2  & 79.3  & 73.0    & 62.4  & 79.3  & 74.5  & 64.2  & 71.8  & 52.9 \\
  He et al. \cite{he2018end} & -     & -     & -     & -     & -     & -     & 91.0    & 89.0    & 86.0    & 93.0    & 92.0    & 87.0    & 82.0    & 77.0    & 63.0    & 85.0    & 80.0    & 65.0    & -     & - \\
  *Liu et al. \cite{liu2018fots} : FOTS & -     & -     & -     & -     & -     & -     & 92.0 & 90.1 & 84.8 & 95.9 & 93.9  & 87.8 & 83.6 & 79.1 & 65.3 & \textbf{87.0} & \textbf{82.4} & 68.0 & -     & - \\
  *Liao et al. \cite{liao2018textboxes++} : TextBoxes++ & 84.0    & 64.0    & -     & -     & -     & -     & 93.0    & \textbf{92.0}    & 85.0    & \textbf{96.0}    & \textbf{95.0}    & 87.0    & 73.3  & 65.9  & 51.9  & 76.5  & 69.0    & 54.4  & -     & - \\
  Liao et al. \cite{liao2019mask} : Mask TextSpotter & -     & -     & -     & -     & -     & -     & \textbf{93.3}  & 91.3  & \textbf{88.2}  & 92.7  & 91.7  & 87.7  & 83.0    & 77.7  & \textbf{73.5}  & 82.4  & 78.1  & \textbf{73.6} & 77.4  & 65.3  \\
  % Neumann et al. \cite{neumann2015efficient} & -     & -     & -     & -     & -     & -     & 93.3  & 91.3  & 88.2  & 92.7  & 91.7  & 87.7  & 83    & 77.7  & 73.5  & 82.4  & 78.1  & 73.6  & 65.3  & 77.4 \\
  % \midrule
  *Xing et al. \cite{xing2019convo} : CharNet & -     & -     & -     & -     & -     & -     & -  & -  & -  & -  & -  & -  & 85.1  & 81.3  & 71.1  & -  & -  & -  & - & 69.2  \\
  Feng et al. \cite{feng2019textdragon} : TextDragon & -     & -     & -     & -     & -     & -     & -  & -  & -  & -  & -  & -  & 82.5  & 78.3  & 65.2  & 86.2  & 81.6  & 68.0  & 74.8 & 48.8  \\
  % \midrule
  Qin et al. \cite{qin2019towards} & -     & -     & -     & -     & -     & -     & -  & -  & -  & -  & -  & -  & \textbf{85.5}  & \textbf{81.9}  & 69.9  & -  & -  & -  & - & \textbf{70.7} \\  
  Wang et al. \cite{wang2019all} : Boundary & -     & -     & -     & -     & -     & -     & 88.2  & 87.7  & 84.1  & -  & -  & -  & 79.7  & 75.2  & 64.1  & -  & -  & -  & 76.1 & 65.0 \\ 
  Qiao et al. \cite{qiao2019text} : Text Perceptron & -     & -     & -     & -     & -     & -     & 91.4  & 90.7  & 85.8  & 94.9  & 94.0  & \textbf{88.5}  & 80.5  & 76.6  & 65.1  & 84.1  & 79.4  & 67.9   & \textbf{78.3}  & 69.7  \\ 
  \bottomrule
\end{tabular}}
\end{table*}%

\begin{table*}[htbp]
\tiny
\centering
\caption{Performance comparison for competitions. NED stands for the normalized edit distance.}
\label{Table_competition_performance}
\begin{tabular}{ccccccc}
  \toprule
  \multirow{2}[1]{*}{\textbf{Competition}} & \multicolumn{3}{c}{\textbf{Detection}} & \multicolumn{3}{c}{\textbf{End-to-End}} \\
  \cmidrule(lr){2-4} \cmidrule(lr){5-7}      & Team Name & Protocol & Result (\%) & Team Name & Protocol & Result (\%) \\
  \midrule
  RCTW  & Foo \& Bar  & F-score & 66.10 & NLPR\_PAL & 1 - NED   & 67.99 \\
  MTWI  & nelslip(iflytek\&ustc) & F-score & 79.60 & nelslip(iflytek\&ustc) & F-score & 81.50 \\
  LSVT  & Tencent-DPPR Team & F-score & 86.42 & Tencent-DPPR Team & F-score & 60.97 \\
  ArT   & pil\_maskrcnn & F-score & 82.65 & baseline\_0.5\_class\_5435 & F-score & 50.17 \\
  ReCTS & SANHL & F-score & 93.36 & Tencent-DPPR & 1 - NED & 81.50 \\
  MLT   & Tencent-DPPR Team & F-score & 83.61 & Tencent-DPPR Team \& USTB-PRIR & F-score & 59.15 \\
  \bottomrule
\end{tabular}%
\end{table*}%

\subsubsection{Evaluation Protocols for Multilingual Text}
\label{evaluation_protocols_for_multilingual_text}

% In this content, we briefly introduce the evaluation protocols for some competitions, such as \textit{RCTW Competition}, \textit{MTWI Competition}, \textit{LSVT Competition}, \textit{ArT Competition}, \textit{ReCTS Competition} and \textit{MLT Competition}.
In this section, we briefly introduce the evaluation protocols for multilingual text widely used in recent competitions, such as \textit{RCTW} \cite{shi2017icdar2017}, \textit{MTWI} \cite{he2018icpr2018}, \textit{LSVT} \cite{sun2019icdar}, \textit{ArT} \cite{chng2019icdar2019}, \textit{ReCTS} \cite{liu2019icdar}, and \textit{MLT} \cite{nayef2019icdar2019} competitions.

\noindent\textbf{Recognition Protocols}
\label{Recognition_Protocols}

Most competitions \cite{he2018icpr2018}, \cite{chng2019icdar2019}, \cite{liu2019icdar} measured the algorithm recognition performance by a traditional evaluation metric — the \textit{normalized edit distance (NED)}:
\begin{equation}
NED =\frac{1}{N}\sum^N_{i=1}D(s_i, \hat{s}_i)/max(l_i, \hat{l}_i),
\end{equation}
where $D(.)$ stands for the Levenshtein distance.
$s_i$ and $\hat{s}_i$ denote the predicted text and the corresponding ground truth, respectively.
Furthermore, $l_i$ and $\hat{l}_i$ are their text length.
$N$ is the total number of text lines.
The \textit{NED} protocol measures the mis-matching between the predicted text and the corresponding ground truth.
Therefore, the recognition score is usually calculated as \textit{$1$-NED}.

\noindent\textbf{End-to-End Protocols}
\label{End_to_End_Protocols}

Two main evaluation protocols for end-to-end systems have been used during recent competitions: 

\begin{itemize} 
\item
% One protocol 
%compared 
The first protocol evaluates the algorithm performance in several aspects, including \textit{precision}, \textit{recall}, and \textit{F-score} based on \textit{NED}.
%In particular, the \textit{NEDs} between the predicted text and the corresponding ground truth are used to evaluate the performance of each algorithm.
%According to the matched rules in \textit{Precision} and \textit{Recall}, the corresponding \textit{NEDs} are summed and divided by the number of test instances, which is called \textit{NED Recall} and \textit{NED Precision} respectively.
According to the matching relationship between the predicted and ground truth bounding boxes, the $1$-\textit{NED} of the predicted text and ground truth text serves as \textit{precision} and \textit{recall} score.
The \textit{F-score} is the harmonic average of the score of \textit{precision} and \textit{recall}.
This is a mainstream metric to evaluate detection and recognition performance simultaneously.
The protocol is widely used in \cite{he2018icpr2018}, \cite{sun2019icdar}, \cite{chng2019icdar2019}, \cite{liu2019icdar}, \cite{nayef2019icdar2019}.

%Based on \textit{NED Recall} and \textit{NED Precision}, the harmonic average of them can be calculated, which serves as a metric to evaluate detection and recognition performance simultaneously.

%This protocol evaluation was widely used in some competitions \cite{he2018icpr2018}, \cite{sun2019icdar}, \cite{chng2019icdar2019}, \cite{liu2019icdar}, \cite{nayef2019icdar2019}.
%In addition, \cite{nayef2019icdar2019} required the predicted text must exactly match the corresponding ground truth instead of the normalized metric with \textit{NEDs}, which was stricter.
%Note that the matched rules may be different in different competitions.

\item
The second protocol measures the algorithm performance by the average \textit{NED}, namely, \textit{AED}.
In particular, the \textit{NED} between the predicted text and the corresponding ground truth are calculated.
%between all matching pairs.
Then, all the \textit{NEDs} are summed and divided by the number of test images, and the result is called \textit{AED}.
Specifically, a lower \textit{AED} means a better performance.
This protocol evaluation was introduced in \cite{shi2017icdar2017} to improve the fairness for long text detection and recognition, which is practical useful for real-world systems.
\end{itemize}

These two types of evaluation protocols evaluate the algorithm from different perspectives.
%and enjoy their own advantages.
%Note that both evaluation protocols are not strictly positively related, i.e., the algorithm rankings under different protocols may be different.
As illustrated in Table~\ref{Table_competition_performance}, the performances of winning systems of several recent end-to-end competitions indicate that the problem of end-to-end recognition remains unsolved.

\subsection{Discussion}
\label{dataset_competition_disscussion}

Various new challenging datasets inspire new research that promotes the progress in the field.
However, it is hard to assess whether and how a newly proposed algorithm improves upon the current art because of the varieties of different datasets, priors, evaluation protocols, and testing environments.
% Different data, priors, evaluation protocols and testing environments make it difficult to compare reported numbers at face value (Table~\ref{Table_recognition_performance} and Table~\ref{Table_end_to_end_performance}).
Therefore, a holistic and fair comparison is necessary for future work \cite{baek2019wrong}, \cite{liu2019tightness}.

Recent datasets and competitions show that the community is moving toward more challenging text recognition tasks (e.g., from horizontal text to irregular text, and from Latin text to multilingual text).
Beyond the challenges, high-quality annotations are also critical for a good dataset.
Moreover, new datasets and competitions may bridge the gap between academia and industry.

\section{Discussion and Future Directions}
\label{discussion_and_future_directions}

Text has played an important role in human lives.
Automatically reading text in natural scenes has a great practical value.
Therefore, scene text recognition has become an important and vibrant research area in computer vision and pattern recognition.
% , as one of the most influential inventions of humanity, has played an important role in human life.
% Automatically reading text in natural scenes has a large potential practical value.
% Therefore, scene text recognition has become an important and vibrant research area in computer vision.
This paper summarizes the fundamental problems and the state-of-the-art methods associated with scene text recognition, introduces new insights and ideas, and provides a comprehensive review of publicly available resources.
In the past decades, there have been substantial advancements in innovation, practicality, and efficiency of recognition methods.
However, there is ample room remaining for future research:

%  benefiting from the development of deep learning, a large number of scene text recognition approaches have been reported with notable success.
% The community has witnessed substantial advancements in the innovation, practicality and efficiency of these approaches.
% However, reading text in natural scenes is still confronted with some challenges. 
% There are ample room remaining for future research:

\begin{itemize} 
\item \textbf{Generalization Ability.}
Generalization ability refers to the ability of recognition algorithms to be effective across a range of inputs and applications.
% Most applications should be adaptable to varying inputs, such as size, shape, and font style. 
% Most applications require recognitionshould be adaptable to varying inputs, such as size, shape, and font style. 
Although the recognition algorithms trained by the synthetic datasets achieve good performance on several realistic evaluation datasets, they fail to adapt to varying inputs, such as text instances with longer characters, smaller sizes, and unseen font styles. 
Moreover, most recognition algorithms are sensitive to environmental interferences and hard to deal with real-world complexity, e.g., the poor performance reported on the COCO-Text dataset.
Therefore, researchers and practitioners have to train models from scratch based on specific inputs and scenarios.

In contrast, humans are adept at recognizing different styles of text under complex scenarios with little supervision learning, which indicates that there still exists a giant gap between the current understanding level of machines and human-level performance.
In addition to simply employing rich and diverse data as training samples, a feasible solution might be to explore the unique and essential representation of text, such as visual-level and semantic-level.

 % and more complex scenarios
% , e.g., COCO-Text dataset.
% Moreover, recognition algorithms should be insensitive to environmental interferences and be able to deal with real-world complexity.
% , which endowing the recognition system with the ability to deal with real-world complexity.

% Generalization ability is important for algorithms as most applications would require the adaptability to varying environments.

\item \textbf{Evaluation Protocols.}
Numerous approaches proposed in recent years claimed to have pushed the boundary of the technology.
% It is hard to assess whether and how a newly proposed algorithm push the boundary of the technology.
However, the inconsistency of datasets, priors, and testing environments makes it difficult to fairly evaluate the reported numbers at face value in Table~\ref{Table_recognition_performance} and Table~\ref{Table_end_to_end_performance}.
Researchers and practitioners have to confirm and compare the experimental settings in newly proposed algorithms.
For example, which training datasets were used, e.g., synthetic datasets, realistic datasets or a mixture of both? which annotations were used, e.g., word-level, character-level or pixel-level?
Considering this, a fair comparison is required in the community.
For example, future work might report recognition performance on the unified training/testing datasets or even report recognition performance on a single model, i.e., evaluate the performance of the same model across different datasets.
Moreover, clear and detailed experimental settings introduced in papers are also important in advancing research progress.
% A fair comparison has been largely missing in the community.

\item \textbf{Data Issues.}
Most deep learning algorithms highly depend on a sufficient amount of high quality data.
The existing realistic datasets only contains thousands of data samples, which is relatively small for training a accurate scene text recognizer.
Moreover, manually collecting and annotating large amount of real-world data will involve huge efforts and resources.
Therefore, there are two aspects to be considered.
On one hand, synthesizing as realistic and effective data as possible has a potential in the community.
Compared with realistic datasets, multi-level annotation information (i.e., word-level, character-level and pixel-level) can be easily obtained during synthesizing, which can be used to train data-hungry algorithms.
For example, some researchers \cite{Long2020UnrealText} are working to synthesis realistic text instances by a $3$D engine.
% algorithms. data-hungry deep learning algorithms instead of real-world data.
On the other hand, approaches of using unlabeled real-world data are worth considering in the future.

It is valuable to explore how to use the existing data efficiently.
For example, with the emergence of many realistic datasets, we should reconsider whether unified synthetic datasets are the only choice for training models and then evaluated with realistic datasets. (Such strategy is widely adopted in most of current researches.)
The balance between realistic datasets and synthetic datasets needs to be further developed.
Moreover, developing efficient data augmentation approaches for text might be a feasible and promising solution, which should focus more on the style of multi-objects.
 % necessary for evaluating recognition algorithms.

\item \textbf{Scenarios.}
The research aims to improve human quality of life.
However, for STR, the gap between the research and applications still exists.
 % (e.g., more complex backgrounds and more noise in the real world).
In practical applications, text usually appears with worse image quality, more complex backgrounds, and more noise, which requires the recognition systems with the ability to deal with real-world complexity.
Meanwhile, for simple but private vision-based scenarios, such as bank cards, recognition performance is especially important.
Thus, researchers and practitioners should not be limited to several standard benchmarks.
Challenges in real-world applications may provide new research opportunities and advance research progress in the future, such as multilingual text recognition in modern cities, ultra-high precision recognition in private scenarios, and fast text recognition for mobiles.

\item \textbf{Image Preprocessing.}
To improve the recognition performance of algorithms, increasingly complex 
%neural networks 
recognizers have become a new trend in the community.
However, this is not the only perspective worth considering.
%Is this the only resolution?
Some potential image preprocessing issues deserve the attention of researchers, such as TextSR \cite{peyrard2015icdar2015} and background removal \cite{Luo2020Separating}, which can significantly reduce the difficulties of STR and improve performance from a new perspective.

\item \textbf{End-to-End Systems.}
% Given imagery with complex backgrounds as input, an end-to-end system can directly convert all text regions into string sequences, which has a wide range of application scenarios.
Constructing a real-time and efficient end-to-end system has attracted the interest of researchers and practitioners.
However, the performance of end-to-end systems remains far behind compared with that of OCR in scanned documents.
Some difficulties should be considered, such as efficiently bridging and sharing information between text detection and recognition, balancing the significant differences in learning difficulty and convergence speed between text detection and recognition, and improving joint optimization.
% Simple, compact, yet powerful end-to-end systems may be a new trend.
In this area, there is much work to be done.
Furthermore, it is worth considering whether end-to-end solutions are necessary for industrial applications.
% In addition to improving text detection and recognition branches, exploring the full potential of information sharing between text detection and recognition is worthy to study in the future.

\item \textbf{Languages.}
Representatives of multiple cultures live and communicate in modern cities.
Multilingual text recognition is critical to human communication as well as smart city development.
In addition to construct large-scale synthetic/realistic multilingual training datasets, a feasible solution might be combined with script identification.
Moreover, although many recognition algorithms exist, most of them focus on Latin text only.
Recognition of non-Latin has not been extensively investigated, such as Chinese scene text, which is large-scale category text and has unique characteristics compared with Latin text.
Existing recognition algorithms cannot be well generalized to different languages.
Developing language-dependent recognition algorithms for specific language might be a feasible solution.

\item \textbf{Security.}
% In recent years, adversarial attack has attracted the attention of numerous researchers.
As STR algorithms can be adapted to many private vision-based scenarios (such as bank cards, ID cards, and driver licenses), the security of recognition approaches is very important.
Despite high performance, most deep learning-based text recognizers are highly vulnerable to adversarial examples.
Strengthening the security of STR algorithms will be a potential direction in the future.

\item \textbf{STR + NLP.}
NLP is a bridge in human–computer communication.
Meanwhile, text is the most important carrier of communication and perception in the world.
A combination of NLP and STR may be an important trend in various fields, such as text VQA \cite{singh2019towards}, \cite{Xinyu2020On}, document understanding \cite{katti2018chargrid}, and information extraction \cite{liu2019graph}, \cite{dang2019end}.

\end{itemize}

\bibliographystyle{ACM-Reference-Format}
\bibliography{acmart}


\begin{thebibliography}{243}

%%% the Web address.

\ifx \showCODEN    \undefined \def \showCODEN     #1{\unskip}     \fi
\ifx \showDOI      \undefined \def \showDOI       #1{#1}\fi
\ifx \showISBNx    \undefined \def \showISBNx     #1{\unskip}     \fi
\ifx \showISBNxiii \undefined \def \showISBNxiii  #1{\unskip}     \fi
\ifx \showISSN     \undefined \def \showISSN      #1{\unskip}     \fi
\ifx \showLCCN     \undefined \def \showLCCN      #1{\unskip}     \fi
\ifx \shownote     \undefined \def \shownote      #1{#1}          \fi
\ifx \showarticletitle \undefined \def \showarticletitle #1{#1}   \fi
\ifx \showURL      \undefined \def \showURL       {\relax}        \fi
% invisible to TeX
\providecommand\bibfield[2]{#2}
\providecommand\bibinfo[2]{#2}
\providecommand\natexlab[1]{#1}
\providecommand\showeprint[2][]{arXiv:#2}

\bibitem[\protect\citeauthoryear{Almaz{\'a}n, Gordo, Forn{\'e}s, and
  Valveny}{Almaz{\'a}n et~al\mbox{.}}{2014}]%
        {almazan2014word}
\bibfield{author}{\bibinfo{person}{Jon Almaz{\'a}n}, \bibinfo{person}{Albert
  Gordo}, \bibinfo{person}{Alicia Forn{\'e}s}, {and} \bibinfo{person}{Ernest
  Valveny}.} \bibinfo{year}{2014}\natexlab{}.
\newblock \showarticletitle{Word spotting and recognition with embedded
  attributes}.
\newblock \bibinfo{journal}{\emph{IEEE Trans. Pattern Anal. Mach. Intell}}
  \bibinfo{volume}{36}, \bibinfo{number}{12} (\bibinfo{year}{2014}),
  \bibinfo{pages}{2552--2566}.
\newblock


\bibitem[\protect\citeauthoryear{Alsharif and Pineau}{Alsharif and
  Pineau}{2014}]%
        {Alsharif2014end}
\bibfield{author}{\bibinfo{person}{Ouais Alsharif} {and}
  \bibinfo{person}{Joelle Pineau}.} \bibinfo{year}{2014}\natexlab{}.
\newblock \showarticletitle{End-to-End Text Recognition with Hybrid {HMM}
  Maxout Models}. In \bibinfo{booktitle}{\emph{Proceedings of ICLR: Workshop}}.
\newblock


\bibitem[\protect\citeauthoryear{Anderson, He, Buehler, Teney, Johnson, Gould,
  and Zhang}{Anderson et~al\mbox{.}}{2018}]%
        {anderson2018bottom}
\bibfield{author}{\bibinfo{person}{Peter Anderson}, \bibinfo{person}{Xiaodong
  He}, \bibinfo{person}{Chris Buehler}, \bibinfo{person}{Damien Teney},
  \bibinfo{person}{Mark Johnson}, \bibinfo{person}{Stephen Gould}, {and}
  \bibinfo{person}{Lei Zhang}.} \bibinfo{year}{2018}\natexlab{}.
\newblock \showarticletitle{Bottom-up and top-down attention for image
  captioning and visual question answering}. In
  \bibinfo{booktitle}{\emph{Proceedings of CVPR}}. \bibinfo{pages}{6077--6086}.
\newblock


\bibitem[\protect\citeauthoryear{Baek, Kim, Lee, Park, Han, Yun, Oh, and
  Lee}{Baek et~al\mbox{.}}{2019a}]%
        {baek2019wrong}
\bibfield{author}{\bibinfo{person}{Jeonghun Baek}, \bibinfo{person}{Geewook
  Kim}, \bibinfo{person}{Junyeop Lee}, \bibinfo{person}{Sungrae Park},
  \bibinfo{person}{Dongyoon Han}, \bibinfo{person}{Sangdoo Yun},
  \bibinfo{person}{Seong~Joon Oh}, {and} \bibinfo{person}{Hwalsuk Lee}.}
  \bibinfo{year}{2019}\natexlab{a}.
\newblock \showarticletitle{What is wrong with scene text recognition model
  comparisons? dataset and model analysis}. In
  \bibinfo{booktitle}{\emph{Proceedings of ICCV}}. \bibinfo{pages}{4714--4722}.
\newblock


\bibitem[\protect\citeauthoryear{Baek, Lee, Han, Yun, and Lee}{Baek
  et~al\mbox{.}}{2019b}]%
        {baek2019character}
\bibfield{author}{\bibinfo{person}{Youngmin Baek}, \bibinfo{person}{Bado Lee},
  \bibinfo{person}{Dongyoon Han}, \bibinfo{person}{Sangdoo Yun}, {and}
  \bibinfo{person}{Hwalsuk Lee}.} \bibinfo{year}{2019}\natexlab{b}.
\newblock \showarticletitle{Character region awareness for text detection}. In
  \bibinfo{booktitle}{\emph{Proceedings of CVPR}}. \bibinfo{pages}{9365--9374}.
\newblock


\bibitem[\protect\citeauthoryear{Bahdanau, Cho, and Bengio}{Bahdanau
  et~al\mbox{.}}{2015}]%
        {bahdanau2014neural}
\bibfield{author}{\bibinfo{person}{Dzmitry Bahdanau},
  \bibinfo{person}{Kyunghyun Cho}, {and} \bibinfo{person}{Yoshua Bengio}.}
  \bibinfo{year}{2015}\natexlab{}.
\newblock \showarticletitle{Neural machine translation by jointly learning to
  align and translate}. In \bibinfo{booktitle}{\emph{Proceedings of ICLR}}.
\newblock


\bibitem[\protect\citeauthoryear{Bahdanau, Chorowski, Serdyuk, Brakel, and
  Bengio}{Bahdanau et~al\mbox{.}}{2016}]%
        {bahdanau2016end}
\bibfield{author}{\bibinfo{person}{Dzmitry Bahdanau}, \bibinfo{person}{Jan
  Chorowski}, \bibinfo{person}{Dmitriy Serdyuk}, \bibinfo{person}{Philemon
  Brakel}, {and} \bibinfo{person}{Yoshua Bengio}.}
  \bibinfo{year}{2016}\natexlab{}.
\newblock \showarticletitle{End-to-end attention-based large vocabulary speech
  recognition}. In \bibinfo{booktitle}{\emph{Proceedings of ICASSP}}.
  \bibinfo{pages}{4945--4949}.
\newblock


\bibitem[\protect\citeauthoryear{Bai, Cheng, Niu, Pu, and Zhou}{Bai
  et~al\mbox{.}}{2018a}]%
        {bai2018edit}
\bibfield{author}{\bibinfo{person}{Fan Bai}, \bibinfo{person}{Zhanzhan Cheng},
  \bibinfo{person}{Yi Niu}, \bibinfo{person}{Shiliang Pu}, {and}
  \bibinfo{person}{Shuigeng Zhou}.} \bibinfo{year}{2018}\natexlab{a}.
\newblock \showarticletitle{Edit Probability for Scene Text Recognition}. In
  \bibinfo{booktitle}{\emph{Proceedings of CVPR}}. \bibinfo{pages}{1508--1516}.
\newblock


\bibitem[\protect\citeauthoryear{Bai, Yang, Lyu, Xu, and Luo}{Bai
  et~al\mbox{.}}{2018b}]%
        {bai2018integrating}
\bibfield{author}{\bibinfo{person}{Xiang Bai}, \bibinfo{person}{Mingkun Yang},
  \bibinfo{person}{Pengyuan Lyu}, \bibinfo{person}{Yongchao Xu}, {and}
  \bibinfo{person}{Jiebo Luo}.} \bibinfo{year}{2018}\natexlab{b}.
\newblock \showarticletitle{Integrating scene text and visual appearance for
  fine-grained image classification}.
\newblock \bibinfo{journal}{\emph{IEEE Access}}  \bibinfo{volume}{6}
  (\bibinfo{year}{2018}), \bibinfo{pages}{66322--66335}.
\newblock


\bibitem[\protect\citeauthoryear{Baker and Kanade}{Baker and Kanade}{2002}]%
        {baker2002limits}
\bibfield{author}{\bibinfo{person}{Simon Baker} {and} \bibinfo{person}{Takeo
  Kanade}.} \bibinfo{year}{2002}\natexlab{}.
\newblock \showarticletitle{Limits on super-resolution and how to break them}.
\newblock \bibinfo{journal}{\emph{IEEE Trans. Pattern Anal. Mach. Intell}}
  \bibinfo{number}{9} (\bibinfo{year}{2002}), \bibinfo{pages}{1167--1183}.
\newblock


\bibitem[\protect\citeauthoryear{Bartz, Yang, and Meinel}{Bartz
  et~al\mbox{.}}{2018}]%
        {bartz2018see}
\bibfield{author}{\bibinfo{person}{Christian Bartz}, \bibinfo{person}{Haojin
  Yang}, {and} \bibinfo{person}{Christoph Meinel}.}
  \bibinfo{year}{2018}\natexlab{}.
\newblock \showarticletitle{{SEE}: towards semi-supervised end-to-end scene
  text recognition}. In \bibinfo{booktitle}{\emph{Proceedings of AAAI}}.
  \bibinfo{pages}{6674--6681}.
\newblock


\bibitem[\protect\citeauthoryear{Bissacco, Cummins, Netzer, and Neven}{Bissacco
  et~al\mbox{.}}{2013}]%
        {bissacco2013photoocr}
\bibfield{author}{\bibinfo{person}{Alessandro Bissacco}, \bibinfo{person}{Mark
  Cummins}, \bibinfo{person}{Yuval Netzer}, {and} \bibinfo{person}{Hartmut
  Neven}.} \bibinfo{year}{2013}\natexlab{}.
\newblock \showarticletitle{Photoocr: Reading text in uncontrolled conditions}.
  In \bibinfo{booktitle}{\emph{Proceedings of ICCV}}.
  \bibinfo{pages}{785--792}.
\newblock


\bibitem[\protect\citeauthoryear{Biten, Tito, Mafla, Gomez, Rusinol, Valveny,
  Jawahar, and Karatzas}{Biten et~al\mbox{.}}{2019}]%
        {biten2019scene}
\bibfield{author}{\bibinfo{person}{Ali~Furkan Biten}, \bibinfo{person}{Ruben
  Tito}, \bibinfo{person}{Andres Mafla}, \bibinfo{person}{Lluis Gomez},
  \bibinfo{person}{Mar{\c{c}}al Rusinol}, \bibinfo{person}{Ernest Valveny},
  \bibinfo{person}{CV Jawahar}, {and} \bibinfo{person}{Dimosthenis Karatzas}.}
  \bibinfo{year}{2019}\natexlab{}.
\newblock \showarticletitle{Scene text visual question answering}. In
  \bibinfo{booktitle}{\emph{Proceedings of ICCV}}. \bibinfo{pages}{4291--4301}.
\newblock


\bibitem[\protect\citeauthoryear{Bluche}{Bluche}{2016}]%
        {bluche2016joint}
\bibfield{author}{\bibinfo{person}{Th{\'e}odore Bluche}.}
  \bibinfo{year}{2016}\natexlab{}.
\newblock \showarticletitle{Joint line segmentation and transcription for
  end-to-end handwritten paragraph recognition}. In
  \bibinfo{booktitle}{\emph{Proceedings of NIPS}}. \bibinfo{pages}{838--846}.
\newblock


\bibitem[\protect\citeauthoryear{Busta, Neumann, and Matas}{Busta
  et~al\mbox{.}}{2017}]%
        {busta2017deep}
\bibfield{author}{\bibinfo{person}{Michal Busta}, \bibinfo{person}{Lukas
  Neumann}, {and} \bibinfo{person}{Jiri Matas}.}
  \bibinfo{year}{2017}\natexlab{}.
\newblock \showarticletitle{Deep textspotter: An end-to-end trainable scene
  text localization and recognition framework}. In
  \bibinfo{booktitle}{\emph{Proceedings of ICCV}}. \bibinfo{pages}{2204--2212}.
\newblock


\bibitem[\protect\citeauthoryear{Caner and Haritaoglu}{Caner and
  Haritaoglu}{2010}]%
        {caner2010shape}
\bibfield{author}{\bibinfo{person}{Gulcin Caner} {and} \bibinfo{person}{Ismail
  Haritaoglu}.} \bibinfo{year}{2010}\natexlab{}.
\newblock \showarticletitle{Shape-dna: effective character restoration and
  enhancement for Arabic text documents}. In
  \bibinfo{booktitle}{\emph{Proceedings of ICPR}}. \bibinfo{pages}{2053--2056}.
\newblock


\bibitem[\protect\citeauthoryear{Canny}{Canny}{1986}]%
        {canny1986computational}
\bibfield{author}{\bibinfo{person}{John Canny}.}
  \bibinfo{year}{1986}\natexlab{}.
\newblock \showarticletitle{A computational approach to edge detection}.
\newblock \bibinfo{journal}{\emph{IEEE Trans. Pattern Anal. Mach. Intell}}
  \bibinfo{number}{6} (\bibinfo{year}{1986}), \bibinfo{pages}{679--698}.
\newblock


\bibitem[\protect\citeauthoryear{Casey and Lecolinet}{Casey and
  Lecolinet}{1996}]%
        {casey1996survey}
\bibfield{author}{\bibinfo{person}{Richard~G Casey} {and} \bibinfo{person}{Eric
  Lecolinet}.} \bibinfo{year}{1996}\natexlab{}.
\newblock \showarticletitle{A survey of methods and strategies in character
  segmentation}.
\newblock \bibinfo{journal}{\emph{IEEE Trans. Pattern Anal. Mach. Intell}}
  \bibinfo{volume}{18}, \bibinfo{number}{7} (\bibinfo{year}{1996}),
  \bibinfo{pages}{690--706}.
\newblock


\bibitem[\protect\citeauthoryear{Chen, Desai, and Zhou}{Chen
  et~al\mbox{.}}{2007}]%
        {chen2007cindi}
\bibfield{author}{\bibinfo{person}{Rui Chen}, \bibinfo{person}{Bipin~C Desai},
  {and} \bibinfo{person}{Cong Zhou}.} \bibinfo{year}{2007}\natexlab{}.
\newblock \showarticletitle{{CINDI} robot: an intelligent Web crawler based on
  multi-level inspection}. In \bibinfo{booktitle}{\emph{Eleventh International
  Database Engineering and Applications Symposium (IDEAS)}}.
  \bibinfo{pages}{93--101}.
\newblock


\bibitem[\protect\citeauthoryear{Chen, Wang, Zhu, Jin, and Luo}{Chen
  et~al\mbox{.}}{2020}]%
        {chen2019adaptive}
\bibfield{author}{\bibinfo{person}{Xiaoxue Chen}, \bibinfo{person}{Tianwei
  Wang}, \bibinfo{person}{Yuanzhi Zhu}, \bibinfo{person}{Lianwen Jin}, {and}
  \bibinfo{person}{Canjie Luo}.} \bibinfo{year}{2020}\natexlab{}.
\newblock \showarticletitle{Adaptive Embedding Gate for Attention-Based Scene
  Text Recognition}.
\newblock \bibinfo{journal}{\emph{Neurocomputing}}  \bibinfo{volume}{381}
  (\bibinfo{year}{2020}), \bibinfo{pages}{261--271}.
\newblock


\bibitem[\protect\citeauthoryear{Chen, Yang, Zhang, and Waibel}{Chen
  et~al\mbox{.}}{2004}]%
        {chen2004automatic}
\bibfield{author}{\bibinfo{person}{Xilin Chen}, \bibinfo{person}{Jie Yang},
  \bibinfo{person}{Jing Zhang}, {and} \bibinfo{person}{Alex Waibel}.}
  \bibinfo{year}{2004}\natexlab{}.
\newblock \showarticletitle{Automatic detection and recognition of signs from
  natural scenes}.
\newblock \bibinfo{journal}{\emph{IEEE Transactions on Image Processing}}
  \bibinfo{volume}{13}, \bibinfo{number}{1} (\bibinfo{year}{2004}),
  \bibinfo{pages}{87--99}.
\newblock


\bibitem[\protect\citeauthoryear{Cheng, Huang, Bai, Feng, and Liu}{Cheng
  et~al\mbox{.}}{2019}]%
        {cheng2019patch}
\bibfield{author}{\bibinfo{person}{Changxu Cheng}, \bibinfo{person}{Qiuhui
  Huang}, \bibinfo{person}{Xiang Bai}, \bibinfo{person}{Bin Feng}, {and}
  \bibinfo{person}{Wenyu Liu}.} \bibinfo{year}{2019}\natexlab{}.
\newblock \showarticletitle{Patch Aggregator for Scene Text Script
  Identification}. In \bibinfo{booktitle}{\emph{Proceedings of ICDAR}}.
  \bibinfo{pages}{1077--1083}.
\newblock


\bibitem[\protect\citeauthoryear{Cheng}{Cheng}{2019}]%
        {cheng2019semi}
\bibfield{author}{\bibinfo{person}{Yong Cheng}.}
  \bibinfo{year}{2019}\natexlab{}.
\newblock \showarticletitle{Semi-supervised learning for neural machine
  translation}.
\newblock In \bibinfo{booktitle}{\emph{Joint Training for Neural Machine
  Translation}}. \bibinfo{pages}{25--40}.
\newblock


\bibitem[\protect\citeauthoryear{Cheng, Bai, Xu, Zheng, Pu, and Zhou}{Cheng
  et~al\mbox{.}}{2017}]%
        {cheng2017focusing}
\bibfield{author}{\bibinfo{person}{Zhanzhan Cheng}, \bibinfo{person}{Fan Bai},
  \bibinfo{person}{Yunlu Xu}, \bibinfo{person}{Gang Zheng},
  \bibinfo{person}{Shiliang Pu}, {and} \bibinfo{person}{Shuigeng Zhou}.}
  \bibinfo{year}{2017}\natexlab{}.
\newblock \showarticletitle{Focusing attention: Towards accurate text
  recognition in natural images}. In \bibinfo{booktitle}{\emph{Proceedings of
  ICCV}}. \bibinfo{pages}{5086--5094}.
\newblock


\bibitem[\protect\citeauthoryear{Cheng, Xu, Bai, Niu, Pu, and Zhou}{Cheng
  et~al\mbox{.}}{2018}]%
        {cheng2018aon}
\bibfield{author}{\bibinfo{person}{Zhanzhan Cheng}, \bibinfo{person}{Yangliu
  Xu}, \bibinfo{person}{Fan Bai}, \bibinfo{person}{Yi Niu},
  \bibinfo{person}{Shiliang Pu}, {and} \bibinfo{person}{Shuigeng Zhou}.}
  \bibinfo{year}{2018}\natexlab{}.
\newblock \showarticletitle{{AON}: Towards arbitrarily-oriented text
  recognition}. In \bibinfo{booktitle}{\emph{Proceedings of CVPR}}.
  \bibinfo{pages}{5571--5579}.
\newblock


\bibitem[\protect\citeauthoryear{Ch’ng, Chan, and Liu}{Ch’ng
  et~al\mbox{.}}{2019}]%
        {ch2017total}
\bibfield{author}{\bibinfo{person}{Chee-Kheng Ch’ng},
  \bibinfo{person}{Chee~Seng Chan}, {and} \bibinfo{person}{Cheng-Lin Liu}.}
  \bibinfo{year}{2019}\natexlab{}.
\newblock \showarticletitle{Total-Text: toward orientation robustness in scene
  text detection}.
\newblock \bibinfo{journal}{\emph{International Journal on Document Analysis
  and Recognition (IJDAR)}} (\bibinfo{year}{2019}), \bibinfo{pages}{1--22}.
\newblock


\bibitem[\protect\citeauthoryear{Chng, Liu, Sun, Ng, Luo, Ni, Fang, Zhang, Han,
  Ding, et~al\mbox{.}}{Chng et~al\mbox{.}}{2019}]%
        {chng2019icdar2019}
\bibfield{author}{\bibinfo{person}{Chee-Kheng Chng}, \bibinfo{person}{Yuliang
  Liu}, \bibinfo{person}{Yipeng Sun}, \bibinfo{person}{Chun~Chet Ng},
  \bibinfo{person}{Canjie Luo}, \bibinfo{person}{Zihan Ni},
  \bibinfo{person}{ChuanMing Fang}, \bibinfo{person}{Shuaitao Zhang},
  \bibinfo{person}{Junyu Han}, \bibinfo{person}{Errui Ding}, {et~al\mbox{.}}}
  \bibinfo{year}{2019}\natexlab{}.
\newblock \showarticletitle{{ICDAR}2019 Robust Reading Challenge on
  Arbitrary-Shaped Text ({RRC-ArT})}. In \bibinfo{booktitle}{\emph{Proceedings
  of ICDAR}}. \bibinfo{pages}{1571--1576}.
\newblock


\bibitem[\protect\citeauthoryear{Cho, Sung, and Jun}{Cho et~al\mbox{.}}{2016}]%
        {cho2016canny}
\bibfield{author}{\bibinfo{person}{Hojin Cho}, \bibinfo{person}{Myungchul
  Sung}, {and} \bibinfo{person}{Bongjin Jun}.} \bibinfo{year}{2016}\natexlab{}.
\newblock \showarticletitle{Canny text detector: Fast and robust scene text
  localization algorithm}. In \bibinfo{booktitle}{\emph{Proceedings of CVPR}}.
  \bibinfo{pages}{3566--3573}.
\newblock


\bibitem[\protect\citeauthoryear{Cho, Van~Merri{\"e}nboer, Gulcehre, Bahdanau,
  Bougares, Schwenk, and Bengio}{Cho et~al\mbox{.}}{2014}]%
        {cho2014properties}
\bibfield{author}{\bibinfo{person}{Kyunghyun Cho}, \bibinfo{person}{Bart
  Van~Merri{\"e}nboer}, \bibinfo{person}{Caglar Gulcehre},
  \bibinfo{person}{Dzmitry Bahdanau}, \bibinfo{person}{Fethi Bougares},
  \bibinfo{person}{Holger Schwenk}, {and} \bibinfo{person}{Yoshua Bengio}.}
  \bibinfo{year}{2014}\natexlab{}.
\newblock \showarticletitle{Learning phrase representations using {RNN}
  encoder-decoder for statistical machine translation}. In
  \bibinfo{booktitle}{\emph{Proceedings of EMNLP}}.
  \bibinfo{pages}{1724--1734}.
\newblock


\bibitem[\protect\citeauthoryear{Chowdhury}{Chowdhury}{2003}]%
        {chowdhury2003natural}
\bibfield{author}{\bibinfo{person}{Gobinda~G Chowdhury}.}
  \bibinfo{year}{2003}\natexlab{}.
\newblock \showarticletitle{Natural language processing}.
\newblock \bibinfo{journal}{\emph{Annual review of information science and
  technology}} \bibinfo{volume}{37}, \bibinfo{number}{1}
  (\bibinfo{year}{2003}), \bibinfo{pages}{51--89}.
\newblock


\bibitem[\protect\citeauthoryear{Cong, Hu, Qiang, and Guo}{Cong
  et~al\mbox{.}}{2019}]%
        {cong2019acomparative}
\bibfield{author}{\bibinfo{person}{Fuze Cong}, \bibinfo{person}{Wenping Hu},
  \bibinfo{person}{Huo Qiang}, {and} \bibinfo{person}{Li Guo}.}
  \bibinfo{year}{2019}\natexlab{}.
\newblock \showarticletitle{A Comparative Study of Attention-based
  Encoder-Decoder Approaches to Natural Scene Text Recognition}. In
  \bibinfo{booktitle}{\emph{Proceedings of ICDAR}}. \bibinfo{pages}{916--921}.
\newblock


\bibitem[\protect\citeauthoryear{Corbelli, Baraldi, Grana, and
  Cucchiara}{Corbelli et~al\mbox{.}}{2016}]%
        {corbelli2016historical}
\bibfield{author}{\bibinfo{person}{Andrea Corbelli}, \bibinfo{person}{Lorenzo
  Baraldi}, \bibinfo{person}{Costantino Grana}, {and} \bibinfo{person}{Rita
  Cucchiara}.} \bibinfo{year}{2016}\natexlab{}.
\newblock \showarticletitle{Historical document digitization through layout
  analysis and deep content classification}. In
  \bibinfo{booktitle}{\emph{Proceedings of ICPR}}. \bibinfo{pages}{4077--4082}.
\newblock


\bibitem[\protect\citeauthoryear{Dai, Zhang, and Cao}{Dai
  et~al\mbox{.}}{2019}]%
        {dai2019deep}
\bibfield{author}{\bibinfo{person}{Pengwen Dai}, \bibinfo{person}{Hua Zhang},
  {and} \bibinfo{person}{Xiaochun Cao}.} \bibinfo{year}{2019}\natexlab{}.
\newblock \showarticletitle{Deep Multi-Scale Context Aware Feature Aggregation
  for Curved Scene Text Detection}.
\newblock \bibinfo{journal}{\emph{IEEE Transactions on Multimedia}}
  (\bibinfo{year}{2019}).
\newblock


\bibitem[\protect\citeauthoryear{Dalal and Triggs}{Dalal and Triggs}{2005}]%
        {dalal2005histograms}
\bibfield{author}{\bibinfo{person}{Navneet Dalal} {and} \bibinfo{person}{Bill
  Triggs}.} \bibinfo{year}{2005}\natexlab{}.
\newblock \showarticletitle{Histograms of oriented gradients for human
  detection}. In \bibinfo{booktitle}{\emph{Proceedings of CVPR}}.
  \bibinfo{pages}{886--893}.
\newblock


\bibitem[\protect\citeauthoryear{Dang and Thanh}{Dang and Thanh}{2019}]%
        {dang2019end}
\bibfield{author}{\bibinfo{person}{Tuan Anh~Nguyen Dang} {and}
  \bibinfo{person}{Dat~Nguyen Thanh}.} \bibinfo{year}{2019}\natexlab{}.
\newblock \showarticletitle{End-to-End Information Extraction by
  Character-Level Embedding and Multi-Stage Attentional U-Net}. In
  \bibinfo{booktitle}{\emph{Proceedings of BMVC}}. \bibinfo{pages}{96}.
\newblock


\bibitem[\protect\citeauthoryear{Das, Datta, Gkioxari, Lee, Parikh, and
  Batra}{Das et~al\mbox{.}}{2018}]%
        {das2018embodied}
\bibfield{author}{\bibinfo{person}{Abhishek Das}, \bibinfo{person}{Samyak
  Datta}, \bibinfo{person}{Georgia Gkioxari}, \bibinfo{person}{Stefan Lee},
  \bibinfo{person}{Devi Parikh}, {and} \bibinfo{person}{Dhruv Batra}.}
  \bibinfo{year}{2018}\natexlab{}.
\newblock \showarticletitle{Embodied question answering}. In
  \bibinfo{booktitle}{\emph{Proceedings of CVPR: Workshops}}.
  \bibinfo{pages}{2054--2063}.
\newblock


\bibitem[\protect\citeauthoryear{Dean, Corrado, Monga, Chen, Devin, Mao,
  Ranzato, Senior, Tucker, Yang, et~al\mbox{.}}{Dean et~al\mbox{.}}{2012}]%
        {dean2012large}
\bibfield{author}{\bibinfo{person}{Jeffrey Dean}, \bibinfo{person}{Greg
  Corrado}, \bibinfo{person}{Rajat Monga}, \bibinfo{person}{Kai Chen},
  \bibinfo{person}{Matthieu Devin}, \bibinfo{person}{Mark Mao},
  \bibinfo{person}{Marc'aurelio Ranzato}, \bibinfo{person}{Andrew Senior},
  \bibinfo{person}{Paul Tucker}, \bibinfo{person}{Ke Yang}, {et~al\mbox{.}}}
  \bibinfo{year}{2012}\natexlab{}.
\newblock \showarticletitle{Large scale distributed deep networks}. In
  \bibinfo{booktitle}{\emph{Proceedings of NIPS}}. \bibinfo{pages}{1223--1231}.
\newblock


\bibitem[\protect\citeauthoryear{DeSouza and Kak}{DeSouza and Kak}{2002}]%
        {desouza2002vision}
\bibfield{author}{\bibinfo{person}{Guilherme~N DeSouza} {and}
  \bibinfo{person}{Avinash~C Kak}.} \bibinfo{year}{2002}\natexlab{}.
\newblock \showarticletitle{Vision for mobile robot navigation: A survey}.
\newblock \bibinfo{journal}{\emph{IEEE Trans. Pattern Anal. Mach. Intell}}
  \bibinfo{volume}{24}, \bibinfo{number}{2} (\bibinfo{year}{2002}),
  \bibinfo{pages}{237--267}.
\newblock


\bibitem[\protect\citeauthoryear{Dong, Loy, He, and Tang}{Dong
  et~al\mbox{.}}{2015}]%
        {dong2015image}
\bibfield{author}{\bibinfo{person}{Chao Dong}, \bibinfo{person}{Chen~Change
  Loy}, \bibinfo{person}{Kaiming He}, {and} \bibinfo{person}{Xiaoou Tang}.}
  \bibinfo{year}{2015}\natexlab{}.
\newblock \showarticletitle{Image super-resolution using deep convolutional
  networks}.
\newblock \bibinfo{journal}{\emph{IEEE Trans. Pattern Anal. Mach. Intell}}
  \bibinfo{volume}{38}, \bibinfo{number}{2} (\bibinfo{year}{2015}),
  \bibinfo{pages}{295--307}.
\newblock


\bibitem[\protect\citeauthoryear{Elhabian, El-Sayed, and Ahmed}{Elhabian
  et~al\mbox{.}}{2008}]%
        {elhabian2008moving}
\bibfield{author}{\bibinfo{person}{Shireen~Y Elhabian},
  \bibinfo{person}{Khaled~M El-Sayed}, {and} \bibinfo{person}{Sumaya~H Ahmed}.}
  \bibinfo{year}{2008}\natexlab{}.
\newblock \showarticletitle{Moving object detection in spatial domain using
  background removal techniques-state-of-art}.
\newblock \bibinfo{journal}{\emph{Recent patents on computer science}}
  \bibinfo{volume}{1}, \bibinfo{number}{1} (\bibinfo{year}{2008}),
  \bibinfo{pages}{32--54}.
\newblock


\bibitem[\protect\citeauthoryear{Epshtein, Ofek, and Wexler}{Epshtein
  et~al\mbox{.}}{2010}]%
        {epshtein2010detecting}
\bibfield{author}{\bibinfo{person}{Boris Epshtein}, \bibinfo{person}{Eyal
  Ofek}, {and} \bibinfo{person}{Yonatan Wexler}.}
  \bibinfo{year}{2010}\natexlab{}.
\newblock \showarticletitle{Detecting text in natural scenes with stroke width
  transform}. In \bibinfo{booktitle}{\emph{Proceedings of CVPR}}. IEEE,
  \bibinfo{pages}{2963--2970}.
\newblock


\bibitem[\protect\citeauthoryear{Ezaki, Kiyota, Minh, Bulacu, and
  Schomaker}{Ezaki et~al\mbox{.}}{2005}]%
        {ezaki2005improved}
\bibfield{author}{\bibinfo{person}{Nobuo Ezaki}, \bibinfo{person}{Kimiyasu
  Kiyota}, \bibinfo{person}{Bui~Truong Minh}, \bibinfo{person}{Marius Bulacu},
  {and} \bibinfo{person}{Lambert Schomaker}.} \bibinfo{year}{2005}\natexlab{}.
\newblock \showarticletitle{Improved text-detection methods for a camera-based
  text reading system for blind persons}. In
  \bibinfo{booktitle}{\emph{Proceedings of ICDAR}}. \bibinfo{pages}{257--261}.
\newblock


\bibitem[\protect\citeauthoryear{Fang, Xie, Chen, Tan, and Zhang}{Fang
  et~al\mbox{.}}{2019}]%
        {fang2019learning}
\bibfield{author}{\bibinfo{person}{Shancheng Fang}, \bibinfo{person}{Hongtao
  Xie}, \bibinfo{person}{Jianjun Chen}, \bibinfo{person}{Jianlong Tan}, {and}
  \bibinfo{person}{Yongdong Zhang}.} \bibinfo{year}{2019}\natexlab{}.
\newblock \showarticletitle{Learning to draw text in natural images with
  conditional adversarial networks}. In \bibinfo{booktitle}{\emph{Proceedings
  of IJCAI}}. \bibinfo{pages}{715--722}.
\newblock


\bibitem[\protect\citeauthoryear{Fang, Xie, Zha, Sun, Tan, and Zhang}{Fang
  et~al\mbox{.}}{2018}]%
        {fang2018attention}
\bibfield{author}{\bibinfo{person}{Shancheng Fang}, \bibinfo{person}{Hongtao
  Xie}, \bibinfo{person}{Zheng-Jun Zha}, \bibinfo{person}{Nannan Sun},
  \bibinfo{person}{Jianlong Tan}, {and} \bibinfo{person}{Yongdong Zhang}.}
  \bibinfo{year}{2018}\natexlab{}.
\newblock \showarticletitle{Attention and language ensemble for scene text
  recognition with convolutional sequence modeling}. In
  \bibinfo{booktitle}{\emph{ACM Multimedia Conference on Multimedia
  Conference}}. \bibinfo{pages}{248--256}.
\newblock


\bibitem[\protect\citeauthoryear{Feng, He, Yin, Zhang, and Liu}{Feng
  et~al\mbox{.}}{2019a}]%
        {feng2019textdragon}
\bibfield{author}{\bibinfo{person}{Wei Feng}, \bibinfo{person}{Wenhao He},
  \bibinfo{person}{Fei Yin}, \bibinfo{person}{Xu-Yao Zhang}, {and}
  \bibinfo{person}{Cheng-Lin Liu}.} \bibinfo{year}{2019}\natexlab{a}.
\newblock \showarticletitle{{TextDragon}: An End-to-End Framework for Arbitrary
  Shaped Text Spotting}. In \bibinfo{booktitle}{\emph{Proceedings of ICCV}}.
  \bibinfo{pages}{9076--9085}.
\newblock


\bibitem[\protect\citeauthoryear{Feng, Yao, and Zhang}{Feng
  et~al\mbox{.}}{2019b}]%
        {feng2019focal}
\bibfield{author}{\bibinfo{person}{Xinjie Feng}, \bibinfo{person}{Hongxun Yao},
  {and} \bibinfo{person}{Shengping Zhang}.} \bibinfo{year}{2019}\natexlab{b}.
\newblock \showarticletitle{Focal {CTC} Loss for Chinese Optical Character
  Recognition on Unbalanced Datasets}.
\newblock \bibinfo{journal}{\emph{Complexity}}  \bibinfo{volume}{2019}
  (\bibinfo{year}{2019}), \bibinfo{pages}{9345861:1--9345861:11}.
\newblock


\bibitem[\protect\citeauthoryear{Gao, Chen, Wang, Tang, and Lu}{Gao
  et~al\mbox{.}}{2018}]%
        {gao2018dense}
\bibfield{author}{\bibinfo{person}{Yunze Gao}, \bibinfo{person}{Yingying Chen},
  \bibinfo{person}{Jinqiao Wang}, \bibinfo{person}{Ming Tang}, {and}
  \bibinfo{person}{Hanqing Lu}.} \bibinfo{year}{2018}\natexlab{}.
\newblock \showarticletitle{Dense Chained Attention Network for Scene Text
  Recognition}. In \bibinfo{booktitle}{\emph{Proceedings of ICIP}}.
  \bibinfo{pages}{679--683}.
\newblock


\bibitem[\protect\citeauthoryear{Gao, Chen, Wang, Tang, and Lu}{Gao
  et~al\mbox{.}}{2019}]%
        {gao2019reading}
\bibfield{author}{\bibinfo{person}{Yunze Gao}, \bibinfo{person}{Yingying Chen},
  \bibinfo{person}{Jinqiao Wang}, \bibinfo{person}{Ming Tang}, {and}
  \bibinfo{person}{Hanqing Lu}.} \bibinfo{year}{2019}\natexlab{}.
\newblock \showarticletitle{Reading scene text with fully convolutional
  sequence modeling}.
\newblock \bibinfo{journal}{\emph{Neurocomputing}}  \bibinfo{volume}{339}
  (\bibinfo{year}{2019}), \bibinfo{pages}{161--170}.
\newblock


\bibitem[\protect\citeauthoryear{Girshick}{Girshick}{2015}]%
        {girshick2015fast}
\bibfield{author}{\bibinfo{person}{Ross Girshick}.}
  \bibinfo{year}{2015}\natexlab{}.
\newblock \showarticletitle{Fast r-cnn}. In
  \bibinfo{booktitle}{\emph{Proceedings of ICCV}}. \bibinfo{pages}{1440--1448}.
\newblock


\bibitem[\protect\citeauthoryear{Goel, Mishra, Alahari, and Jawahar}{Goel
  et~al\mbox{.}}{2013}]%
        {goel2013whole}
\bibfield{author}{\bibinfo{person}{Vibhor Goel}, \bibinfo{person}{Anand
  Mishra}, \bibinfo{person}{Karteek Alahari}, {and} \bibinfo{person}{CV
  Jawahar}.} \bibinfo{year}{2013}\natexlab{}.
\newblock \showarticletitle{Whole is greater than sum of parts: Recognizing
  scene text words}. In \bibinfo{booktitle}{\emph{Proceedings of ICDAR}}.
  \bibinfo{pages}{398--402}.
\newblock


\bibitem[\protect\citeauthoryear{Gomez and Karatzas}{Gomez and
  Karatzas}{2016}]%
        {gomez2016fine}
\bibfield{author}{\bibinfo{person}{Lluis Gomez} {and}
  \bibinfo{person}{Dimosthenis Karatzas}.} \bibinfo{year}{2016}\natexlab{}.
\newblock \showarticletitle{A fine-grained approach to scene text script
  identification}. In \bibinfo{booktitle}{\emph{IAPR Workshop on Document
  Analysis Systems (DAS)}}. \bibinfo{pages}{192--197}.
\newblock


\bibitem[\protect\citeauthoryear{Gomez, Nicolaou, and Karatzas}{Gomez
  et~al\mbox{.}}{2017}]%
        {gomez2017improving}
\bibfield{author}{\bibinfo{person}{Lluis Gomez}, \bibinfo{person}{Anguelos
  Nicolaou}, {and} \bibinfo{person}{Dimosthenis Karatzas}.}
  \bibinfo{year}{2017}\natexlab{}.
\newblock \showarticletitle{Improving patch-based scene text script
  identification with ensembles of conjoined networks}.
\newblock \bibinfo{journal}{\emph{Pattern Recognition}}  \bibinfo{volume}{67}
  (\bibinfo{year}{2017}), \bibinfo{pages}{85--96}.
\newblock


\bibitem[\protect\citeauthoryear{Goodfellow, Pouget-Abadie, Mirza, Xu,
  Warde-Farley, Ozair, Courville, and Bengio}{Goodfellow et~al\mbox{.}}{2014}]%
        {goodfellow2014generative}
\bibfield{author}{\bibinfo{person}{Ian Goodfellow}, \bibinfo{person}{Jean
  Pouget-Abadie}, \bibinfo{person}{Mehdi Mirza}, \bibinfo{person}{Bing Xu},
  \bibinfo{person}{David Warde-Farley}, \bibinfo{person}{Sherjil Ozair},
  \bibinfo{person}{Aaron Courville}, {and} \bibinfo{person}{Yoshua Bengio}.}
  \bibinfo{year}{2014}\natexlab{}.
\newblock \showarticletitle{Generative adversarial nets}. In
  \bibinfo{booktitle}{\emph{Proceedings of NIPS}}. \bibinfo{pages}{2672--2680}.
\newblock


\bibitem[\protect\citeauthoryear{Goodfellow, Warde-Farley, Mirza, Courville,
  and Bengio}{Goodfellow et~al\mbox{.}}{2013}]%
        {goodfellow2013maxout}
\bibfield{author}{\bibinfo{person}{Ian~J Goodfellow}, \bibinfo{person}{David
  Warde-Farley}, \bibinfo{person}{Mehdi Mirza}, \bibinfo{person}{Aaron
  Courville}, {and} \bibinfo{person}{Yoshua Bengio}.}
  \bibinfo{year}{2013}\natexlab{}.
\newblock \showarticletitle{Maxout networks}. In
  \bibinfo{booktitle}{\emph{Proceedings of ICML}}. \bibinfo{pages}{1319--1327}.
\newblock


\bibitem[\protect\citeauthoryear{Gordo}{Gordo}{2015}]%
        {gordo2015supervised}
\bibfield{author}{\bibinfo{person}{Albert Gordo}.}
  \bibinfo{year}{2015}\natexlab{}.
\newblock \showarticletitle{Supervised mid-level features for word image
  representation}. In \bibinfo{booktitle}{\emph{Proceedings of CVPR}}.
  \bibinfo{pages}{2956--2964}.
\newblock


\bibitem[\protect\citeauthoryear{Graves}{Graves}{2012}]%
        {graves2012supervised}
\bibfield{author}{\bibinfo{person}{Alex Graves}.}
  \bibinfo{year}{2012}\natexlab{}.
\newblock \showarticletitle{Supervised sequence labelling}.
\newblock In \bibinfo{booktitle}{\emph{Supervised sequence labelling with
  recurrent neural networks}}. \bibinfo{pages}{5--13}.
\newblock


\bibitem[\protect\citeauthoryear{Graves, Fern{\'a}ndez, Gomez, and
  Schmidhuber}{Graves et~al\mbox{.}}{2006}]%
        {graves2006connectionist}
\bibfield{author}{\bibinfo{person}{Alex Graves}, \bibinfo{person}{Santiago
  Fern{\'a}ndez}, \bibinfo{person}{Faustino Gomez}, {and}
  \bibinfo{person}{J{\"u}rgen Schmidhuber}.} \bibinfo{year}{2006}\natexlab{}.
\newblock \showarticletitle{Connectionist temporal classification: labelling
  unsegmented sequence data with recurrent neural networks}. In
  \bibinfo{booktitle}{\emph{Proceedings of ICML}}. \bibinfo{pages}{369--376}.
\newblock


\bibitem[\protect\citeauthoryear{Graves and Jaitly}{Graves and Jaitly}{2014}]%
        {graves2014towards}
\bibfield{author}{\bibinfo{person}{Alex Graves} {and} \bibinfo{person}{Navdeep
  Jaitly}.} \bibinfo{year}{2014}\natexlab{}.
\newblock \showarticletitle{Towards end-to-end speech recognition with
  recurrent neural networks}. In \bibinfo{booktitle}{\emph{Proceedings of
  ICML}}. \bibinfo{pages}{1764--1772}.
\newblock


\bibitem[\protect\citeauthoryear{Graves, Liwicki, Fern{\'a}ndez, Bertolami,
  Bunke, and Schmidhuber}{Graves et~al\mbox{.}}{2008}]%
        {graves2008novel}
\bibfield{author}{\bibinfo{person}{Alex Graves}, \bibinfo{person}{Marcus
  Liwicki}, \bibinfo{person}{Santiago Fern{\'a}ndez}, \bibinfo{person}{Roman
  Bertolami}, \bibinfo{person}{Horst Bunke}, {and} \bibinfo{person}{J{\"u}rgen
  Schmidhuber}.} \bibinfo{year}{2008}\natexlab{}.
\newblock \showarticletitle{A novel connectionist system for unconstrained
  handwriting recognition}.
\newblock \bibinfo{journal}{\emph{IEEE Trans. Pattern Anal. Mach. Intell}}
  \bibinfo{volume}{31}, \bibinfo{number}{5} (\bibinfo{year}{2008}),
  \bibinfo{pages}{855--868}.
\newblock


\bibitem[\protect\citeauthoryear{Graves, Mohamed, and Hinton}{Graves
  et~al\mbox{.}}{2013}]%
        {graves2013speech}
\bibfield{author}{\bibinfo{person}{Alex Graves}, \bibinfo{person}{Abdel-rahman
  Mohamed}, {and} \bibinfo{person}{Geoffrey Hinton}.}
  \bibinfo{year}{2013}\natexlab{}.
\newblock \showarticletitle{Speech recognition with deep recurrent neural
  networks}. In \bibinfo{booktitle}{\emph{Proceedings of ICASSP}}.
  \bibinfo{pages}{6645--6649}.
\newblock


\bibitem[\protect\citeauthoryear{Guo, Wang, Lei, Tu, and Li}{Guo
  et~al\mbox{.}}{2016}]%
        {guo2016convolutional}
\bibfield{author}{\bibinfo{person}{Qiang Guo}, \bibinfo{person}{Fenglei Wang},
  \bibinfo{person}{Jun Lei}, \bibinfo{person}{Dan Tu}, {and}
  \bibinfo{person}{Guohui Li}.} \bibinfo{year}{2016}\natexlab{}.
\newblock \showarticletitle{Convolutional feature learning and Hybrid {CNN-HMM}
  for scene number recognition}.
\newblock \bibinfo{journal}{\emph{Neurocomputing}}  \bibinfo{volume}{184}
  (\bibinfo{year}{2016}), \bibinfo{pages}{78--90}.
\newblock


\bibitem[\protect\citeauthoryear{Gupta, Vedaldi, and Zisserman}{Gupta
  et~al\mbox{.}}{2016}]%
        {gupta2016synthetic}
\bibfield{author}{\bibinfo{person}{Ankush Gupta}, \bibinfo{person}{Andrea
  Vedaldi}, {and} \bibinfo{person}{Andrew Zisserman}.}
  \bibinfo{year}{2016}\natexlab{}.
\newblock \showarticletitle{Synthetic data for text localisation in natural
  images}. In \bibinfo{booktitle}{\emph{Proceedings of CVPR}}.
  \bibinfo{pages}{2315--2324}.
\newblock


\bibitem[\protect\citeauthoryear{Ham, Kang, Chung, Park, and Park}{Ham
  et~al\mbox{.}}{1995}]%
        {ham1995recognition}
\bibfield{author}{\bibinfo{person}{Young~Kug Ham}, \bibinfo{person}{Min~Seok
  Kang}, \bibinfo{person}{Hong~Kyu Chung}, \bibinfo{person}{Rae-Hong Park},
  {and} \bibinfo{person}{Gwi~Tae Park}.} \bibinfo{year}{1995}\natexlab{}.
\newblock \showarticletitle{Recognition of raised characters for automatic
  classification of rubber tires}.
\newblock \bibinfo{journal}{\emph{Optical Engineering}} \bibinfo{volume}{34},
  \bibinfo{number}{1} (\bibinfo{year}{1995}), \bibinfo{pages}{102--110}.
\newblock


\bibitem[\protect\citeauthoryear{He, Yang, Liang, Zhou, Ororbi, Kifer, and
  Lee~Giles}{He et~al\mbox{.}}{2017c}]%
        {he2017multi}
\bibfield{author}{\bibinfo{person}{Dafang He}, \bibinfo{person}{Xiao Yang},
  \bibinfo{person}{Chen Liang}, \bibinfo{person}{Zihan Zhou},
  \bibinfo{person}{Alexander~G Ororbi}, \bibinfo{person}{Daniel Kifer}, {and}
  \bibinfo{person}{C Lee~Giles}.} \bibinfo{year}{2017}\natexlab{c}.
\newblock \showarticletitle{Multi-scale {FCN} with cascaded instance aware
  segmentation for arbitrary oriented word spotting in the wild}. In
  \bibinfo{booktitle}{\emph{Proceedings of CVPR}}. \bibinfo{pages}{3519--3528}.
\newblock


\bibitem[\protect\citeauthoryear{He, Gkioxari, Doll{\'a}r, and Girshick}{He
  et~al\mbox{.}}{2017a}]%
        {he2017mask}
\bibfield{author}{\bibinfo{person}{Kaiming He}, \bibinfo{person}{Georgia
  Gkioxari}, \bibinfo{person}{Piotr Doll{\'a}r}, {and} \bibinfo{person}{Ross
  Girshick}.} \bibinfo{year}{2017}\natexlab{a}.
\newblock \showarticletitle{Mask r-cnn}. In
  \bibinfo{booktitle}{\emph{Proceedings of ICCV}}. \bibinfo{pages}{2961--2969}.
\newblock


\bibitem[\protect\citeauthoryear{He, Zhang, Ren, and Sun}{He
  et~al\mbox{.}}{2016b}]%
        {he2016deep}
\bibfield{author}{\bibinfo{person}{Kaiming He}, \bibinfo{person}{Xiangyu
  Zhang}, \bibinfo{person}{Shaoqing Ren}, {and} \bibinfo{person}{Jian Sun}.}
  \bibinfo{year}{2016}\natexlab{b}.
\newblock \showarticletitle{Deep residual learning for image recognition}. In
  \bibinfo{booktitle}{\emph{Proceedings of CVPR}}. \bibinfo{pages}{770--778}.
\newblock


\bibitem[\protect\citeauthoryear{He, Liu, Yang, Zhang, Luo, Gao, Zheng, Wang,
  Zhang, and Jin}{He et~al\mbox{.}}{2018a}]%
        {he2018icpr2018}
\bibfield{author}{\bibinfo{person}{Mengchao He}, \bibinfo{person}{Yuliang Liu},
  \bibinfo{person}{Zhibo Yang}, \bibinfo{person}{Sheng Zhang},
  \bibinfo{person}{Canjie Luo}, \bibinfo{person}{Feiyu Gao},
  \bibinfo{person}{Qi Zheng}, \bibinfo{person}{Yongpan Wang},
  \bibinfo{person}{Xin Zhang}, {and} \bibinfo{person}{Lianwen Jin}.}
  \bibinfo{year}{2018}\natexlab{a}.
\newblock \showarticletitle{{ICPR}2018 Contest on Robust Reading for Multi-Type
  Web Images}. In \bibinfo{booktitle}{\emph{Proceedings of ICPR}}.
  \bibinfo{pages}{7--12}.
\newblock


\bibitem[\protect\citeauthoryear{He, Huang, He, Zhu, Qiao, and Li}{He
  et~al\mbox{.}}{2017b}]%
        {he2017single}
\bibfield{author}{\bibinfo{person}{Pan He}, \bibinfo{person}{Weilin Huang},
  \bibinfo{person}{Tong He}, \bibinfo{person}{Qile Zhu}, \bibinfo{person}{Yu
  Qiao}, {and} \bibinfo{person}{Xiaolin Li}.} \bibinfo{year}{2017}\natexlab{b}.
\newblock \showarticletitle{Single shot text detector with regional attention}.
  In \bibinfo{booktitle}{\emph{Proceedings of ICCV}}.
  \bibinfo{pages}{3047--3055}.
\newblock


\bibitem[\protect\citeauthoryear{He, Huang, Qiao, Loy, and Tang}{He
  et~al\mbox{.}}{2016a}]%
        {He2016reading}
\bibfield{author}{\bibinfo{person}{Pan He}, \bibinfo{person}{Weilin Huang},
  \bibinfo{person}{Yu Qiao}, \bibinfo{person}{Chen~Change Loy}, {and}
  \bibinfo{person}{Xiaoou Tang}.} \bibinfo{year}{2016}\natexlab{a}.
\newblock \showarticletitle{Reading Scene Text in Deep Convolutional
  Sequences}. In \bibinfo{booktitle}{\emph{Proceedings of AAAI}}.
  \bibinfo{pages}{3501--3508}.
\newblock


\bibitem[\protect\citeauthoryear{He, Tian, Huang, Shen, Qiao, and Sun}{He
  et~al\mbox{.}}{2018b}]%
        {he2018end}
\bibfield{author}{\bibinfo{person}{Tong He}, \bibinfo{person}{Zhi Tian},
  \bibinfo{person}{Weilin Huang}, \bibinfo{person}{Chunhua Shen},
  \bibinfo{person}{Yu Qiao}, {and} \bibinfo{person}{Changming Sun}.}
  \bibinfo{year}{2018}\natexlab{b}.
\newblock \showarticletitle{An end-to-end textspotter with explicit alignment
  and attention}. In \bibinfo{booktitle}{\emph{Proceedings of CVPR}}.
  \bibinfo{pages}{5020--5029}.
\newblock


\bibitem[\protect\citeauthoryear{He, Yang, Shi, and Bai}{He
  et~al\mbox{.}}{2019}]%
        {he2019vd}
\bibfield{author}{\bibinfo{person}{Xinwei He}, \bibinfo{person}{Yang Yang},
  \bibinfo{person}{Baoguang Shi}, {and} \bibinfo{person}{Xiang Bai}.}
  \bibinfo{year}{2019}\natexlab{}.
\newblock \showarticletitle{{VD-SAN}: Visual-Densely Semantic Attention Network
  for Image Caption Generation}.
\newblock \bibinfo{journal}{\emph{Neurocomputing}}  \bibinfo{volume}{328}
  (\bibinfo{year}{2019}), \bibinfo{pages}{48--55}.
\newblock


\bibitem[\protect\citeauthoryear{Hochreiter and Schmidhuber}{Hochreiter and
  Schmidhuber}{1997}]%
        {hochreiter1997long}
\bibfield{author}{\bibinfo{person}{Sepp Hochreiter} {and}
  \bibinfo{person}{J{\"u}rgen Schmidhuber}.} \bibinfo{year}{1997}\natexlab{}.
\newblock \showarticletitle{Long short-term memory}.
\newblock \bibinfo{journal}{\emph{Neural computation}} \bibinfo{volume}{9},
  \bibinfo{number}{8} (\bibinfo{year}{1997}), \bibinfo{pages}{1735--1780}.
\newblock


\bibitem[\protect\citeauthoryear{Hu, Cai, Hou, Yi, and Lin}{Hu
  et~al\mbox{.}}{2020}]%
        {hu2020gtc}
\bibfield{author}{\bibinfo{person}{Wenyang Hu}, \bibinfo{person}{Xiaocong Cai},
  \bibinfo{person}{Jun Hou}, \bibinfo{person}{Shuai Yi}, {and}
  \bibinfo{person}{Zhiping Lin}.} \bibinfo{year}{2020}\natexlab{}.
\newblock \showarticletitle{{GTC}: Guided Training of CTC Towards Efficient and
  Accurate Scene Text Recognition}. In \bibinfo{booktitle}{\emph{Proceedings of
  AAAI}}.
\newblock


\bibitem[\protect\citeauthoryear{Huang, Liu, Van Der~Maaten, and
  Weinberger}{Huang et~al\mbox{.}}{2017}]%
        {huang2017densely}
\bibfield{author}{\bibinfo{person}{Gao Huang}, \bibinfo{person}{Zhuang Liu},
  \bibinfo{person}{Laurens Van Der~Maaten}, {and} \bibinfo{person}{Kilian~Q
  Weinberger}.} \bibinfo{year}{2017}\natexlab{}.
\newblock \showarticletitle{Densely connected convolutional networks}. In
  \bibinfo{booktitle}{\emph{Proceedings of CVPR}}. \bibinfo{pages}{4700--4708}.
\newblock


\bibitem[\protect\citeauthoryear{Huang, Zhong, Yin, Xiang, He, Lv, and
  Huang}{Huang et~al\mbox{.}}{2019}]%
        {huang2019express}
\bibfield{author}{\bibinfo{person}{Hu Huang}, \bibinfo{person}{Ya Zhong},
  \bibinfo{person}{Shiying Yin}, \bibinfo{person}{Junlin Xiang},
  \bibinfo{person}{Lijun He}, \bibinfo{person}{Yu Lv}, {and}
  \bibinfo{person}{Peng Huang}.} \bibinfo{year}{2019}\natexlab{}.
\newblock \showarticletitle{Express Delivery System based on Fingerprint
  Identification}. In \bibinfo{booktitle}{\emph{Proceedings of ITNEC}}. IEEE,
  \bibinfo{pages}{363--367}.
\newblock


\bibitem[\protect\citeauthoryear{Huang, Sun, Jin, and Luo}{Huang
  et~al\mbox{.}}{2020}]%
        {huang2019epan}
\bibfield{author}{\bibinfo{person}{Yunlong Huang}, \bibinfo{person}{Zenghui
  Sun}, \bibinfo{person}{Lianwen Jin}, {and} \bibinfo{person}{Canjie Luo}.}
  \bibinfo{year}{2020}\natexlab{}.
\newblock \showarticletitle{{EPAN}: Effective parts attention network for scene
  text recognition}.
\newblock \bibinfo{journal}{\emph{Neurocomputing}}  \bibinfo{volume}{376}
  (\bibinfo{year}{2020}), \bibinfo{pages}{202--213}.
\newblock


\bibitem[\protect\citeauthoryear{Jaderberg, Simonyan, Vedaldi, and
  Zisserman}{Jaderberg et~al\mbox{.}}{2014a}]%
        {jaderberg2014synthetic}
\bibfield{author}{\bibinfo{person}{Max Jaderberg}, \bibinfo{person}{Karen
  Simonyan}, \bibinfo{person}{Andrea Vedaldi}, {and} \bibinfo{person}{Andrew
  Zisserman}.} \bibinfo{year}{2014}\natexlab{a}.
\newblock \showarticletitle{Synthetic data and artificial neural networks for
  natural scene text recognition}. In \bibinfo{booktitle}{\emph{Proceedings of
  NIPS-W}}.
\newblock


\bibitem[\protect\citeauthoryear{Jaderberg, Simonyan, Vedaldi, and
  Zisserman}{Jaderberg et~al\mbox{.}}{2016}]%
        {jaderberg2016reading}
\bibfield{author}{\bibinfo{person}{Max Jaderberg}, \bibinfo{person}{Karen
  Simonyan}, \bibinfo{person}{Andrea Vedaldi}, {and} \bibinfo{person}{Andrew
  Zisserman}.} \bibinfo{year}{2016}\natexlab{}.
\newblock \showarticletitle{Reading text in the wild with convolutional neural
  networks}.
\newblock \bibinfo{journal}{\emph{Int. J. Comput. Vis}} \bibinfo{volume}{116},
  \bibinfo{number}{1} (\bibinfo{year}{2016}), \bibinfo{pages}{1--20}.
\newblock


\bibitem[\protect\citeauthoryear{Jaderberg, Simonyan, and Zisserman}{Jaderberg
  et~al\mbox{.}}{2015a}]%
        {jaderberg2015deep}
\bibfield{author}{\bibinfo{person}{Max Jaderberg}, \bibinfo{person}{Karen
  Simonyan}, {and} \bibinfo{person}{Andrew Zisserman}.}
  \bibinfo{year}{2015}\natexlab{a}.
\newblock \showarticletitle{Deep structured output learning for unconstrained
  text recognition}. In \bibinfo{booktitle}{\emph{Proceedings of ICLR}}.
\newblock


\bibitem[\protect\citeauthoryear{Jaderberg, Simonyan, Zisserman,
  et~al\mbox{.}}{Jaderberg et~al\mbox{.}}{2015b}]%
        {jaderberg2015spatial}
\bibfield{author}{\bibinfo{person}{Max Jaderberg}, \bibinfo{person}{Karen
  Simonyan}, \bibinfo{person}{Andrew Zisserman}, {et~al\mbox{.}}}
  \bibinfo{year}{2015}\natexlab{b}.
\newblock \showarticletitle{Spatial transformer networks}. In
  \bibinfo{booktitle}{\emph{Proceedings of NIPS}}. \bibinfo{pages}{2017--2025}.
\newblock


\bibitem[\protect\citeauthoryear{Jaderberg, Vedaldi, and Zisserman}{Jaderberg
  et~al\mbox{.}}{2014b}]%
        {jaderberg2014deep}
\bibfield{author}{\bibinfo{person}{Max Jaderberg}, \bibinfo{person}{Andrea
  Vedaldi}, {and} \bibinfo{person}{Andrew Zisserman}.}
  \bibinfo{year}{2014}\natexlab{b}.
\newblock \showarticletitle{Deep features for text spotting}. In
  \bibinfo{booktitle}{\emph{Proceedings of ECCV}}. \bibinfo{pages}{512--528}.
\newblock


\bibitem[\protect\citeauthoryear{Karatzas, Gomez-Bigorda, Nicolaou, Ghosh,
  Bagdanov, Iwamura, Matas, Neumann, Chandrasekhar, Lu, et~al\mbox{.}}{Karatzas
  et~al\mbox{.}}{2015}]%
        {karatzas2015icdar}
\bibfield{author}{\bibinfo{person}{Dimosthenis Karatzas},
  \bibinfo{person}{Lluis Gomez-Bigorda}, \bibinfo{person}{Anguelos Nicolaou},
  \bibinfo{person}{Suman Ghosh}, \bibinfo{person}{Andrew Bagdanov},
  \bibinfo{person}{Masakazu Iwamura}, \bibinfo{person}{Jiri Matas},
  \bibinfo{person}{Lukas Neumann}, \bibinfo{person}{Vijay~Ramaseshan
  Chandrasekhar}, \bibinfo{person}{Shijian Lu}, {et~al\mbox{.}}}
  \bibinfo{year}{2015}\natexlab{}.
\newblock \showarticletitle{{ICDAR} 2015 competition on robust reading}. In
  \bibinfo{booktitle}{\emph{Proceedings of ICDAR}}.
  \bibinfo{pages}{1156--1160}.
\newblock


\bibitem[\protect\citeauthoryear{Karatzas, Shafait, Uchida, Iwamura, i~Bigorda,
  Mestre, Mas, Mota, Almazan, and De~Las~Heras}{Karatzas et~al\mbox{.}}{2013}]%
        {karatzas2013icdar}
\bibfield{author}{\bibinfo{person}{Dimosthenis Karatzas},
  \bibinfo{person}{Faisal Shafait}, \bibinfo{person}{Seiichi Uchida},
  \bibinfo{person}{Masakazu Iwamura}, \bibinfo{person}{Lluis~Gomez i Bigorda},
  \bibinfo{person}{Sergi~Robles Mestre}, \bibinfo{person}{Joan Mas},
  \bibinfo{person}{David~Fernandez Mota}, \bibinfo{person}{Jon~Almazan
  Almazan}, {and} \bibinfo{person}{Lluis~Pere De~Las~Heras}.}
  \bibinfo{year}{2013}\natexlab{}.
\newblock \showarticletitle{{ICDAR} 2013 robust reading competition}. In
  \bibinfo{booktitle}{\emph{Proceedings of ICDAR}}.
  \bibinfo{pages}{1484--1493}.
\newblock


\bibitem[\protect\citeauthoryear{Katti, Reisswig, Guder, Brarda, Bickel,
  H{\"o}hne, and Faddoul}{Katti et~al\mbox{.}}{2018}]%
        {katti2018chargrid}
\bibfield{author}{\bibinfo{person}{Anoop~Raveendra Katti},
  \bibinfo{person}{Christian Reisswig}, \bibinfo{person}{Cordula Guder},
  \bibinfo{person}{Sebastian Brarda}, \bibinfo{person}{Steffen Bickel},
  \bibinfo{person}{Johannes H{\"o}hne}, {and} \bibinfo{person}{Jean~Baptiste
  Faddoul}.} \bibinfo{year}{2018}\natexlab{}.
\newblock \showarticletitle{Chargrid: Towards understanding 2d documents}. In
  \bibinfo{booktitle}{\emph{Proceedings of EMNLP}}.
  \bibinfo{pages}{4459--4469}.
\newblock


\bibitem[\protect\citeauthoryear{Kim and Kim}{Kim and Kim}{2008}]%
        {kim2008new}
\bibfield{author}{\bibinfo{person}{Wonjun Kim} {and} \bibinfo{person}{Changick
  Kim}.} \bibinfo{year}{2008}\natexlab{}.
\newblock \showarticletitle{A new approach for overlay text detection and
  extraction from complex video scene}.
\newblock \bibinfo{journal}{\emph{IEEE Transactions on Image Processing}}
  \bibinfo{volume}{18}, \bibinfo{number}{2} (\bibinfo{year}{2008}),
  \bibinfo{pages}{401--411}.
\newblock


\bibitem[\protect\citeauthoryear{Kipf and Welling}{Kipf and Welling}{2017}]%
        {kipf2016semi}
\bibfield{author}{\bibinfo{person}{Thomas~N Kipf} {and} \bibinfo{person}{Max
  Welling}.} \bibinfo{year}{2017}\natexlab{}.
\newblock \showarticletitle{Semi-supervised classification with graph
  convolutional networks}. In \bibinfo{booktitle}{\emph{Proceedings of ICLR}}.
\newblock


\bibitem[\protect\citeauthoryear{Koo and Kim}{Koo and Kim}{2013}]%
        {Hyung2013text}
\bibfield{author}{\bibinfo{person}{Hyung~Il Koo} {and}
  \bibinfo{person}{Duck~Hoon Kim}.} \bibinfo{year}{2013}\natexlab{}.
\newblock \showarticletitle{Scene Text Detection via Connected Component
  Clustering and Nontext Filtering}.
\newblock \bibinfo{journal}{\emph{IEEE Transactions on Image Processing}}
  \bibinfo{volume}{22}, \bibinfo{number}{6} (\bibinfo{year}{2013}),
  \bibinfo{pages}{2296--2305}.
\newblock


\bibitem[\protect\citeauthoryear{Krasin, Duerig, Alldrin, Ferrari,
  Abu-El-Haija, Kuznetsova, Rom, Uijlings, Popov, Veit, et~al\mbox{.}}{Krasin
  et~al\mbox{.}}{2017}]%
        {krasin2017openimages}
\bibfield{author}{\bibinfo{person}{Ivan Krasin}, \bibinfo{person}{Tom Duerig},
  \bibinfo{person}{Neil Alldrin}, \bibinfo{person}{Vittorio Ferrari},
  \bibinfo{person}{Sami Abu-El-Haija}, \bibinfo{person}{Alina Kuznetsova},
  \bibinfo{person}{Hassan Rom}, \bibinfo{person}{Jasper Uijlings},
  \bibinfo{person}{Stefan Popov}, \bibinfo{person}{Andreas Veit},
  {et~al\mbox{.}}} \bibinfo{year}{2017}\natexlab{}.
\newblock \showarticletitle{Openimages: A public dataset for large-scale
  multi-label and multi-class image classification}.
\newblock \bibinfo{journal}{\emph{Dataset available from https://github.
  com/openimages}}  \bibinfo{volume}{2} (\bibinfo{year}{2017}),
  \bibinfo{pages}{3}.
\newblock


\bibitem[\protect\citeauthoryear{Lafferty, McCallum, and Pereira}{Lafferty
  et~al\mbox{.}}{2001}]%
        {LaffertyMP01}
\bibfield{author}{\bibinfo{person}{John~D. Lafferty}, \bibinfo{person}{Andrew
  McCallum}, {and} \bibinfo{person}{Fernando C.~N. Pereira}.}
  \bibinfo{year}{2001}\natexlab{}.
\newblock \showarticletitle{Conditional Random Fields: Probabilistic Models for
  Segmenting and Labeling Sequence Data}. In
  \bibinfo{booktitle}{\emph{Proceedings of ICML}}. \bibinfo{pages}{282--289}.
\newblock


\bibitem[\protect\citeauthoryear{LeCun, Bottou, Bengio, Haffner,
  et~al\mbox{.}}{LeCun et~al\mbox{.}}{1998}]%
        {lecun1998gradient}
\bibfield{author}{\bibinfo{person}{Yann LeCun}, \bibinfo{person}{L{\'e}on
  Bottou}, \bibinfo{person}{Yoshua Bengio}, \bibinfo{person}{Patrick Haffner},
  {et~al\mbox{.}}} \bibinfo{year}{1998}\natexlab{}.
\newblock \showarticletitle{Gradient-based learning applied to document
  recognition}.
\newblock \bibinfo{journal}{\emph{Proc. IEEE}} \bibinfo{volume}{86},
  \bibinfo{number}{11} (\bibinfo{year}{1998}), \bibinfo{pages}{2278--2324}.
\newblock


\bibitem[\protect\citeauthoryear{Lee and Osindero}{Lee and Osindero}{2016}]%
        {lee2016recursive}
\bibfield{author}{\bibinfo{person}{Chen-Yu Lee} {and} \bibinfo{person}{Simon
  Osindero}.} \bibinfo{year}{2016}\natexlab{}.
\newblock \showarticletitle{Recursive recurrent nets with attention modeling
  for {OCR} in the wild}. In \bibinfo{booktitle}{\emph{Proceedings of CVPR}}.
  \bibinfo{pages}{2231--2239}.
\newblock


\bibitem[\protect\citeauthoryear{Lee, Cho, Jung, and Kim}{Lee
  et~al\mbox{.}}{2010}]%
        {lee2010scene}
\bibfield{author}{\bibinfo{person}{SeongHun Lee}, \bibinfo{person}{Min~Su Cho},
  \bibinfo{person}{Kyomin Jung}, {and} \bibinfo{person}{Jin~Hyung Kim}.}
  \bibinfo{year}{2010}\natexlab{}.
\newblock \showarticletitle{Scene text extraction with edge constraint and text
  collinearity}. In \bibinfo{booktitle}{\emph{Proceedings of ICPR}}.
  \bibinfo{pages}{3983--3986}.
\newblock


\bibitem[\protect\citeauthoryear{Li, Wang, and Shen}{Li et~al\mbox{.}}{2017}]%
        {li2017towards}
\bibfield{author}{\bibinfo{person}{Hui Li}, \bibinfo{person}{Peng Wang}, {and}
  \bibinfo{person}{Chunhua Shen}.} \bibinfo{year}{2017}\natexlab{}.
\newblock \showarticletitle{Towards end-to-end text spotting with convolutional
  recurrent neural networks}. In \bibinfo{booktitle}{\emph{Proceedings of
  ICCV}}. \bibinfo{pages}{5238--5246}.
\newblock


\bibitem[\protect\citeauthoryear{Li, Wang, Shen, and Zhang}{Li
  et~al\mbox{.}}{2019}]%
        {li2019show}
\bibfield{author}{\bibinfo{person}{Hui Li}, \bibinfo{person}{Peng Wang},
  \bibinfo{person}{Chunhua Shen}, {and} \bibinfo{person}{Guyu Zhang}.}
  \bibinfo{year}{2019}\natexlab{}.
\newblock \showarticletitle{Show, attend and read: A simple and strong baseline
  for irregular text recognition}. In \bibinfo{booktitle}{\emph{Proceedings of
  AAAI}}. \bibinfo{pages}{8610--8617}.
\newblock


\bibitem[\protect\citeauthoryear{Li and Wang}{Li and Wang}{2008}]%
        {li2008adaptive}
\bibfield{author}{\bibinfo{person}{Minhua Li} {and} \bibinfo{person}{Chunheng
  Wang}.} \bibinfo{year}{2008}\natexlab{}.
\newblock \showarticletitle{An adaptive text detection approach in images and
  video frames}. In \bibinfo{booktitle}{\emph{Proceedings of IJCNN}}.
  \bibinfo{pages}{72--77}.
\newblock


\bibitem[\protect\citeauthoryear{Li, Wang, Li, and Wu}{Li
  et~al\mbox{.}}{2018}]%
        {li2018employing}
\bibfield{author}{\bibinfo{person}{Peipei Li}, \bibinfo{person}{Haixun Wang},
  \bibinfo{person}{Hongsong Li}, {and} \bibinfo{person}{Xindong Wu}.}
  \bibinfo{year}{2018}\natexlab{}.
\newblock \showarticletitle{Employing Semantic Context for Sparse Information
  Extraction Assessment}.
\newblock \bibinfo{journal}{\emph{ACM Transactions on Knowledge Discovery from
  Data (TKDD)}} \bibinfo{volume}{12}, \bibinfo{number}{5}
  (\bibinfo{year}{2018}), \bibinfo{pages}{54}.
\newblock


\bibitem[\protect\citeauthoryear{Liang and Hu}{Liang and Hu}{2015}]%
        {liang2015recurrent}
\bibfield{author}{\bibinfo{person}{Ming Liang} {and} \bibinfo{person}{Xiaolin
  Hu}.} \bibinfo{year}{2015}\natexlab{}.
\newblock \showarticletitle{Recurrent convolutional neural network for object
  recognition}. In \bibinfo{booktitle}{\emph{Proceedings of CVPR}}.
  \bibinfo{pages}{3367--3375}.
\newblock


\bibitem[\protect\citeauthoryear{Liao, Lyu, He, Yao, Wu, and Bai}{Liao
  et~al\mbox{.}}{2019a}]%
        {liao2019mask}
\bibfield{author}{\bibinfo{person}{Minghui Liao}, \bibinfo{person}{Pengyuan
  Lyu}, \bibinfo{person}{Minghang He}, \bibinfo{person}{Cong Yao},
  \bibinfo{person}{Wenhao Wu}, {and} \bibinfo{person}{Xiang Bai}.}
  \bibinfo{year}{2019}\natexlab{a}.
\newblock \showarticletitle{Mask textspotter: An end-to-end trainable neural
  network for spotting text with arbitrary shapes}.
\newblock \bibinfo{journal}{\emph{IEEE Trans. Pattern Anal. Mach. Intell}}
  (\bibinfo{year}{2019}).
\newblock


\bibitem[\protect\citeauthoryear{Liao, Shi, and Bai}{Liao
  et~al\mbox{.}}{2018}]%
        {liao2018textboxes++}
\bibfield{author}{\bibinfo{person}{Minghui Liao}, \bibinfo{person}{Baoguang
  Shi}, {and} \bibinfo{person}{Xiang Bai}.} \bibinfo{year}{2018}\natexlab{}.
\newblock \showarticletitle{Textboxes++: A single-shot oriented scene text
  detector}.
\newblock \bibinfo{journal}{\emph{IEEE Transactions on Image Processing}}
  \bibinfo{volume}{27}, \bibinfo{number}{8} (\bibinfo{year}{2018}),
  \bibinfo{pages}{3676--3690}.
\newblock


\bibitem[\protect\citeauthoryear{Liao, Shi, Bai, Wang, and Liu}{Liao
  et~al\mbox{.}}{2017}]%
        {liao2017textboxes}
\bibfield{author}{\bibinfo{person}{Minghui Liao}, \bibinfo{person}{Baoguang
  Shi}, \bibinfo{person}{Xiang Bai}, \bibinfo{person}{Xinggang Wang}, {and}
  \bibinfo{person}{Wenyu Liu}.} \bibinfo{year}{2017}\natexlab{}.
\newblock \showarticletitle{Textboxes: A fast text detector with a single deep
  neural network}. In \bibinfo{booktitle}{\emph{Proceedings of AAAI}}.
  \bibinfo{pages}{4161--4167}.
\newblock


\bibitem[\protect\citeauthoryear{Liao, Zhang, Wan, Xie, Liang, Lyu, Yao, and
  Bai}{Liao et~al\mbox{.}}{2019b}]%
        {liao2019scene}
\bibfield{author}{\bibinfo{person}{Minghui Liao}, \bibinfo{person}{Jian Zhang},
  \bibinfo{person}{Zhaoyi Wan}, \bibinfo{person}{Fengming Xie},
  \bibinfo{person}{Jiajun Liang}, \bibinfo{person}{Pengyuan Lyu},
  \bibinfo{person}{Cong Yao}, {and} \bibinfo{person}{Xiang Bai}.}
  \bibinfo{year}{2019}\natexlab{b}.
\newblock \showarticletitle{Scene text recognition from two-dimensional
  perspective}. In \bibinfo{booktitle}{\emph{Proceedings of AAAI}}.
  \bibinfo{pages}{8714--8721}.
\newblock


\bibitem[\protect\citeauthoryear{Lienhart and Wernicke}{Lienhart and
  Wernicke}{2002}]%
        {lienhart2002localizing}
\bibfield{author}{\bibinfo{person}{Rainer Lienhart} {and} \bibinfo{person}{Axel
  Wernicke}.} \bibinfo{year}{2002}\natexlab{}.
\newblock \showarticletitle{Localizing and segmenting text in images and
  videos}.
\newblock \bibinfo{journal}{\emph{IEEE Transactions on circuits and systems for
  video technology}} \bibinfo{volume}{12}, \bibinfo{number}{4}
  (\bibinfo{year}{2002}), \bibinfo{pages}{256--268}.
\newblock


\bibitem[\protect\citeauthoryear{Litman, Anschel, Tsiper, Litman, Mazor, and
  Manmatha}{Litman et~al\mbox{.}}{2020}]%
        {Ron2020Scatter}
\bibfield{author}{\bibinfo{person}{Ron Litman}, \bibinfo{person}{Oron Anschel},
  \bibinfo{person}{Shahar Tsiper}, \bibinfo{person}{Roee Litman},
  \bibinfo{person}{Shai Mazor}, {and} \bibinfo{person}{R. Manmatha}.}
  \bibinfo{year}{2020}\natexlab{}.
\newblock \showarticletitle{{SCATTER:} Selective Context Attentional Scene Text
  Recognizer}. In \bibinfo{booktitle}{\emph{Proceedings of CVPR}}.
\newblock


\bibitem[\protect\citeauthoryear{Liu, Koga, and Fujisawa}{Liu
  et~al\mbox{.}}{2002}]%
        {liu2002lexicon}
\bibfield{author}{\bibinfo{person}{Cheng-Lin Liu}, \bibinfo{person}{Masashi
  Koga}, {and} \bibinfo{person}{Hiromichi Fujisawa}.}
  \bibinfo{year}{2002}\natexlab{}.
\newblock \showarticletitle{Lexicon-driven segmentation and recognition of
  handwritten character strings for Japanese address reading}.
\newblock \bibinfo{journal}{\emph{IEEE Trans. Pattern Anal. Mach. Intell}}
  \bibinfo{volume}{24}, \bibinfo{number}{11} (\bibinfo{year}{2002}),
  \bibinfo{pages}{1425--1437}.
\newblock


\bibitem[\protect\citeauthoryear{Liu, Flanigan, Thomson, Sadeh, and Smith}{Liu
  et~al\mbox{.}}{2018b}]%
        {liu2018toward}
\bibfield{author}{\bibinfo{person}{Fei Liu}, \bibinfo{person}{Jeffrey
  Flanigan}, \bibinfo{person}{Sam Thomson}, \bibinfo{person}{Norman Sadeh},
  {and} \bibinfo{person}{Noah~A Smith}.} \bibinfo{year}{2018}\natexlab{b}.
\newblock \showarticletitle{Toward abstractive summarization using semantic
  representations}.
\newblock \bibinfo{journal}{\emph{CoRR abs/1805.10399}} (\bibinfo{year}{2018}).
\newblock


\bibitem[\protect\citeauthoryear{Liu, Jin, and Zhang}{Liu
  et~al\mbox{.}}{2018c}]%
        {liu2018connectionist}
\bibfield{author}{\bibinfo{person}{Hu Liu}, \bibinfo{person}{Sheng Jin}, {and}
  \bibinfo{person}{Changshui Zhang}.} \bibinfo{year}{2018}\natexlab{c}.
\newblock \showarticletitle{Connectionist temporal classification with maximum
  entropy regularization}. In \bibinfo{booktitle}{\emph{Proceedings of NIPS}}.
  \bibinfo{pages}{831--841}.
\newblock


\bibitem[\protect\citeauthoryear{Liu, Chen, and Wong}{Liu
  et~al\mbox{.}}{2018a}]%
        {liu2018char}
\bibfield{author}{\bibinfo{person}{Wei Liu}, \bibinfo{person}{Chaofeng Chen},
  {and} \bibinfo{person}{Kwan-Yee~K Wong}.} \bibinfo{year}{2018}\natexlab{a}.
\newblock \showarticletitle{{Char-Net}: A Character-Aware Neural Network for
  Distorted Scene Text Recognition.}. In \bibinfo{booktitle}{\emph{Proceedings
  of AAAI}}. \bibinfo{pages}{7154--7161}.
\newblock


\bibitem[\protect\citeauthoryear{Liu, Chen, Wong, Su, and Han}{Liu
  et~al\mbox{.}}{2016a}]%
        {liu2016star}
\bibfield{author}{\bibinfo{person}{Wei Liu}, \bibinfo{person}{Chaofeng Chen},
  \bibinfo{person}{Kwan-Yee~K Wong}, \bibinfo{person}{Zhizhong Su}, {and}
  \bibinfo{person}{Junyu Han}.} \bibinfo{year}{2016}\natexlab{a}.
\newblock \showarticletitle{{STAR-Net}: A SpaTial Attention Residue Network for
  Scene Text Recognition}. In \bibinfo{booktitle}{\emph{Proceedings of BMVC}}.
  \bibinfo{pages}{7}.
\newblock


\bibitem[\protect\citeauthoryear{Liu}{Liu}{2008}]%
        {liu2008camera}
\bibfield{author}{\bibinfo{person}{Xu Liu}.} \bibinfo{year}{2008}\natexlab{}.
\newblock \showarticletitle{A camera phone based currency reader for the
  visually impaired}. In \bibinfo{booktitle}{\emph{Proceedings of {ACM}
  {SIGACCESS} International Conference on Computers and Accessibility}}.
  \bibinfo{pages}{305--306}.
\newblock


\bibitem[\protect\citeauthoryear{Liu, Gao, Zhang, and Zhao}{Liu
  et~al\mbox{.}}{2019a}]%
        {liu2019graph}
\bibfield{author}{\bibinfo{person}{Xiaojing Liu}, \bibinfo{person}{Feiyu Gao},
  \bibinfo{person}{Qiong Zhang}, {and} \bibinfo{person}{Huasha Zhao}.}
  \bibinfo{year}{2019}\natexlab{a}.
\newblock \showarticletitle{Graph convolution for multimodal information
  extraction from visually rich documents}. In
  \bibinfo{booktitle}{\emph{Proceedings of NAACL}}. \bibinfo{pages}{32--39}.
\newblock


\bibitem[\protect\citeauthoryear{Liu, Kawanishi, Wu, and Kashino}{Liu
  et~al\mbox{.}}{2016b}]%
        {liu2016scene}
\bibfield{author}{\bibinfo{person}{Xinhao Liu}, \bibinfo{person}{Takahito
  Kawanishi}, \bibinfo{person}{Xiaomeng Wu}, {and} \bibinfo{person}{Kunio
  Kashino}.} \bibinfo{year}{2016}\natexlab{b}.
\newblock \showarticletitle{Scene text recognition with {CNN} classifier and
  {WFST}-based word labeling}. In \bibinfo{booktitle}{\emph{Proceedings of
  ICPR}}. \bibinfo{pages}{3999--4004}.
\newblock


\bibitem[\protect\citeauthoryear{Liu, Liang, Yan, Chen, Qiao, and Yan}{Liu
  et~al\mbox{.}}{2018e}]%
        {liu2018fots}
\bibfield{author}{\bibinfo{person}{Xuebo Liu}, \bibinfo{person}{Ding Liang},
  \bibinfo{person}{Shi Yan}, \bibinfo{person}{Dagui Chen}, \bibinfo{person}{Yu
  Qiao}, {and} \bibinfo{person}{Junjie Yan}.} \bibinfo{year}{2018}\natexlab{e}.
\newblock \showarticletitle{Fots: Fast oriented text spotting with a unified
  network}. In \bibinfo{booktitle}{\emph{Proceedings of CVPR}}.
  \bibinfo{pages}{5676--5685}.
\newblock


\bibitem[\protect\citeauthoryear{Liu and Wang}{Liu and Wang}{2011}]%
        {liu2011robustly}
\bibfield{author}{\bibinfo{person}{Xiaoqian Liu} {and}
  \bibinfo{person}{Weiqiang Wang}.} \bibinfo{year}{2011}\natexlab{}.
\newblock \showarticletitle{Robustly extracting captions in videos based on
  stroke-like edges and spatio-temporal analysis}.
\newblock \bibinfo{journal}{\emph{IEEE Transactions on Multimedia}}
  \bibinfo{volume}{14}, \bibinfo{number}{2} (\bibinfo{year}{2011}),
  \bibinfo{pages}{482--489}.
\newblock


\bibitem[\protect\citeauthoryear{Liu, Zhang, Zhou, Jiang, Song, Li, Zhou, Wang,
  Wang, Liao, et~al\mbox{.}}{Liu et~al\mbox{.}}{2019d}]%
        {liu2019icdar}
\bibfield{author}{\bibinfo{person}{Xi Liu}, \bibinfo{person}{Rui Zhang},
  \bibinfo{person}{Yongsheng Zhou}, \bibinfo{person}{Qianyi Jiang},
  \bibinfo{person}{Qi Song}, \bibinfo{person}{Nan Li}, \bibinfo{person}{Kai
  Zhou}, \bibinfo{person}{Lei Wang}, \bibinfo{person}{Dong Wang},
  \bibinfo{person}{Minghui Liao}, {et~al\mbox{.}}}
  \bibinfo{year}{2019}\natexlab{d}.
\newblock \showarticletitle{{ICDAR} 2019 Robust Reading Challenge on Reading
  Chinese Text on Signboard}. In \bibinfo{booktitle}{\emph{Proceedings of
  ICDAR}}. \bibinfo{pages}{1577--1581}.
\newblock


\bibitem[\protect\citeauthoryear{Liu, Chen, Shen, He, Jin, and Wang}{Liu
  et~al\mbox{.}}{2020a}]%
        {Yuliang2020ABCNet}
\bibfield{author}{\bibinfo{person}{Yuliang Liu}, \bibinfo{person}{Hao Chen},
  \bibinfo{person}{Chunhua Shen}, \bibinfo{person}{Tong He},
  \bibinfo{person}{Lianwen Jin}, {and} \bibinfo{person}{Liangwei Wang}.}
  \bibinfo{year}{2020}\natexlab{a}.
\newblock \showarticletitle{{ABCNet}: Real-time Scene Text Spotting with
  Adaptive {Bezier}-Curve Network}. In \bibinfo{booktitle}{\emph{Proceedings of
  CVPR}}.
\newblock


\bibitem[\protect\citeauthoryear{Liu and Jin}{Liu and Jin}{2017}]%
        {liu2017deep}
\bibfield{author}{\bibinfo{person}{Yuliang Liu} {and} \bibinfo{person}{Lianwen
  Jin}.} \bibinfo{year}{2017}\natexlab{}.
\newblock \showarticletitle{Deep matching prior network: Toward tighter
  multi-oriented text detection}. In \bibinfo{booktitle}{\emph{Proceedings of
  CVPR}}. \bibinfo{pages}{1962--1969}.
\newblock


\bibitem[\protect\citeauthoryear{Liu, Jin, and Fang}{Liu
  et~al\mbox{.}}{2020b}]%
        {liu2019arbitrarily}
\bibfield{author}{\bibinfo{person}{Yuliang Liu}, \bibinfo{person}{Lianwen Jin},
  {and} \bibinfo{person}{Chuanming Fang}.} \bibinfo{year}{2020}\natexlab{b}.
\newblock \showarticletitle{Arbitrarily Shaped Scene Text Detection with a Mask
  Tightness Text Detector}.
\newblock \bibinfo{journal}{\emph{IEEE Transactions on Image Processing}}
  \bibinfo{volume}{29} (\bibinfo{year}{2020}), \bibinfo{pages}{2918--2930}.
\newblock


\bibitem[\protect\citeauthoryear{Liu, Jin, Xie, Luo, Zhang, and Xie}{Liu
  et~al\mbox{.}}{2019b}]%
        {liu2019tightness}
\bibfield{author}{\bibinfo{person}{Yuliang Liu}, \bibinfo{person}{Lianwen Jin},
  \bibinfo{person}{Zecheng Xie}, \bibinfo{person}{Canjie Luo},
  \bibinfo{person}{Shuaitao Zhang}, {and} \bibinfo{person}{Lele Xie}.}
  \bibinfo{year}{2019}\natexlab{b}.
\newblock \showarticletitle{Tightness-aware evaluation protocol for scene text
  detection}. In \bibinfo{booktitle}{\emph{Proceedings of CVPR}}.
  \bibinfo{pages}{9612--9620}.
\newblock


\bibitem[\protect\citeauthoryear{Liu, Jin, Zhang, Luo, and Zhang}{Liu
  et~al\mbox{.}}{2019c}]%
        {liu2019curved}
\bibfield{author}{\bibinfo{person}{Yuliang Liu}, \bibinfo{person}{Lianwen Jin},
  \bibinfo{person}{Shuaitao Zhang}, \bibinfo{person}{Canjie Luo}, {and}
  \bibinfo{person}{Sheng Zhang}.} \bibinfo{year}{2019}\natexlab{c}.
\newblock \showarticletitle{Curved scene text detection via transverse and
  longitudinal sequence connection}.
\newblock \bibinfo{journal}{\emph{Pattern Recognition}}  \bibinfo{volume}{90}
  (\bibinfo{year}{2019}), \bibinfo{pages}{337--345}.
\newblock


\bibitem[\protect\citeauthoryear{Liu, Wang, Jin, and Wassell}{Liu
  et~al\mbox{.}}{2018f}]%
        {liu2018synthetically}
\bibfield{author}{\bibinfo{person}{Yang Liu}, \bibinfo{person}{Zhaowen Wang},
  \bibinfo{person}{Hailin Jin}, {and} \bibinfo{person}{Ian Wassell}.}
  \bibinfo{year}{2018}\natexlab{f}.
\newblock \showarticletitle{Synthetically supervised feature learning for scene
  text recognition}. In \bibinfo{booktitle}{\emph{Proceedings of ECCV}}.
  \bibinfo{pages}{449--465}.
\newblock


\bibitem[\protect\citeauthoryear{Liu, Li, Ren, Goh, and Yu}{Liu
  et~al\mbox{.}}{2018d}]%
        {liu2018squeezedtext}
\bibfield{author}{\bibinfo{person}{Zichuan Liu}, \bibinfo{person}{Yixing Li},
  \bibinfo{person}{Fengbo Ren}, \bibinfo{person}{Wang~Ling Goh}, {and}
  \bibinfo{person}{Hao Yu}.} \bibinfo{year}{2018}\natexlab{d}.
\newblock \showarticletitle{Squeezedtext: A real-time scene text recognition by
  binary convolutional encoder-decoder network}. In
  \bibinfo{booktitle}{\emph{Proceedings of AAAI}}. \bibinfo{pages}{7194--7201}.
\newblock


\bibitem[\protect\citeauthoryear{Long, He, and Ya}{Long et~al\mbox{.}}{2018}]%
        {long2018scene}
\bibfield{author}{\bibinfo{person}{Shangbang Long}, \bibinfo{person}{Xin He},
  {and} \bibinfo{person}{Cong Ya}.} \bibinfo{year}{2018}\natexlab{}.
\newblock \showarticletitle{Scene text detection and recognition: The deep
  learning era}.
\newblock \bibinfo{journal}{\emph{CoRR abs/1811.04256}} (\bibinfo{year}{2018}).
\newblock


\bibitem[\protect\citeauthoryear{Long and Yao}{Long and Yao}{2020}]%
        {Long2020UnrealText}
\bibfield{author}{\bibinfo{person}{Shangbang Long} {and} \bibinfo{person}{Cong
  Yao}.} \bibinfo{year}{2020}\natexlab{}.
\newblock \showarticletitle{{UnrealText}: Synthesizing Realistic Scene Text
  Images from the Unreal World}. In \bibinfo{booktitle}{\emph{Proceedings of
  CVPR}}.
\newblock


\bibitem[\protect\citeauthoryear{Lu, McCaffrey, and Kuo}{Lu
  et~al\mbox{.}}{2011}]%
        {lu2011foreign}
\bibfield{author}{\bibinfo{person}{Fang Lu}, \bibinfo{person}{Corey~S
  McCaffrey}, {and} \bibinfo{person}{Elaine~I Kuo}.}
  \bibinfo{year}{2011}\natexlab{}.
\newblock \bibinfo{title}{Foreign language abbreviation translation in an
  instant messaging system}.
\newblock
\newblock
\newblock
\shownote{US Patent 7,890,525.}


\bibitem[\protect\citeauthoryear{Lucas}{Lucas}{2005}]%
        {lucas2005icdartext}
\bibfield{author}{\bibinfo{person}{Simon~M Lucas}.}
  \bibinfo{year}{2005}\natexlab{}.
\newblock \showarticletitle{{ICDAR} 2005 text locating competition results}. In
  \bibinfo{booktitle}{\emph{Proceedings of ICDAR}}. \bibinfo{pages}{80--84}.
\newblock


\bibitem[\protect\citeauthoryear{Lucas, Panaretos, Sosa, Tang, Wong, and
  Young}{Lucas et~al\mbox{.}}{2003}]%
        {lucas2005icdar}
\bibfield{author}{\bibinfo{person}{Simon~M Lucas}, \bibinfo{person}{Alex
  Panaretos}, \bibinfo{person}{Luis Sosa}, \bibinfo{person}{Anthony Tang},
  \bibinfo{person}{Shirley Wong}, {and} \bibinfo{person}{Robert Young}.}
  \bibinfo{year}{2003}\natexlab{}.
\newblock \showarticletitle{{ICDAR} 2003 robust reading competitions}. In
  \bibinfo{booktitle}{\emph{Proceedings of ICDAR}}. \bibinfo{pages}{682--687}.
\newblock


\bibitem[\protect\citeauthoryear{Luo, Jin, and Sun}{Luo et~al\mbox{.}}{2019}]%
        {cluo2019moran}
\bibfield{author}{\bibinfo{person}{Canjie Luo}, \bibinfo{person}{Lianwen Jin},
  {and} \bibinfo{person}{Zenghui Sun}.} \bibinfo{year}{2019}\natexlab{}.
\newblock \showarticletitle{{MORAN}: A Multi-Object Rectified Attention Network
  for Scene Text Recognition}.
\newblock \bibinfo{journal}{\emph{Pattern Recognition}}  \bibinfo{volume}{90}
  (\bibinfo{year}{2019}), \bibinfo{pages}{109--118}.
\newblock


\bibitem[\protect\citeauthoryear{Luo, Lin, Liu, Lianwen, and Chunhua}{Luo
  et~al\mbox{.}}{2020}]%
        {Luo2020Separating}
\bibfield{author}{\bibinfo{person}{Canjie Luo}, \bibinfo{person}{Qingxiang
  Lin}, \bibinfo{person}{Yuliang Liu}, \bibinfo{person}{Jin Lianwen}, {and}
  \bibinfo{person}{Shen Chunhua}.} \bibinfo{year}{2020}\natexlab{}.
\newblock \showarticletitle{Separating Content from Style Using Adversarial
  Learning for Recognizing Text in the Wild}.
\newblock \bibinfo{journal}{\emph{CoRR abs/2001.04189}} (\bibinfo{year}{2020}).
\newblock


\bibitem[\protect\citeauthoryear{Lyu, Liao, Yao, Wu, and Bai}{Lyu
  et~al\mbox{.}}{2018}]%
        {lyu2018mask}
\bibfield{author}{\bibinfo{person}{Pengyuan Lyu}, \bibinfo{person}{Minghui
  Liao}, \bibinfo{person}{Cong Yao}, \bibinfo{person}{Wenhao Wu}, {and}
  \bibinfo{person}{Xiang Bai}.} \bibinfo{year}{2018}\natexlab{}.
\newblock \showarticletitle{Mask textspotter: An end-to-end trainable neural
  network for spotting text with arbitrary shapes}. In
  \bibinfo{booktitle}{\emph{Proceedings of ECCV}}. \bibinfo{pages}{67--83}.
\newblock


\bibitem[\protect\citeauthoryear{Mei, Dai, Shi, and Bai}{Mei
  et~al\mbox{.}}{2016}]%
        {mei2016scene}
\bibfield{author}{\bibinfo{person}{Jieru Mei}, \bibinfo{person}{Luo Dai},
  \bibinfo{person}{Baoguang Shi}, {and} \bibinfo{person}{Xiang Bai}.}
  \bibinfo{year}{2016}\natexlab{}.
\newblock \showarticletitle{Scene text script identification with convolutional
  recurrent neural networks}. In \bibinfo{booktitle}{\emph{Proceedings of
  ICPR}}. \bibinfo{pages}{4053--4058}.
\newblock


\bibitem[\protect\citeauthoryear{Miao, Gowayyed, and Metze}{Miao
  et~al\mbox{.}}{2015}]%
        {miao2015eesen}
\bibfield{author}{\bibinfo{person}{Yajie Miao}, \bibinfo{person}{Mohammad
  Gowayyed}, {and} \bibinfo{person}{Florian Metze}.}
  \bibinfo{year}{2015}\natexlab{}.
\newblock \showarticletitle{{EESEN}: End-to-end speech recognition using deep
  RNN models and {WFST}-based decoding}. In \bibinfo{booktitle}{\emph{IEEE
  Workshop on Automatic Speech Recognition and Understanding (ASRU)}}.
  \bibinfo{pages}{167--174}.
\newblock


\bibitem[\protect\citeauthoryear{Mishra, Alahari, and Jawahar}{Mishra
  et~al\mbox{.}}{2012a}]%
        {mishra2012scene}
\bibfield{author}{\bibinfo{person}{Anand Mishra}, \bibinfo{person}{Karteek
  Alahari}, {and} \bibinfo{person}{CV Jawahar}.}
  \bibinfo{year}{2012}\natexlab{a}.
\newblock \showarticletitle{Scene text recognition using higher order language
  priors}. In \bibinfo{booktitle}{\emph{Proceedings of BMVC}}.
  \bibinfo{pages}{1--11}.
\newblock


\bibitem[\protect\citeauthoryear{Mishra, Alahari, and Jawahar}{Mishra
  et~al\mbox{.}}{2012b}]%
        {mishra2012top}
\bibfield{author}{\bibinfo{person}{Anand Mishra}, \bibinfo{person}{Karteek
  Alahari}, {and} \bibinfo{person}{CV Jawahar}.}
  \bibinfo{year}{2012}\natexlab{b}.
\newblock \showarticletitle{Top-down and bottom-up cues for scene text
  recognition}. In \bibinfo{booktitle}{\emph{Proceedings of CVPR}}.
  \bibinfo{pages}{2687--2694}.
\newblock


\bibitem[\protect\citeauthoryear{Mishra, Alahari, and Jawahar}{Mishra
  et~al\mbox{.}}{2016}]%
        {mishra2016enhancing}
\bibfield{author}{\bibinfo{person}{Anand Mishra}, \bibinfo{person}{Karteek
  Alahari}, {and} \bibinfo{person}{CV Jawahar}.}
  \bibinfo{year}{2016}\natexlab{}.
\newblock \showarticletitle{Enhancing energy minimization framework for scene
  text recognition with top-down cues}.
\newblock \bibinfo{journal}{\emph{Computer Vision and Image Understanding}}
  \bibinfo{volume}{145} (\bibinfo{year}{2016}), \bibinfo{pages}{30--42}.
\newblock


\bibitem[\protect\citeauthoryear{Mohri, Pereira, and Riley}{Mohri
  et~al\mbox{.}}{2002}]%
        {mohri2002weighted}
\bibfield{author}{\bibinfo{person}{Mehryar Mohri}, \bibinfo{person}{Fernando
  Pereira}, {and} \bibinfo{person}{Michael Riley}.}
  \bibinfo{year}{2002}\natexlab{}.
\newblock \showarticletitle{Weighted finite-state transducers in speech
  recognition}.
\newblock \bibinfo{journal}{\emph{Computer Speech and Language}}
  \bibinfo{volume}{16}, \bibinfo{number}{1} (\bibinfo{year}{2002}),
  \bibinfo{pages}{69--88}.
\newblock


\bibitem[\protect\citeauthoryear{Mosleh, Bouguila, and Hamza}{Mosleh
  et~al\mbox{.}}{2012}]%
        {mosleh2012image}
\bibfield{author}{\bibinfo{person}{Ali Mosleh}, \bibinfo{person}{Nizar
  Bouguila}, {and} \bibinfo{person}{A~Ben Hamza}.}
  \bibinfo{year}{2012}\natexlab{}.
\newblock \showarticletitle{Image Text Detection Using a Bandlet-Based Edge
  Detector and Stroke Width Transform.}. In
  \bibinfo{booktitle}{\emph{Proceedings of BMVC}}. \bibinfo{pages}{1--12}.
\newblock


\bibitem[\protect\citeauthoryear{Nagy}{Nagy}{2000}]%
        {nagy2000twenty}
\bibfield{author}{\bibinfo{person}{George Nagy}.}
  \bibinfo{year}{2000}\natexlab{}.
\newblock \showarticletitle{Twenty years of document image analysis in PAMI}.
\newblock \bibinfo{journal}{\emph{IEEE Trans. Pattern Anal. Mach. Intell}}
  \bibinfo{volume}{22}, \bibinfo{number}{1} (\bibinfo{year}{2000}),
  \bibinfo{pages}{38--62}.
\newblock


\bibitem[\protect\citeauthoryear{Nayef, Patel, Busta, Chowdhury, Karatzas,
  Khlif, Matas, Pal, Burie, Liu, et~al\mbox{.}}{Nayef et~al\mbox{.}}{2019}]%
        {nayef2019icdar2019}
\bibfield{author}{\bibinfo{person}{Nibal Nayef}, \bibinfo{person}{Yash Patel},
  \bibinfo{person}{Michal Busta}, \bibinfo{person}{Pinaki~Nath Chowdhury},
  \bibinfo{person}{Dimosthenis Karatzas}, \bibinfo{person}{Wafa Khlif},
  \bibinfo{person}{Jiri Matas}, \bibinfo{person}{Umapada Pal},
  \bibinfo{person}{Jean-Christophe Burie}, \bibinfo{person}{Cheng-lin Liu},
  {et~al\mbox{.}}} \bibinfo{year}{2019}\natexlab{}.
\newblock \showarticletitle{{ICDAR}2019 Robust Reading Challenge on
  Multi-lingual Scene Text Detection and Recognition--{RRC-MLT}-2019}. In
  \bibinfo{booktitle}{\emph{Proceedings of ICDAR}}.
  \bibinfo{pages}{1582--1587}.
\newblock


\bibitem[\protect\citeauthoryear{Netzer, Wang, Coates, Bissacco, Wu, and
  Ng}{Netzer et~al\mbox{.}}{2011}]%
        {netzer2011reading}
\bibfield{author}{\bibinfo{person}{Yuval Netzer}, \bibinfo{person}{Tao Wang},
  \bibinfo{person}{Adam Coates}, \bibinfo{person}{Alessandro Bissacco},
  \bibinfo{person}{Bo Wu}, {and} \bibinfo{person}{Andrew~Y Ng}.}
  \bibinfo{year}{2011}\natexlab{}.
\newblock \showarticletitle{Reading digits in natural images with unsupervised
  feature learning}. In \bibinfo{booktitle}{\emph{Proceedings of NIPS}}.
\newblock


\bibitem[\protect\citeauthoryear{Neumann and Matas}{Neumann and Matas}{2010}]%
        {neumann2010method}
\bibfield{author}{\bibinfo{person}{Lukas Neumann} {and} \bibinfo{person}{Jiri
  Matas}.} \bibinfo{year}{2010}\natexlab{}.
\newblock \showarticletitle{A method for text localization and recognition in
  real-world images}. In \bibinfo{booktitle}{\emph{Proceedings of ACCV}}.
  \bibinfo{pages}{770--783}.
\newblock


\bibitem[\protect\citeauthoryear{Neumann and Matas}{Neumann and Matas}{2012}]%
        {neumann2012real}
\bibfield{author}{\bibinfo{person}{Luk{\'a}{\v{s}} Neumann} {and}
  \bibinfo{person}{Ji{\v{r}}{\'\i} Matas}.} \bibinfo{year}{2012}\natexlab{}.
\newblock \showarticletitle{Real-time scene text localization and recognition}.
  In \bibinfo{booktitle}{\emph{Proceedings of CVPR}}.
  \bibinfo{pages}{3538--3545}.
\newblock


\bibitem[\protect\citeauthoryear{Neumann and Matas}{Neumann and Matas}{2015a}]%
        {neumann2015efficient}
\bibfield{author}{\bibinfo{person}{Luk{\'a}{\v{s}} Neumann} {and}
  \bibinfo{person}{Ji{\v{r}}{\'\i} Matas}.} \bibinfo{year}{2015}\natexlab{a}.
\newblock \showarticletitle{Efficient scene text localization and recognition
  with local character refinement}. In \bibinfo{booktitle}{\emph{Proceedings of
  ICDAR}}. \bibinfo{pages}{746--750}.
\newblock


\bibitem[\protect\citeauthoryear{Neumann and Matas}{Neumann and Matas}{2015b}]%
        {neumann2015real}
\bibfield{author}{\bibinfo{person}{Luk{\'a}{\v{s}} Neumann} {and}
  \bibinfo{person}{Ji{\v{r}}{\'\i} Matas}.} \bibinfo{year}{2015}\natexlab{b}.
\newblock \showarticletitle{Real-time lexicon-free scene text localization and
  recognition}.
\newblock \bibinfo{journal}{\emph{IEEE Trans. Pattern Anal. Mach. Intell}}
  \bibinfo{volume}{38}, \bibinfo{number}{9} (\bibinfo{year}{2015}),
  \bibinfo{pages}{1872--1885}.
\newblock


\bibitem[\protect\citeauthoryear{Nomura, Yamanaka, Katai, Kawakami, and
  Shiose}{Nomura et~al\mbox{.}}{2005}]%
        {nomura2005novel}
\bibfield{author}{\bibinfo{person}{Shigueo Nomura}, \bibinfo{person}{Keiji
  Yamanaka}, \bibinfo{person}{Osamu Katai}, \bibinfo{person}{Hiroshi Kawakami},
  {and} \bibinfo{person}{Takayuki Shiose}.} \bibinfo{year}{2005}\natexlab{}.
\newblock \showarticletitle{A novel adaptive morphological approach for
  degraded character image segmentation}.
\newblock \bibinfo{journal}{\emph{Pattern Recognition}} \bibinfo{volume}{38},
  \bibinfo{number}{11} (\bibinfo{year}{2005}), \bibinfo{pages}{1961--1975}.
\newblock


\bibitem[\protect\citeauthoryear{Pan, Hou, and Liu}{Pan et~al\mbox{.}}{2011}]%
        {Yi2011AHybrid}
\bibfield{author}{\bibinfo{person}{Yi{-}Feng Pan}, \bibinfo{person}{Xinwen
  Hou}, {and} \bibinfo{person}{Cheng{-}Lin Liu}.}
  \bibinfo{year}{2011}\natexlab{}.
\newblock \showarticletitle{A Hybrid Approach to Detect and Localize Texts in
  Natural Scene Images}.
\newblock \bibinfo{journal}{\emph{IEEE Transactions on Image Processing}}
  \bibinfo{volume}{20}, \bibinfo{number}{3} (\bibinfo{year}{2011}),
  \bibinfo{pages}{800--813}.
\newblock


\bibitem[\protect\citeauthoryear{Peyrard, Baccouche, Mamalet, and
  Garcia}{Peyrard et~al\mbox{.}}{2015}]%
        {peyrard2015icdar2015}
\bibfield{author}{\bibinfo{person}{Cl{\'e}ment Peyrard}, \bibinfo{person}{Moez
  Baccouche}, \bibinfo{person}{Franck Mamalet}, {and}
  \bibinfo{person}{Christophe Garcia}.} \bibinfo{year}{2015}\natexlab{}.
\newblock \showarticletitle{{ICDAR}2015 competition on text image
  super-resolution}. In \bibinfo{booktitle}{\emph{Proceedings of ICDAR}}.
  \bibinfo{pages}{1201--1205}.
\newblock


\bibitem[\protect\citeauthoryear{Qi, Chen, Xiao, Li, Zou, and Cui}{Qi
  et~al\mbox{.}}{2019}]%
        {qi2019novel}
\bibfield{author}{\bibinfo{person}{Xianbiao Qi}, \bibinfo{person}{Yihao Chen},
  \bibinfo{person}{Rong Xiao}, \bibinfo{person}{Chun-Guang Li},
  \bibinfo{person}{Qin Zou}, {and} \bibinfo{person}{Shuguang Cui}.}
  \bibinfo{year}{2019}\natexlab{}.
\newblock \showarticletitle{A Novel Joint Character Categorization and
  Localization Approach for Character-Level Scene Text Recognition}. In
  \bibinfo{booktitle}{\emph{Proceedings of ICDAR: Workshops}}.
  \bibinfo{pages}{83--90}.
\newblock


\bibitem[\protect\citeauthoryear{Qiao, Tang, Cheng, Xu, Niu, Pu, and Wu}{Qiao
  et~al\mbox{.}}{2020a}]%
        {qiao2019text}
\bibfield{author}{\bibinfo{person}{Liang Qiao}, \bibinfo{person}{Sanli Tang},
  \bibinfo{person}{Zhanzhan Cheng}, \bibinfo{person}{Yunlu Xu},
  \bibinfo{person}{Yi Niu}, \bibinfo{person}{Shiliang Pu}, {and}
  \bibinfo{person}{Fei Wu}.} \bibinfo{year}{2020}\natexlab{a}.
\newblock \showarticletitle{Text Perceptron: Towards End-to-End
  Arbitrary-Shaped Text Spotting}. In \bibinfo{booktitle}{\emph{Proceedings of
  AAAI}}.
\newblock


\bibitem[\protect\citeauthoryear{Qiao, Zhou, Yang, Zhou, and Wang}{Qiao
  et~al\mbox{.}}{2020b}]%
        {qiao2020seed}
\bibfield{author}{\bibinfo{person}{Zhi Qiao}, \bibinfo{person}{Yu Zhou},
  \bibinfo{person}{Dongbao Yang}, \bibinfo{person}{Yucan Zhou}, {and}
  \bibinfo{person}{Weiping Wang}.} \bibinfo{year}{2020}\natexlab{b}.
\newblock \showarticletitle{SEED: Semantics Enhanced Encoder-Decoder Framework
  for Scene Text Recognition}. In \bibinfo{booktitle}{\emph{Proceedings of
  CVPR}}.
\newblock


\bibitem[\protect\citeauthoryear{Qin, Bissacco, Raptis, Fujii, and Xiao}{Qin
  et~al\mbox{.}}{2019}]%
        {qin2019towards}
\bibfield{author}{\bibinfo{person}{Siyang Qin}, \bibinfo{person}{Alessandro
  Bissacco}, \bibinfo{person}{Michalis Raptis}, \bibinfo{person}{Yasuhisa
  Fujii}, {and} \bibinfo{person}{Ying Xiao}.} \bibinfo{year}{2019}\natexlab{}.
\newblock \showarticletitle{Towards Unconstrained End-to-End Text Spotting}. In
  \bibinfo{booktitle}{\emph{Proceedings of ICCV}}. \bibinfo{pages}{4704--4714}.
\newblock


\bibitem[\protect\citeauthoryear{Qiu and Yuille}{Qiu and Yuille}{2016}]%
        {Qiu2016UnrealCV}
\bibfield{author}{\bibinfo{person}{Weichao Qiu} {and} \bibinfo{person}{Alan~L.
  Yuille}.} \bibinfo{year}{2016}\natexlab{}.
\newblock \showarticletitle{UnrealCV: Connecting Computer Vision to Unreal
  Engine}. In \bibinfo{booktitle}{\emph{Proceedings of ECCV}}.
  \bibinfo{pages}{909--916}.
\newblock


\bibitem[\protect\citeauthoryear{Quy~Phan, Shivakumara, Tian, and
  Lim~Tan}{Quy~Phan et~al\mbox{.}}{2013}]%
        {quy2013recognizing}
\bibfield{author}{\bibinfo{person}{Trung Quy~Phan},
  \bibinfo{person}{Palaiahnakote Shivakumara}, \bibinfo{person}{Shangxuan
  Tian}, {and} \bibinfo{person}{Chew Lim~Tan}.}
  \bibinfo{year}{2013}\natexlab{}.
\newblock \showarticletitle{Recognizing text with perspective distortion in
  natural scenes}. In \bibinfo{booktitle}{\emph{Proceedings of ICCV}}.
  \bibinfo{pages}{569--576}.
\newblock


\bibitem[\protect\citeauthoryear{Risnumawan, Shivakumara, Chan, and
  Tan}{Risnumawan et~al\mbox{.}}{2014}]%
        {risnumawan2014robust}
\bibfield{author}{\bibinfo{person}{Anhar Risnumawan},
  \bibinfo{person}{Palaiahankote Shivakumara}, \bibinfo{person}{Chee~Seng
  Chan}, {and} \bibinfo{person}{Chew~Lim Tan}.}
  \bibinfo{year}{2014}\natexlab{}.
\newblock \showarticletitle{A robust arbitrary text detection system for
  natural scene images}.
\newblock \bibinfo{journal}{\emph{Expert Systems with Applications}}
  \bibinfo{volume}{41}, \bibinfo{number}{18} (\bibinfo{year}{2014}),
  \bibinfo{pages}{8027--8048}.
\newblock


\bibitem[\protect\citeauthoryear{Rodriguez-Serrano, Gordo, and
  Perronnin}{Rodriguez-Serrano et~al\mbox{.}}{2015}]%
        {rodriguez2015label}
\bibfield{author}{\bibinfo{person}{Jose~A Rodriguez-Serrano},
  \bibinfo{person}{Albert Gordo}, {and} \bibinfo{person}{Florent Perronnin}.}
  \bibinfo{year}{2015}\natexlab{}.
\newblock \showarticletitle{Label embedding: A frugal baseline for text
  recognition}.
\newblock \bibinfo{journal}{\emph{Int. J. Comput. Vis}} \bibinfo{volume}{113},
  \bibinfo{number}{3} (\bibinfo{year}{2015}), \bibinfo{pages}{193--207}.
\newblock


\bibitem[\protect\citeauthoryear{Rouh and Beaudet}{Rouh and Beaudet}{2019}]%
        {rouh2019method}
\bibfield{author}{\bibinfo{person}{Alain Rouh} {and} \bibinfo{person}{Jean
  Beaudet}.} \bibinfo{year}{2019}\natexlab{}.
\newblock \bibinfo{title}{Method and a device for tracking characters that
  appear on a plurality of images of a video stream of a text}.
\newblock
\newblock
\newblock
\shownote{US Patent App. 10/185,873.}


\bibitem[\protect\citeauthoryear{Sanchez, Romero, Toselli, Villegas, and
  Vidal}{Sanchez et~al\mbox{.}}{2017}]%
        {sanchez2017icdar2017}
\bibfield{author}{\bibinfo{person}{Joan~Andreu Sanchez},
  \bibinfo{person}{Ver{\'o}nica Romero}, \bibinfo{person}{Alejandro~H Toselli},
  \bibinfo{person}{Mauricio Villegas}, {and} \bibinfo{person}{Enrique Vidal}.}
  \bibinfo{year}{2017}\natexlab{}.
\newblock \showarticletitle{{ICDAR}2017 competition on handwritten text
  recognition on the read dataset}. In \bibinfo{booktitle}{\emph{Proceedings of
  ICDAR}}. \bibinfo{pages}{1383--1388}.
\newblock


\bibitem[\protect\citeauthoryear{Sermanet, Chintala, and LeCun}{Sermanet
  et~al\mbox{.}}{2012}]%
        {Sermanet2012convolutional}
\bibfield{author}{\bibinfo{person}{Pierre Sermanet}, \bibinfo{person}{Soumith
  Chintala}, {and} \bibinfo{person}{Yann LeCun}.}
  \bibinfo{year}{2012}\natexlab{}.
\newblock \showarticletitle{Convolutional neural networks applied to house
  numbers digit classification}. In \bibinfo{booktitle}{\emph{Proceedings of
  ICPR}}. \bibinfo{pages}{3288--3291}.
\newblock


\bibitem[\protect\citeauthoryear{Shahab, Shafait, and Dengel}{Shahab
  et~al\mbox{.}}{2011}]%
        {shahab2011icdar}
\bibfield{author}{\bibinfo{person}{Asif Shahab}, \bibinfo{person}{Faisal
  Shafait}, {and} \bibinfo{person}{Andreas Dengel}.}
  \bibinfo{year}{2011}\natexlab{}.
\newblock \showarticletitle{{ICDAR} 2011 robust reading competition challenge
  2: Reading text in scene images}. In \bibinfo{booktitle}{\emph{Proceedings of
  ICDAR}}. \bibinfo{pages}{1491--1496}.
\newblock


\bibitem[\protect\citeauthoryear{Sheng, Chen, and Xu}{Sheng
  et~al\mbox{.}}{2019}]%
        {sheng2018nrtr}
\bibfield{author}{\bibinfo{person}{Fenfen Sheng}, \bibinfo{person}{Zhineng
  Chen}, {and} \bibinfo{person}{Bo Xu}.} \bibinfo{year}{2019}\natexlab{}.
\newblock \showarticletitle{{NRTR}: A No-Recurrence Sequence-to-Sequence Model
  For Scene Text Recognition}. In \bibinfo{booktitle}{\emph{Proceedings of
  ICDAR}}. \bibinfo{pages}{781--786}.
\newblock


\bibitem[\protect\citeauthoryear{Shi, Bai, and Yao}{Shi et~al\mbox{.}}{2016a}]%
        {shi2016script}
\bibfield{author}{\bibinfo{person}{Baoguang Shi}, \bibinfo{person}{Xiang Bai},
  {and} \bibinfo{person}{Cong Yao}.} \bibinfo{year}{2016}\natexlab{a}.
\newblock \showarticletitle{Script identification in the wild via
  discriminative convolutional neural network}.
\newblock \bibinfo{journal}{\emph{Pattern Recognition}}  \bibinfo{volume}{52}
  (\bibinfo{year}{2016}), \bibinfo{pages}{448--458}.
\newblock


\bibitem[\protect\citeauthoryear{Shi, Bai, and Yao}{Shi et~al\mbox{.}}{2017a}]%
        {shi2017end}
\bibfield{author}{\bibinfo{person}{Baoguang Shi}, \bibinfo{person}{Xiang Bai},
  {and} \bibinfo{person}{Cong Yao}.} \bibinfo{year}{2017}\natexlab{a}.
\newblock \showarticletitle{An end-to-end trainable neural network for
  image-based sequence recognition and its application to scene text
  recognition}.
\newblock \bibinfo{journal}{\emph{IEEE Trans. Pattern Anal. Mach. Intell}}
  \bibinfo{volume}{39}, \bibinfo{number}{11} (\bibinfo{year}{2017}),
  \bibinfo{pages}{2298--2304}.
\newblock


\bibitem[\protect\citeauthoryear{Shi, Wang, Lyu, Yao, and Bai}{Shi
  et~al\mbox{.}}{2016b}]%
        {shi2016robust}
\bibfield{author}{\bibinfo{person}{Baoguang Shi}, \bibinfo{person}{Xinggang
  Wang}, \bibinfo{person}{Pengyuan Lyu}, \bibinfo{person}{Cong Yao}, {and}
  \bibinfo{person}{Xiang Bai}.} \bibinfo{year}{2016}\natexlab{b}.
\newblock \showarticletitle{Robust scene text recognition with automatic
  rectification}. In \bibinfo{booktitle}{\emph{Proceedings of CVPR}}.
  \bibinfo{pages}{4168--4176}.
\newblock


\bibitem[\protect\citeauthoryear{Shi, Yang, Wang, Lyu, Yao, and Bai}{Shi
  et~al\mbox{.}}{2019}]%
        {shi2018aster}
\bibfield{author}{\bibinfo{person}{Baoguang Shi}, \bibinfo{person}{Mingkun
  Yang}, \bibinfo{person}{Xinggang Wang}, \bibinfo{person}{Pengyuan Lyu},
  \bibinfo{person}{Cong Yao}, {and} \bibinfo{person}{Xiang Bai}.}
  \bibinfo{year}{2019}\natexlab{}.
\newblock \showarticletitle{{ASTER}: An Attentional Scene Text Recognizer with
  Flexible Rectification}.
\newblock \bibinfo{journal}{\emph{IEEE Trans. Pattern Anal. Mach. Intell}}
  \bibinfo{volume}{41}, \bibinfo{number}{9} (\bibinfo{year}{2019}),
  \bibinfo{pages}{2035--2048}.
\newblock


\bibitem[\protect\citeauthoryear{Shi, Yao, Liao, Yang, Xu, Cui, Belongie, Lu,
  and Bai}{Shi et~al\mbox{.}}{2017b}]%
        {shi2017icdar2017}
\bibfield{author}{\bibinfo{person}{Baoguang Shi}, \bibinfo{person}{Cong Yao},
  \bibinfo{person}{Minghui Liao}, \bibinfo{person}{Mingkun Yang},
  \bibinfo{person}{Pei Xu}, \bibinfo{person}{Linyan Cui},
  \bibinfo{person}{Serge Belongie}, \bibinfo{person}{Shijian Lu}, {and}
  \bibinfo{person}{Xiang Bai}.} \bibinfo{year}{2017}\natexlab{b}.
\newblock \showarticletitle{{ICDAR}2017 competition on reading chinese text in
  the wild (rctw-17)}. In \bibinfo{booktitle}{\emph{Proceedings of ICDAR}}.
  \bibinfo{pages}{1429--1434}.
\newblock


\bibitem[\protect\citeauthoryear{Shi, Yao, Zhang, Guo, Huang, and Bai}{Shi
  et~al\mbox{.}}{2015}]%
        {shi2015automatic}
\bibfield{author}{\bibinfo{person}{Baoguang Shi}, \bibinfo{person}{Cong Yao},
  \bibinfo{person}{Chengquan Zhang}, \bibinfo{person}{Xiaowei Guo},
  \bibinfo{person}{Feiyue Huang}, {and} \bibinfo{person}{Xiang Bai}.}
  \bibinfo{year}{2015}\natexlab{}.
\newblock \showarticletitle{Automatic script identification in the wild}. In
  \bibinfo{booktitle}{\emph{Proceedings of ICDAR}}. \bibinfo{pages}{531--535}.
\newblock


\bibitem[\protect\citeauthoryear{Shi, Wang, Xiao, Zhang, Gao, and Zhang}{Shi
  et~al\mbox{.}}{2013}]%
        {shi2013scene}
\bibfield{author}{\bibinfo{person}{Cunzhao Shi}, \bibinfo{person}{Chunheng
  Wang}, \bibinfo{person}{Baihua Xiao}, \bibinfo{person}{Yang Zhang},
  \bibinfo{person}{Song Gao}, {and} \bibinfo{person}{Zhong Zhang}.}
  \bibinfo{year}{2013}\natexlab{}.
\newblock \showarticletitle{Scene text recognition using part-based
  tree-structured character detection}. In
  \bibinfo{booktitle}{\emph{Proceedings of CVPR}}. \bibinfo{pages}{2961--2968}.
\newblock


\bibitem[\protect\citeauthoryear{Shivakumara, Bhowmick, Su, Tan, and
  Pal}{Shivakumara et~al\mbox{.}}{2011}]%
        {shivakumara2011new}
\bibfield{author}{\bibinfo{person}{Palaiahnakote Shivakumara},
  \bibinfo{person}{Souvik Bhowmick}, \bibinfo{person}{Bolan Su},
  \bibinfo{person}{Chew~Lim Tan}, {and} \bibinfo{person}{Umapada Pal}.}
  \bibinfo{year}{2011}\natexlab{}.
\newblock \showarticletitle{A new gradient based character segmentation method
  for video text recognition}. In \bibinfo{booktitle}{\emph{Proceedings of
  ICDAR}}. \bibinfo{pages}{126--130}.
\newblock


\bibitem[\protect\citeauthoryear{Shivakumara, Huang, Phan, and Tan}{Shivakumara
  et~al\mbox{.}}{2010}]%
        {shivakumara2010accurate}
\bibfield{author}{\bibinfo{person}{Palaiahnakote Shivakumara},
  \bibinfo{person}{Weihua Huang}, \bibinfo{person}{Trung~Quy Phan}, {and}
  \bibinfo{person}{Chew~Lim Tan}.} \bibinfo{year}{2010}\natexlab{}.
\newblock \showarticletitle{Accurate video text detection through
  classification of low and high contrast images}.
\newblock \bibinfo{journal}{\emph{Pattern Recognition}} \bibinfo{volume}{43},
  \bibinfo{number}{6} (\bibinfo{year}{2010}), \bibinfo{pages}{2165--2185}.
\newblock


\bibitem[\protect\citeauthoryear{Shivakumara, Phan, and Tan}{Shivakumara
  et~al\mbox{.}}{2009}]%
        {shivakumara2009gradient}
\bibfield{author}{\bibinfo{person}{Palaiahnakote Shivakumara},
  \bibinfo{person}{Trung~Quy Phan}, {and} \bibinfo{person}{Chew~Lim Tan}.}
  \bibinfo{year}{2009}\natexlab{}.
\newblock \showarticletitle{A gradient difference based technique for video
  text detection}. In \bibinfo{booktitle}{\emph{Proceedings of ICDAR}}.
  \bibinfo{pages}{156--160}.
\newblock


\bibitem[\protect\citeauthoryear{Simonyan and Zisserman}{Simonyan and
  Zisserman}{2015}]%
        {simonyan2014very}
\bibfield{author}{\bibinfo{person}{Karen Simonyan} {and}
  \bibinfo{person}{Andrew Zisserman}.} \bibinfo{year}{2015}\natexlab{}.
\newblock \showarticletitle{Very deep convolutional networks for large-scale
  image recognition}. In \bibinfo{booktitle}{\emph{Proceedings of ICLR}}.
\newblock


\bibitem[\protect\citeauthoryear{Singh, Natarajan, Shah, Jiang, Chen, Batra,
  Parikh, and Rohrbach}{Singh et~al\mbox{.}}{2019}]%
        {singh2019towards}
\bibfield{author}{\bibinfo{person}{Amanpreet Singh}, \bibinfo{person}{Vivek
  Natarajan}, \bibinfo{person}{Meet Shah}, \bibinfo{person}{Yu Jiang},
  \bibinfo{person}{Xinlei Chen}, \bibinfo{person}{Dhruv Batra},
  \bibinfo{person}{Devi Parikh}, {and} \bibinfo{person}{Marcus Rohrbach}.}
  \bibinfo{year}{2019}\natexlab{}.
\newblock \showarticletitle{Towards {VQA} models that can read}. In
  \bibinfo{booktitle}{\emph{Proceedings of CVPR}}. \bibinfo{pages}{8317--8326}.
\newblock


\bibitem[\protect\citeauthoryear{Singh, Mishra, Dabral, and Jawahar}{Singh
  et~al\mbox{.}}{2016}]%
        {singh2016simple}
\bibfield{author}{\bibinfo{person}{Ajeet~Kumar Singh}, \bibinfo{person}{Anand
  Mishra}, \bibinfo{person}{Pranav Dabral}, {and} \bibinfo{person}{CV
  Jawahar}.} \bibinfo{year}{2016}\natexlab{}.
\newblock \showarticletitle{A simple and effective solution for script
  identification in the wild}. In \bibinfo{booktitle}{\emph{IAPR Workshop on
  Document Analysis Systems (DAS)}}. \bibinfo{pages}{428--433}.
\newblock


\bibitem[\protect\citeauthoryear{Su and Lu}{Su and Lu}{2014}]%
        {su2014accurate}
\bibfield{author}{\bibinfo{person}{Bolan Su} {and} \bibinfo{person}{Shijian
  Lu}.} \bibinfo{year}{2014}\natexlab{}.
\newblock \showarticletitle{Accurate scene text recognition based on recurrent
  neural network}. In \bibinfo{booktitle}{\emph{Proceedings of ACCV}}.
  \bibinfo{pages}{35--48}.
\newblock


\bibitem[\protect\citeauthoryear{Su and Lu}{Su and Lu}{2017}]%
        {su2017accurate}
\bibfield{author}{\bibinfo{person}{Bolan Su} {and} \bibinfo{person}{Shijian
  Lu}.} \bibinfo{year}{2017}\natexlab{}.
\newblock \showarticletitle{Accurate recognition of words in scenes without
  character segmentation using recurrent neural network}.
\newblock \bibinfo{journal}{\emph{Pattern Recognition}}  \bibinfo{volume}{63}
  (\bibinfo{year}{2017}), \bibinfo{pages}{397--405}.
\newblock


\bibitem[\protect\citeauthoryear{Sun, Liu, Liu, Han, Ding, and Liu}{Sun
  et~al\mbox{.}}{2019a}]%
        {sun2019chinese}
\bibfield{author}{\bibinfo{person}{Yipeng Sun}, \bibinfo{person}{Jiaming Liu},
  \bibinfo{person}{Wei Liu}, \bibinfo{person}{Junyu Han},
  \bibinfo{person}{Errui Ding}, {and} \bibinfo{person}{Jingtuo Liu}.}
  \bibinfo{year}{2019}\natexlab{a}.
\newblock \showarticletitle{Chinese Street View Text: Large-scale Chinese Text
  Reading with Partially Supervised Learning}. In
  \bibinfo{booktitle}{\emph{Proceedings of ICCV}}. \bibinfo{pages}{9086--9095}.
\newblock


\bibitem[\protect\citeauthoryear{Sun, Ni, Chng, Liu, Luo, Ng, Han, Ding, Liu,
  Karatzas, et~al\mbox{.}}{Sun et~al\mbox{.}}{2019b}]%
        {sun2019icdar}
\bibfield{author}{\bibinfo{person}{Yipeng Sun}, \bibinfo{person}{Zihan Ni},
  \bibinfo{person}{Chee-Kheng Chng}, \bibinfo{person}{Yuliang Liu},
  \bibinfo{person}{Canjie Luo}, \bibinfo{person}{Chun~Chet Ng},
  \bibinfo{person}{Junyu Han}, \bibinfo{person}{Errui Ding},
  \bibinfo{person}{Jingtuo Liu}, \bibinfo{person}{Dimosthenis Karatzas},
  {et~al\mbox{.}}} \bibinfo{year}{2019}\natexlab{b}.
\newblock \showarticletitle{{ICDAR} 2019 Competition on Large-scale Street View
  Text with Partial Labeling--{RRC-LSVT}}. In
  \bibinfo{booktitle}{\emph{Proceedings of ICDAR}}.
  \bibinfo{pages}{1557--1562}.
\newblock


\bibitem[\protect\citeauthoryear{Tang and Wu}{Tang and Wu}{2018}]%
        {tang2018scene}
\bibfield{author}{\bibinfo{person}{Youbao Tang} {and}
  \bibinfo{person}{Xiangqian Wu}.} \bibinfo{year}{2018}\natexlab{}.
\newblock \showarticletitle{Scene text detection using superpixel-based stroke
  feature transform and deep learning based region classification}.
\newblock \bibinfo{journal}{\emph{IEEE Transactions on Multimedia}}
  \bibinfo{volume}{20}, \bibinfo{number}{9} (\bibinfo{year}{2018}),
  \bibinfo{pages}{2276--2288}.
\newblock


\bibitem[\protect\citeauthoryear{Tian, Yin, Su, and Hao}{Tian
  et~al\mbox{.}}{2018}]%
        {shu2018aunified}
\bibfield{author}{\bibinfo{person}{Shu Tian}, \bibinfo{person}{Xu{-}Cheng Yin},
  \bibinfo{person}{Ya Su}, {and} \bibinfo{person}{Hong{-}Wei Hao}.}
  \bibinfo{year}{2018}\natexlab{}.
\newblock \showarticletitle{A Unified Framework for Tracking Based Text
  Detection and Recognition from Web Videos}.
\newblock \bibinfo{journal}{\emph{IEEE Trans. Pattern Anal. Mach. Intell}}
  \bibinfo{volume}{40}, \bibinfo{number}{3} (\bibinfo{year}{2018}),
  \bibinfo{pages}{542--554}.
\newblock


\bibitem[\protect\citeauthoryear{Tsai, Chen, Chen, Schroth, Grzeszczuk, and
  Girod}{Tsai et~al\mbox{.}}{2011}]%
        {tsai2011mobile}
\bibfield{author}{\bibinfo{person}{Sam~S Tsai}, \bibinfo{person}{Huizhong
  Chen}, \bibinfo{person}{David Chen}, \bibinfo{person}{Georg Schroth},
  \bibinfo{person}{Radek Grzeszczuk}, {and} \bibinfo{person}{Bernd Girod}.}
  \bibinfo{year}{2011}\natexlab{}.
\newblock \showarticletitle{Mobile visual search on printed documents using
  text and low bit-rate features}. In \bibinfo{booktitle}{\emph{Proceedings of
  ICIP}}. \bibinfo{pages}{2601--2604}.
\newblock


\bibitem[\protect\citeauthoryear{Uchida}{Uchida}{2014}]%
        {uchida2014text}
\bibfield{author}{\bibinfo{person}{Seiichi Uchida}.}
  \bibinfo{year}{2014}\natexlab{}.
\newblock \showarticletitle{Text localization and recognition in images and
  video}.
\newblock \bibinfo{journal}{\emph{Handbook of Document Image Processing and
  Recognition}} (\bibinfo{year}{2014}), \bibinfo{pages}{843--883}.
\newblock


\bibitem[\protect\citeauthoryear{Unnikrishnan and Smith}{Unnikrishnan and
  Smith}{2009}]%
        {unnikrishnan2009combined}
\bibfield{author}{\bibinfo{person}{Ranjith Unnikrishnan} {and}
  \bibinfo{person}{Ray Smith}.} \bibinfo{year}{2009}\natexlab{}.
\newblock \showarticletitle{Combined script and page orientation estimation
  using the Tesseract {OCR} engine}. In \bibinfo{booktitle}{\emph{Proceedings
  of the International Workshop on Multilingual OCR}}. \bibinfo{pages}{6}.
\newblock


\bibitem[\protect\citeauthoryear{Vaswani, Shazeer, Parmar, Uszkoreit, Jones,
  Gomez, Kaiser, and Polosukhin}{Vaswani et~al\mbox{.}}{2017}]%
        {vaswani2017attention}
\bibfield{author}{\bibinfo{person}{Ashish Vaswani}, \bibinfo{person}{Noam
  Shazeer}, \bibinfo{person}{Niki Parmar}, \bibinfo{person}{Jakob Uszkoreit},
  \bibinfo{person}{Llion Jones}, \bibinfo{person}{Aidan~N Gomez},
  \bibinfo{person}{{\L}ukasz Kaiser}, {and} \bibinfo{person}{Illia
  Polosukhin}.} \bibinfo{year}{2017}\natexlab{}.
\newblock \showarticletitle{Attention is all you need}. In
  \bibinfo{booktitle}{\emph{Proceedings of NIPS}}. \bibinfo{pages}{5998--6008}.
\newblock


\bibitem[\protect\citeauthoryear{Veit, Matera, Neumann, Matas, and
  Belongie}{Veit et~al\mbox{.}}{2016}]%
        {veit2016coco}
\bibfield{author}{\bibinfo{person}{Andreas Veit}, \bibinfo{person}{Tomas
  Matera}, \bibinfo{person}{Lukas Neumann}, \bibinfo{person}{Jiri Matas}, {and}
  \bibinfo{person}{Serge Belongie}.} \bibinfo{year}{2016}\natexlab{}.
\newblock \showarticletitle{Coco-text: Dataset and benchmark for text detection
  and recognition in natural images}.
\newblock \bibinfo{journal}{\emph{CoRR abs/1601.07140}} (\bibinfo{year}{2016}).
\newblock


\bibitem[\protect\citeauthoryear{Von~Ahn, Maurer, McMillen, Abraham, and
  Blum}{Von~Ahn et~al\mbox{.}}{2008}]%
        {von2008recaptcha}
\bibfield{author}{\bibinfo{person}{Luis Von~Ahn}, \bibinfo{person}{Benjamin
  Maurer}, \bibinfo{person}{Colin McMillen}, \bibinfo{person}{David Abraham},
  {and} \bibinfo{person}{Manuel Blum}.} \bibinfo{year}{2008}\natexlab{}.
\newblock \showarticletitle{{Recaptcha}: Human-based character recognition via
  web security measures}.
\newblock \bibinfo{journal}{\emph{Science}} \bibinfo{volume}{321},
  \bibinfo{number}{5895} (\bibinfo{year}{2008}), \bibinfo{pages}{1465--1468}.
\newblock


\bibitem[\protect\citeauthoryear{Wan, He, Chen, Bai, and Yao}{Wan
  et~al\mbox{.}}{2020}]%
        {wan2019textscanner}
\bibfield{author}{\bibinfo{person}{Zhaoyi Wan}, \bibinfo{person}{Mingling He},
  \bibinfo{person}{Haoran Chen}, \bibinfo{person}{Xiang Bai}, {and}
  \bibinfo{person}{Cong Yao}.} \bibinfo{year}{2020}\natexlab{}.
\newblock \showarticletitle{TextScanner: Reading Characters in Order for Robust
  Scene Text Recognition}. In \bibinfo{booktitle}{\emph{Proceedings of AAAI}}.
\newblock


\bibitem[\protect\citeauthoryear{Wan, Xie, Liu, Bai, and Yao}{Wan
  et~al\mbox{.}}{2019}]%
        {wan2019arXiv}
\bibfield{author}{\bibinfo{person}{Zhaoyi Wan}, \bibinfo{person}{Fengming Xie},
  \bibinfo{person}{Yibo Liu}, \bibinfo{person}{Xiang Bai}, {and}
  \bibinfo{person}{Cong Yao}.} \bibinfo{year}{2019}\natexlab{}.
\newblock \showarticletitle{{2D-CTC} for Scene Text Recognition}.
\newblock \bibinfo{journal}{\emph{CoRR abs/1907.09705}} (\bibinfo{year}{2019}).
\newblock


\bibitem[\protect\citeauthoryear{Wang, Yin, and Liu}{Wang
  et~al\mbox{.}}{2018b}]%
        {wang2018memory}
\bibfield{author}{\bibinfo{person}{Cong Wang}, \bibinfo{person}{Fei Yin}, {and}
  \bibinfo{person}{Cheng-Lin Liu}.} \bibinfo{year}{2018}\natexlab{b}.
\newblock \showarticletitle{Memory-Augmented Attention Model for Scene Text
  Recognition}. In \bibinfo{booktitle}{\emph{Proceedings of ICFHR}}.
  \bibinfo{pages}{62--67}.
\newblock


\bibitem[\protect\citeauthoryear{Wang, Lu, Zhang, Yang, Bai, Xu, He, Wang, and
  Liu}{Wang et~al\mbox{.}}{2020b}]%
        {wang2019all}
\bibfield{author}{\bibinfo{person}{Hao Wang}, \bibinfo{person}{Pu Lu},
  \bibinfo{person}{Hui Zhang}, \bibinfo{person}{Mingkun Yang},
  \bibinfo{person}{Xiang Bai}, \bibinfo{person}{Yongchao Xu},
  \bibinfo{person}{Mengchao He}, \bibinfo{person}{Yongpan Wang}, {and}
  \bibinfo{person}{Wenyu Liu}.} \bibinfo{year}{2020}\natexlab{b}.
\newblock \showarticletitle{All You Need Is Boundary: Toward Arbitrary-Shaped
  Text Spotting}. In \bibinfo{booktitle}{\emph{Proceedings of AAAI}}.
\newblock


\bibitem[\protect\citeauthoryear{Wang and Hu}{Wang and Hu}{2017}]%
        {wang2017gated}
\bibfield{author}{\bibinfo{person}{Jianfeng Wang} {and}
  \bibinfo{person}{Xiaolin Hu}.} \bibinfo{year}{2017}\natexlab{}.
\newblock \showarticletitle{Gated recurrent convolution neural network for
  {OCR}}. In \bibinfo{booktitle}{\emph{Proceedings of NIPS}}.
  \bibinfo{pages}{335--344}.
\newblock


\bibitem[\protect\citeauthoryear{Wang, Babenko, and Belongie}{Wang
  et~al\mbox{.}}{2011}]%
        {wang2011end}
\bibfield{author}{\bibinfo{person}{Kai Wang}, \bibinfo{person}{Boris Babenko},
  {and} \bibinfo{person}{Serge Belongie}.} \bibinfo{year}{2011}\natexlab{}.
\newblock \showarticletitle{End-to-end scene text recognition}. In
  \bibinfo{booktitle}{\emph{Proceedings of ICCV}}. \bibinfo{pages}{1457--1464}.
\newblock


\bibitem[\protect\citeauthoryear{Wang and Belongie}{Wang and Belongie}{2010}]%
        {wang2010word}
\bibfield{author}{\bibinfo{person}{Kai Wang} {and} \bibinfo{person}{Serge
  Belongie}.} \bibinfo{year}{2010}\natexlab{}.
\newblock \showarticletitle{Word spotting in the wild}. In
  \bibinfo{booktitle}{\emph{Proceedings of ECCV}}. \bibinfo{pages}{591--604}.
\newblock


\bibitem[\protect\citeauthoryear{Wang, Yang, Li, Deng, Shen, and Zhang}{Wang
  et~al\mbox{.}}{2019e}]%
        {wang2019simple}
\bibfield{author}{\bibinfo{person}{Peng Wang}, \bibinfo{person}{Lu Yang},
  \bibinfo{person}{Hui Li}, \bibinfo{person}{Yuyan Deng},
  \bibinfo{person}{Chunhua Shen}, {and} \bibinfo{person}{Yanning Zhang}.}
  \bibinfo{year}{2019}\natexlab{e}.
\newblock \showarticletitle{A Simple and Robust Convolutional-Attention Network
  for Irregular Text Recognition}.
\newblock \bibinfo{journal}{\emph{CoRR abs/1904.01375}} (\bibinfo{year}{2019}).
\newblock


\bibitem[\protect\citeauthoryear{Wang, Jia, He, Lu, Blumenstein, Huang, and
  Lyu}{Wang et~al\mbox{.}}{2019a}]%
        {wang2019reelfa}
\bibfield{author}{\bibinfo{person}{Qingqing Wang}, \bibinfo{person}{Wenjing
  Jia}, \bibinfo{person}{Xiangjian He}, \bibinfo{person}{Yue Lu},
  \bibinfo{person}{Michael Blumenstein}, \bibinfo{person}{Ye Huang}, {and}
  \bibinfo{person}{Shujing Lyu}.} \bibinfo{year}{2019}\natexlab{a}.
\newblock \showarticletitle{{ReELFA}: A Scene Text Recognizer with Encoded
  Location and Focused Attention}. In \bibinfo{booktitle}{\emph{Proceedings of
  ICDAR: Workshops}}. \bibinfo{pages}{71--76}.
\newblock


\bibitem[\protect\citeauthoryear{Wang, Liu, Chanussot, and Li}{Wang
  et~al\mbox{.}}{2018a}]%
        {wang2018scene}
\bibfield{author}{\bibinfo{person}{Qi Wang}, \bibinfo{person}{Shaoteng Liu},
  \bibinfo{person}{Jocelyn Chanussot}, {and} \bibinfo{person}{Xuelong Li}.}
  \bibinfo{year}{2018}\natexlab{a}.
\newblock \showarticletitle{Scene classification with recurrent attention of
  {VHR} remote sensing images}.
\newblock \bibinfo{journal}{\emph{IEEE Transactions on Geoscience and Remote
  Sensing}} \bibinfo{volume}{57}, \bibinfo{number}{2} (\bibinfo{year}{2018}),
  \bibinfo{pages}{1155--1167}.
\newblock


\bibitem[\protect\citeauthoryear{Wang, Wang, Qin, Zhao, and Tang}{Wang
  et~al\mbox{.}}{2019c}]%
        {wang2019scene}
\bibfield{author}{\bibinfo{person}{Siwei Wang}, \bibinfo{person}{Yongtao Wang},
  \bibinfo{person}{Xiaoran Qin}, \bibinfo{person}{Qijie Zhao}, {and}
  \bibinfo{person}{Zhi Tang}.} \bibinfo{year}{2019}\natexlab{c}.
\newblock \showarticletitle{Scene Text Recognition via Gated Cascade
  Attention}. In \bibinfo{booktitle}{\emph{Proceedings of ICME}}.
  \bibinfo{pages}{1018--1023}.
\newblock


\bibitem[\protect\citeauthoryear{Wang, Wu, Coates, and Ng}{Wang
  et~al\mbox{.}}{2012}]%
        {wang2012end}
\bibfield{author}{\bibinfo{person}{Tao Wang}, \bibinfo{person}{David~J Wu},
  \bibinfo{person}{Adam Coates}, {and} \bibinfo{person}{Andrew~Y Ng}.}
  \bibinfo{year}{2012}\natexlab{}.
\newblock \showarticletitle{End-to-end text recognition with convolutional
  neural networks}. In \bibinfo{booktitle}{\emph{Proceedings of ICPR}}.
  \bibinfo{pages}{3304--3308}.
\newblock


\bibitem[\protect\citeauthoryear{Wang, Zhu, Jin, Luo, Chen, Wu, Wang, and
  Cai}{Wang et~al\mbox{.}}{2020d}]%
        {wang2019decoupled}
\bibfield{author}{\bibinfo{person}{Tianwei Wang}, \bibinfo{person}{Yuanzhi
  Zhu}, \bibinfo{person}{Lianwen Jin}, \bibinfo{person}{Canjie Luo},
  \bibinfo{person}{Xiaoxue Chen}, \bibinfo{person}{Yaqiang Wu},
  \bibinfo{person}{Qianying Wang}, {and} \bibinfo{person}{Mingxiang Cai}.}
  \bibinfo{year}{2020}\natexlab{d}.
\newblock \showarticletitle{Decoupled Attention Network for Text Recognition}.
  In \bibinfo{booktitle}{\emph{Proceedings of AAAI}}.
\newblock


\bibitem[\protect\citeauthoryear{Wang, Xie, Sun, Wang, Tian, Shen, and
  Luo}{Wang et~al\mbox{.}}{2019d}]%
        {wang2019textsr}
\bibfield{author}{\bibinfo{person}{Wenjia Wang}, \bibinfo{person}{Enze Xie},
  \bibinfo{person}{Peize Sun}, \bibinfo{person}{Wenhai Wang},
  \bibinfo{person}{Lixun Tian}, \bibinfo{person}{Chunhua Shen}, {and}
  \bibinfo{person}{Ping Luo}.} \bibinfo{year}{2019}\natexlab{d}.
\newblock \showarticletitle{{TextSR}: Content-Aware Text Super-Resolution
  Guided by Recognition}.
\newblock \bibinfo{journal}{\emph{CoRR abs/1909.07113}} (\bibinfo{year}{2019}).
\newblock


\bibitem[\protect\citeauthoryear{Wang, Liu, Shen, Ng, Luo, Jin, Chan, van~den
  Hengel, and Wang}{Wang et~al\mbox{.}}{2020a}]%
        {Xinyu2020On}
\bibfield{author}{\bibinfo{person}{Xinyu Wang}, \bibinfo{person}{Yuliang Liu},
  \bibinfo{person}{Chunhua Shen}, \bibinfo{person}{Chun~Chet Ng},
  \bibinfo{person}{Canjie Luo}, \bibinfo{person}{Lianwen Jin},
  \bibinfo{person}{Chee~Seng Chan}, \bibinfo{person}{Anton van~den Hengel},
  {and} \bibinfo{person}{Liangwei Wang}.} \bibinfo{year}{2020}\natexlab{a}.
\newblock \showarticletitle{On the General Value of Evidence, and Bilingual
  Scene-Text Visual Question Answering}. In
  \bibinfo{booktitle}{\emph{Proceedings of CVPR}}.
\newblock


\bibitem[\protect\citeauthoryear{Wang, Su, and Qian}{Wang
  et~al\mbox{.}}{2019b}]%
        {yuyang2019text}
\bibfield{author}{\bibinfo{person}{Yuyang Wang}, \bibinfo{person}{Feng Su},
  {and} \bibinfo{person}{Ye Qian}.} \bibinfo{year}{2019}\natexlab{b}.
\newblock \showarticletitle{Text-Attentional Conditional Generative Adversarial
  Network for Super-Resolution of Text Images}. In
  \bibinfo{booktitle}{\emph{Proceedings of ICME}}. \bibinfo{pages}{1024--1029}.
\newblock


\bibitem[\protect\citeauthoryear{Wang, Xie, Zha, Tian, Fu, and Zhang}{Wang
  et~al\mbox{.}}{2020c}]%
        {wang2020r}
\bibfield{author}{\bibinfo{person}{Yuxin Wang}, \bibinfo{person}{Hongtao Xie},
  \bibinfo{person}{Zheng-Jun Zha}, \bibinfo{person}{Youliang Tian},
  \bibinfo{person}{Zilong Fu}, {and} \bibinfo{person}{Yongdong Zhang}.}
  \bibinfo{year}{2020}\natexlab{c}.
\newblock \showarticletitle{R-Net: A Relationship Network for Efficient and
  Accurate Scene Text Detection}.
\newblock \bibinfo{journal}{\emph{IEEE Transactions on Multimedia}}
  (\bibinfo{year}{2020}).
\newblock


\bibitem[\protect\citeauthoryear{Warps}{Warps}{1989}]%
        {warps1989thin}
\bibfield{author}{\bibinfo{person}{Fred L Bookstein~Principal Warps}.}
  \bibinfo{year}{1989}\natexlab{}.
\newblock \showarticletitle{Thin-Plate Splines and the Decompositions of
  Deformations}.
\newblock \bibinfo{journal}{\emph{IEEE Trans. Pattern Anal. Mach. Intell}}
  \bibinfo{volume}{11}, \bibinfo{number}{6} (\bibinfo{year}{1989}).
\newblock


\bibitem[\protect\citeauthoryear{Wu, Zhang, Liu, Han, Liu, Ding, and Bai}{Wu
  et~al\mbox{.}}{2019}]%
        {wu2019editing}
\bibfield{author}{\bibinfo{person}{Liang Wu}, \bibinfo{person}{Chengquan
  Zhang}, \bibinfo{person}{Jiaming Liu}, \bibinfo{person}{Junyu Han},
  \bibinfo{person}{Jingtuo Liu}, \bibinfo{person}{Errui Ding}, {and}
  \bibinfo{person}{Xiang Bai}.} \bibinfo{year}{2019}\natexlab{}.
\newblock \showarticletitle{Editing Text in the Wild}. In
  \bibinfo{booktitle}{\emph{Proceedings of {ACM} International Conference on
  Multimedia}}. \bibinfo{pages}{1500--1508}.
\newblock


\bibitem[\protect\citeauthoryear{Wu and Natarajan}{Wu and Natarajan}{2017}]%
        {wu2017self}
\bibfield{author}{\bibinfo{person}{Yue Wu} {and} \bibinfo{person}{Prem
  Natarajan}.} \bibinfo{year}{2017}\natexlab{}.
\newblock \showarticletitle{Self-organized text detection with minimal
  post-processing via border learning}. In
  \bibinfo{booktitle}{\emph{Proceedings of ICCV}}. \bibinfo{pages}{5000--5009}.
\newblock


\bibitem[\protect\citeauthoryear{Xie, Fang, Zha, Yang, Li, and Zhang}{Xie
  et~al\mbox{.}}{2019a}]%
        {xie2019convolutional}
\bibfield{author}{\bibinfo{person}{Hongtao Xie}, \bibinfo{person}{Shancheng
  Fang}, \bibinfo{person}{Zheng-Jun Zha}, \bibinfo{person}{Yating Yang},
  \bibinfo{person}{Yan Li}, {and} \bibinfo{person}{Yongdong Zhang}.}
  \bibinfo{year}{2019}\natexlab{a}.
\newblock \showarticletitle{Convolutional Attention Networks for Scene Text
  Recognition}.
\newblock \bibinfo{journal}{\emph{ACM Transactions on Multimedia Computing,
  Communications, and Applications (TOMM)}} \bibinfo{volume}{15},
  \bibinfo{number}{1s} (\bibinfo{year}{2019}), \bibinfo{pages}{3}.
\newblock


\bibitem[\protect\citeauthoryear{Xie, Ahmad, Jin, Liu, and Zhang}{Xie
  et~al\mbox{.}}{2018}]%
        {xie2018new}
\bibfield{author}{\bibinfo{person}{Lele Xie}, \bibinfo{person}{Tasweer Ahmad},
  \bibinfo{person}{Lianwen Jin}, \bibinfo{person}{Yuliang Liu}, {and}
  \bibinfo{person}{Sheng Zhang}.} \bibinfo{year}{2018}\natexlab{}.
\newblock \showarticletitle{A new {CNN}-based method for multi-directional car
  license plate detection}.
\newblock \bibinfo{journal}{\emph{IEEE Transactions on Intelligent
  Transportation Systems}} \bibinfo{volume}{19}, \bibinfo{number}{2}
  (\bibinfo{year}{2018}), \bibinfo{pages}{507--517}.
\newblock


\bibitem[\protect\citeauthoryear{Xie, Liu, Jin, and Xie}{Xie
  et~al\mbox{.}}{2019c}]%
        {xie2019derpn}
\bibfield{author}{\bibinfo{person}{Lele Xie}, \bibinfo{person}{Yuliang Liu},
  \bibinfo{person}{Lianwen Jin}, {and} \bibinfo{person}{Zecheng Xie}.}
  \bibinfo{year}{2019}\natexlab{c}.
\newblock \showarticletitle{{DeRPN}: Taking a further step toward more general
  object detection}. In \bibinfo{booktitle}{\emph{Proceedings of AAAI}}.
  \bibinfo{pages}{9046--9053}.
\newblock


\bibitem[\protect\citeauthoryear{Xie, Huang, Zhu, Jin, Liu, and Xie}{Xie
  et~al\mbox{.}}{2019b}]%
        {xie2019aggregation}
\bibfield{author}{\bibinfo{person}{Zecheng Xie}, \bibinfo{person}{Yaoxiong
  Huang}, \bibinfo{person}{Yuanzhi Zhu}, \bibinfo{person}{Lianwen Jin},
  \bibinfo{person}{Yuliang Liu}, {and} \bibinfo{person}{Lele Xie}.}
  \bibinfo{year}{2019}\natexlab{b}.
\newblock \showarticletitle{Aggregation Cross-Entropy for Sequence
  Recognition}. In \bibinfo{booktitle}{\emph{Proceedings of CVPR}}.
  \bibinfo{pages}{6538--6547}.
\newblock


\bibitem[\protect\citeauthoryear{Xie, Sun, Jin, Ni, and Lyons}{Xie
  et~al\mbox{.}}{2017}]%
        {xie2017learning}
\bibfield{author}{\bibinfo{person}{Zecheng Xie}, \bibinfo{person}{Zenghui Sun},
  \bibinfo{person}{Lianwen Jin}, \bibinfo{person}{Hao Ni}, {and}
  \bibinfo{person}{Terry Lyons}.} \bibinfo{year}{2017}\natexlab{}.
\newblock \showarticletitle{Learning spatial-semantic context with fully
  convolutional recurrent network for online handwritten Chinese text
  recognition}.
\newblock \bibinfo{journal}{\emph{IEEE Trans. Pattern Anal. Mach. Intell}}
  \bibinfo{volume}{40}, \bibinfo{number}{8} (\bibinfo{year}{2017}),
  \bibinfo{pages}{1903--1917}.
\newblock


\bibitem[\protect\citeauthoryear{Xing, Tian, Huang, and Scott}{Xing
  et~al\mbox{.}}{2019}]%
        {xing2019convo}
\bibfield{author}{\bibinfo{person}{Linjie Xing}, \bibinfo{person}{Zhi Tian},
  \bibinfo{person}{Weilin Huang}, {and} \bibinfo{person}{Matthew~R. Scott}.}
  \bibinfo{year}{2019}\natexlab{}.
\newblock \showarticletitle{Convolutional Character Networks}. In
  \bibinfo{booktitle}{\emph{Proceedings of ICCV}}. \bibinfo{pages}{9125--9135}.
\newblock


\bibitem[\protect\citeauthoryear{Xing, Qi, Yuan, Li, Zhang, Fu, Xiong, Hu, and
  Peng}{Xing et~al\mbox{.}}{2018}]%
        {xing2018gene}
\bibfield{author}{\bibinfo{person}{Wenhui Xing}, \bibinfo{person}{Junsheng Qi},
  \bibinfo{person}{Xiaohui Yuan}, \bibinfo{person}{Lin Li},
  \bibinfo{person}{Xiaoyu Zhang}, \bibinfo{person}{Yuhua Fu},
  \bibinfo{person}{Shengwu Xiong}, \bibinfo{person}{Lun Hu}, {and}
  \bibinfo{person}{Jing Peng}.} \bibinfo{year}{2018}\natexlab{}.
\newblock \showarticletitle{A gene--phenotype relationship extraction pipeline
  from the biomedical literature using a representation learning approach}.
\newblock \bibinfo{journal}{\emph{Bioinformatics}} \bibinfo{volume}{34},
  \bibinfo{number}{13} (\bibinfo{year}{2018}), \bibinfo{pages}{i386--i394}.
\newblock


\bibitem[\protect\citeauthoryear{Xu and Jia}{Xu and Jia}{2010}]%
        {xu2010two}
\bibfield{author}{\bibinfo{person}{Li Xu} {and} \bibinfo{person}{Jiaya Jia}.}
  \bibinfo{year}{2010}\natexlab{}.
\newblock \showarticletitle{Two-phase kernel estimation for robust motion
  deblurring}. In \bibinfo{booktitle}{\emph{Proceedings of ECCV}}.
  \bibinfo{pages}{157--170}.
\newblock


\bibitem[\protect\citeauthoryear{Xu, Wang, Zhou, Wang, Yang, and Bai}{Xu
  et~al\mbox{.}}{2019}]%
        {xu2019textfield}
\bibfield{author}{\bibinfo{person}{Yongchao Xu}, \bibinfo{person}{Yukang Wang},
  \bibinfo{person}{Wei Zhou}, \bibinfo{person}{Yongpan Wang},
  \bibinfo{person}{Zhibo Yang}, {and} \bibinfo{person}{Xiang Bai}.}
  \bibinfo{year}{2019}\natexlab{}.
\newblock \showarticletitle{TextField: learning a deep direction field for
  irregular scene text detection}.
\newblock \bibinfo{journal}{\emph{IEEE Transactions on Image Processing}}
  \bibinfo{volume}{28}, \bibinfo{number}{11} (\bibinfo{year}{2019}),
  \bibinfo{pages}{5566--5579}.
\newblock


\bibitem[\protect\citeauthoryear{Yan, Xie, Chen, Zha, Hao, Zhang, and Dai}{Yan
  et~al\mbox{.}}{2018}]%
        {yan2018fast}
\bibfield{author}{\bibinfo{person}{Chenggang Yan}, \bibinfo{person}{Hongtao
  Xie}, \bibinfo{person}{Jianjun Chen}, \bibinfo{person}{Zhengjun Zha},
  \bibinfo{person}{Xinhong Hao}, \bibinfo{person}{Yongdong Zhang}, {and}
  \bibinfo{person}{Qionghai Dai}.} \bibinfo{year}{2018}\natexlab{}.
\newblock \showarticletitle{A fast uyghur text detector for complex background
  images}.
\newblock \bibinfo{journal}{\emph{IEEE Transactions on Multimedia}}
  \bibinfo{volume}{20}, \bibinfo{number}{12} (\bibinfo{year}{2018}),
  \bibinfo{pages}{3389--3398}.
\newblock


\bibitem[\protect\citeauthoryear{Yang, Jin, Lai, Gao, and Li}{Yang
  et~al\mbox{.}}{2019b}]%
        {yang2019fully}
\bibfield{author}{\bibinfo{person}{Fan Yang}, \bibinfo{person}{Lianwen Jin},
  \bibinfo{person}{Songxuan Lai}, \bibinfo{person}{Xue Gao}, {and}
  \bibinfo{person}{Zhaohai Li}.} \bibinfo{year}{2019}\natexlab{b}.
\newblock \showarticletitle{Fully Convolutional Sequence Recognition Network
  for Water Meter Number Reading}.
\newblock \bibinfo{journal}{\emph{IEEE Access}}  \bibinfo{volume}{7}
  (\bibinfo{year}{2019}), \bibinfo{pages}{11679--11687}.
\newblock


\bibitem[\protect\citeauthoryear{Yang, Guan, Liao, He, Bian, Bai, Yao, and
  Bai}{Yang et~al\mbox{.}}{2019a}]%
        {yang2019symmetry}
\bibfield{author}{\bibinfo{person}{Mingkun Yang}, \bibinfo{person}{Yushuo
  Guan}, \bibinfo{person}{Minghui Liao}, \bibinfo{person}{Xin He},
  \bibinfo{person}{Kaigui Bian}, \bibinfo{person}{Song Bai},
  \bibinfo{person}{Cong Yao}, {and} \bibinfo{person}{Xiang Bai}.}
  \bibinfo{year}{2019}\natexlab{a}.
\newblock \showarticletitle{Symmetry-constrained rectification network for
  scene text recognition}. In \bibinfo{booktitle}{\emph{Proceedings of ICCV}}.
  \bibinfo{pages}{9147--9156}.
\newblock


\bibitem[\protect\citeauthoryear{Yang, He, Zhou, Kifer, and Giles}{Yang
  et~al\mbox{.}}{2017}]%
        {yang2017learning}
\bibfield{author}{\bibinfo{person}{Xiao Yang}, \bibinfo{person}{Dafang He},
  \bibinfo{person}{Zihan Zhou}, \bibinfo{person}{Daniel Kifer}, {and}
  \bibinfo{person}{C~Lee Giles}.} \bibinfo{year}{2017}\natexlab{}.
\newblock \showarticletitle{Learning to read irregular text with attention
  mechanisms}. In \bibinfo{booktitle}{\emph{Proceedings of IJCAI}}.
  \bibinfo{pages}{3280--3286}.
\newblock


\bibitem[\protect\citeauthoryear{Yao, Bai, and Liu}{Yao et~al\mbox{.}}{2014a}]%
        {yao2014unified}
\bibfield{author}{\bibinfo{person}{Cong Yao}, \bibinfo{person}{Xiang Bai},
  {and} \bibinfo{person}{Wenyu Liu}.} \bibinfo{year}{2014}\natexlab{a}.
\newblock \showarticletitle{A unified framework for multioriented text
  detection and recognition}.
\newblock \bibinfo{journal}{\emph{IEEE Transactions on Image Processing}}
  \bibinfo{volume}{23}, \bibinfo{number}{11} (\bibinfo{year}{2014}),
  \bibinfo{pages}{4737--4749}.
\newblock


\bibitem[\protect\citeauthoryear{Yao, Bai, Liu, Ma, and Tu}{Yao
  et~al\mbox{.}}{2012}]%
        {yao2012detecting}
\bibfield{author}{\bibinfo{person}{Cong Yao}, \bibinfo{person}{Xiang Bai},
  \bibinfo{person}{Wenyu Liu}, \bibinfo{person}{Yi Ma}, {and}
  \bibinfo{person}{Zhuowen Tu}.} \bibinfo{year}{2012}\natexlab{}.
\newblock \showarticletitle{Detecting texts of arbitrary orientations in
  natural images}. In \bibinfo{booktitle}{\emph{Proceedings of CVPR}}.
  \bibinfo{pages}{1083--1090}.
\newblock


\bibitem[\protect\citeauthoryear{Yao, Bai, Shi, and Liu}{Yao
  et~al\mbox{.}}{2014b}]%
        {yao2014strokelets}
\bibfield{author}{\bibinfo{person}{Cong Yao}, \bibinfo{person}{Xiang Bai},
  \bibinfo{person}{Baoguang Shi}, {and} \bibinfo{person}{Wenyu Liu}.}
  \bibinfo{year}{2014}\natexlab{b}.
\newblock \showarticletitle{Strokelets: A learned multi-scale representation
  for scene text recognition}. In \bibinfo{booktitle}{\emph{Proceedings of
  CVPR}}. \bibinfo{pages}{4042--4049}.
\newblock


\bibitem[\protect\citeauthoryear{Yao, Zhang, Bai, Liu, Ma, and Tu}{Yao
  et~al\mbox{.}}{2013}]%
        {yao2013rotation}
\bibfield{author}{\bibinfo{person}{Cong Yao}, \bibinfo{person}{Xin Zhang},
  \bibinfo{person}{Xiang Bai}, \bibinfo{person}{Wenyu Liu}, \bibinfo{person}{Yi
  Ma}, {and} \bibinfo{person}{Zhuowen Tu}.} \bibinfo{year}{2013}\natexlab{}.
\newblock \showarticletitle{Rotation-invariant features for multi-oriented text
  detection in natural images}.
\newblock \bibinfo{journal}{\emph{PloS one}} \bibinfo{volume}{8},
  \bibinfo{number}{8} (\bibinfo{year}{2013}), \bibinfo{pages}{e70173}.
\newblock


\bibitem[\protect\citeauthoryear{Ye and Doermann}{Ye and Doermann}{2014}]%
        {ye2014text}
\bibfield{author}{\bibinfo{person}{Qixiang Ye} {and} \bibinfo{person}{David
  Doermann}.} \bibinfo{year}{2014}\natexlab{}.
\newblock \showarticletitle{Text detection and recognition in imagery: A
  survey}.
\newblock \bibinfo{journal}{\emph{IEEE Trans. Pattern Anal. Mach. Intell}}
  \bibinfo{volume}{37}, \bibinfo{number}{7} (\bibinfo{year}{2014}),
  \bibinfo{pages}{1480--1500}.
\newblock


\bibitem[\protect\citeauthoryear{Ye, Gao, Wang, and Zeng}{Ye
  et~al\mbox{.}}{2003}]%
        {ye2003robust}
\bibfield{author}{\bibinfo{person}{Qixiang Ye}, \bibinfo{person}{Wen Gao},
  \bibinfo{person}{Weiqiang Wang}, {and} \bibinfo{person}{Wei Zeng}.}
  \bibinfo{year}{2003}\natexlab{}.
\newblock \showarticletitle{A robust text detection algorithm in images and
  video frames}. In \bibinfo{booktitle}{\emph{Proceedings of Joint Conf. Inf.,
  Commun. Signal Process. Pac. Rim Conf. Multimedia}}. IEEE,
  \bibinfo{pages}{802--806}.
\newblock


\bibitem[\protect\citeauthoryear{Ye, Huang, Gao, and Zhao}{Ye
  et~al\mbox{.}}{2005}]%
        {ye2005fast}
\bibfield{author}{\bibinfo{person}{Qixiang Ye}, \bibinfo{person}{Qingming
  Huang}, \bibinfo{person}{Wen Gao}, {and} \bibinfo{person}{Debin Zhao}.}
  \bibinfo{year}{2005}\natexlab{}.
\newblock \showarticletitle{Fast and robust text detection in images and video
  frames}.
\newblock \bibinfo{journal}{\emph{Image and vision computing}}
  \bibinfo{volume}{23}, \bibinfo{number}{6} (\bibinfo{year}{2005}),
  \bibinfo{pages}{565--576}.
\newblock


\bibitem[\protect\citeauthoryear{Yi and Tian}{Yi and Tian}{2011}]%
        {yi2011text}
\bibfield{author}{\bibinfo{person}{Chucai Yi} {and} \bibinfo{person}{YingLi
  Tian}.} \bibinfo{year}{2011}\natexlab{}.
\newblock \showarticletitle{Text string detection from natural scenes by
  structure-based partition and grouping}.
\newblock \bibinfo{journal}{\emph{IEEE Transactions on Image Processing}}
  \bibinfo{volume}{20}, \bibinfo{number}{9} (\bibinfo{year}{2011}),
  \bibinfo{pages}{2594--2605}.
\newblock


\bibitem[\protect\citeauthoryear{Yin, Wu, Yu, and Sun}{Yin
  et~al\mbox{.}}{2019}]%
        {yin2019video}
\bibfield{author}{\bibinfo{person}{Fang Yin}, \bibinfo{person}{Rui Wu},
  \bibinfo{person}{Xiaoyang Yu}, {and} \bibinfo{person}{Guanglu Sun}.}
  \bibinfo{year}{2019}\natexlab{}.
\newblock \showarticletitle{Video text localization based on Adaboost}.
\newblock \bibinfo{journal}{\emph{Multimedia Tools and Applications}}
  \bibinfo{volume}{78}, \bibinfo{number}{5} (\bibinfo{year}{2019}),
  \bibinfo{pages}{5345--5354}.
\newblock


\bibitem[\protect\citeauthoryear{Yin, Wu, Zhang, and Liu}{Yin
  et~al\mbox{.}}{2017}]%
        {yin2017scene}
\bibfield{author}{\bibinfo{person}{Fei Yin}, \bibinfo{person}{Yi-Chao Wu},
  \bibinfo{person}{Xu-Yao Zhang}, {and} \bibinfo{person}{Cheng-Lin Liu}.}
  \bibinfo{year}{2017}\natexlab{}.
\newblock \showarticletitle{Scene text recognition with sliding convolutional
  character models}. In \bibinfo{booktitle}{\emph{Proceedings of ICCV}}.
\newblock


\bibitem[\protect\citeauthoryear{Yin, Zuo, Tian, and Liu}{Yin
  et~al\mbox{.}}{2016}]%
        {yin2016text}
\bibfield{author}{\bibinfo{person}{Xu-Cheng Yin}, \bibinfo{person}{Ze-Yu Zuo},
  \bibinfo{person}{Shu Tian}, {and} \bibinfo{person}{Cheng-Lin Liu}.}
  \bibinfo{year}{2016}\natexlab{}.
\newblock \showarticletitle{Text detection, tracking and recognition in video:
  a comprehensive survey}.
\newblock \bibinfo{journal}{\emph{IEEE Transactions on Image Processing}}
  \bibinfo{volume}{25}, \bibinfo{number}{6} (\bibinfo{year}{2016}),
  \bibinfo{pages}{2752--2773}.
\newblock


\bibitem[\protect\citeauthoryear{Yu, Li, Zhang, Han, Liu, and Ding}{Yu
  et~al\mbox{.}}{2020}]%
        {Deli2020Towards}
\bibfield{author}{\bibinfo{person}{Deli Yu}, \bibinfo{person}{Xuan Li},
  \bibinfo{person}{Chengquan Zhang}, \bibinfo{person}{Junyu Han},
  \bibinfo{person}{Jingtuo Liu}, {and} \bibinfo{person}{Errui Ding}.}
  \bibinfo{year}{2020}\natexlab{}.
\newblock \showarticletitle{Towards Accurate Scene Text Recognition with
  Semantic Reasoning Networks}. In \bibinfo{booktitle}{\emph{Proceedings of
  CVPR}}.
\newblock


\bibitem[\protect\citeauthoryear{Yuan, Zhu, Xu, Li, and Hu}{Yuan
  et~al\mbox{.}}{2018}]%
        {yuan2018chinese}
\bibfield{author}{\bibinfo{person}{Tai-Ling Yuan}, \bibinfo{person}{Zhe Zhu},
  \bibinfo{person}{Kun Xu}, \bibinfo{person}{Cheng-Jun Li}, {and}
  \bibinfo{person}{Shi-Min Hu}.} \bibinfo{year}{2018}\natexlab{}.
\newblock \showarticletitle{Chinese text in the wild}.
\newblock \bibinfo{journal}{\emph{CoRR abs/1803.00085}} (\bibinfo{year}{2018}).
\newblock


\bibitem[\protect\citeauthoryear{Yuliang, Lianwen, Shuaitao, and Sheng}{Yuliang
  et~al\mbox{.}}{2017}]%
        {yuliang2017detecting}
\bibfield{author}{\bibinfo{person}{Liu Yuliang}, \bibinfo{person}{Jin Lianwen},
  \bibinfo{person}{Zhang Shuaitao}, {and} \bibinfo{person}{Zhang Sheng}.}
  \bibinfo{year}{2017}\natexlab{}.
\newblock \showarticletitle{Detecting curve text in the wild: New dataset and
  new solution}.
\newblock \bibinfo{journal}{\emph{CoRR abs/1712.02170}} (\bibinfo{year}{2017}).
\newblock


\bibitem[\protect\citeauthoryear{Zaeem, German, and Barber}{Zaeem
  et~al\mbox{.}}{2018}]%
        {zaeem2018privacycheck}
\bibfield{author}{\bibinfo{person}{Razieh~Nokhbeh Zaeem},
  \bibinfo{person}{Rachel~L German}, {and} \bibinfo{person}{K~Suzanne Barber}.}
  \bibinfo{year}{2018}\natexlab{}.
\newblock \showarticletitle{PrivacyCheck: Automatic Summarization of Privacy
  Policies Using Data Mining}.
\newblock \bibinfo{journal}{\emph{ACM Transactions on Internet Technology
  (TOIT)}} \bibinfo{volume}{18}, \bibinfo{number}{4} (\bibinfo{year}{2018}),
  \bibinfo{pages}{53}.
\newblock


\bibitem[\protect\citeauthoryear{Zhan and Lu}{Zhan and Lu}{2019}]%
        {zhan2019esir}
\bibfield{author}{\bibinfo{person}{Fangneng Zhan} {and}
  \bibinfo{person}{Shijian Lu}.} \bibinfo{year}{2019}\natexlab{}.
\newblock \showarticletitle{{ESIR}: End-to-end scene text recognition via
  iterative image rectification}. In \bibinfo{booktitle}{\emph{Proceedings of
  CVPR}}. \bibinfo{pages}{2059--2068}.
\newblock


\bibitem[\protect\citeauthoryear{Zhan, Lu, and Xue}{Zhan et~al\mbox{.}}{2018}]%
        {zhan2018verisimilar}
\bibfield{author}{\bibinfo{person}{Fangneng Zhan}, \bibinfo{person}{Shijian
  Lu}, {and} \bibinfo{person}{Chuhui Xue}.} \bibinfo{year}{2018}\natexlab{}.
\newblock \showarticletitle{Verisimilar image synthesis for accurate detection
  and recognition of texts in scenes}. In \bibinfo{booktitle}{\emph{Proceedings
  of ECCV}}. \bibinfo{pages}{249--266}.
\newblock


\bibitem[\protect\citeauthoryear{Zhan, Zhu, and Lu}{Zhan et~al\mbox{.}}{2019}]%
        {zhan2019spatial}
\bibfield{author}{\bibinfo{person}{Fangneng Zhan}, \bibinfo{person}{Hongyuan
  Zhu}, {and} \bibinfo{person}{Shijian Lu}.} \bibinfo{year}{2019}\natexlab{}.
\newblock \showarticletitle{Spatial fusion gan for image synthesis}. In
  \bibinfo{booktitle}{\emph{Proceedings of CVPR}}. \bibinfo{pages}{3653--3662}.
\newblock


\bibitem[\protect\citeauthoryear{Zhang, Zhao, Song, and Guo}{Zhang
  et~al\mbox{.}}{2013}]%
        {zhang2013text}
\bibfield{author}{\bibinfo{person}{Honggang Zhang}, \bibinfo{person}{Kaili
  Zhao}, \bibinfo{person}{Yi-Zhe Song}, {and} \bibinfo{person}{Jun Guo}.}
  \bibinfo{year}{2013}\natexlab{}.
\newblock \showarticletitle{Text extraction from natural scene image: A
  survey}.
\newblock \bibinfo{journal}{\emph{Neurocomputing}}  \bibinfo{volume}{122}
  (\bibinfo{year}{2013}), \bibinfo{pages}{310--323}.
\newblock


\bibitem[\protect\citeauthoryear{Zhang, Liu, Jin, and Luo}{Zhang
  et~al\mbox{.}}{2018}]%
        {zhang2018feature}
\bibfield{author}{\bibinfo{person}{Sheng Zhang}, \bibinfo{person}{Yuliang Liu},
  \bibinfo{person}{Lianwen Jin}, {and} \bibinfo{person}{Canjie Luo}.}
  \bibinfo{year}{2018}\natexlab{}.
\newblock \showarticletitle{Feature enhancement network: A refined scene text
  detector}. In \bibinfo{booktitle}{\emph{Proceedings of AAAI}}.
  \bibinfo{pages}{2612--2619}.
\newblock


\bibitem[\protect\citeauthoryear{Zhang, Nie, Liu, Xu, Zhang, and Shen}{Zhang
  et~al\mbox{.}}{2019}]%
        {zhang2019sequence}
\bibfield{author}{\bibinfo{person}{Yaping Zhang}, \bibinfo{person}{Shuai Nie},
  \bibinfo{person}{Wenju Liu}, \bibinfo{person}{Xing Xu},
  \bibinfo{person}{Dongxiang Zhang}, {and} \bibinfo{person}{Heng~Tao Shen}.}
  \bibinfo{year}{2019}\natexlab{}.
\newblock \showarticletitle{Sequence-To-Sequence Domain Adaptation Network for
  Robust Text Image Recognition}. In \bibinfo{booktitle}{\emph{Proceedings of
  CVPR}}. \bibinfo{pages}{2740--2749}.
\newblock


\bibitem[\protect\citeauthoryear{Zhao, Lin, Fu, Hu, Liu, and Huang}{Zhao
  et~al\mbox{.}}{2010}]%
        {zhao2010text}
\bibfield{author}{\bibinfo{person}{Xu Zhao}, \bibinfo{person}{Kai-Hsiang Lin},
  \bibinfo{person}{Yun Fu}, \bibinfo{person}{Yuxiao Hu},
  \bibinfo{person}{Yuncai Liu}, {and} \bibinfo{person}{Thomas~S Huang}.}
  \bibinfo{year}{2010}\natexlab{}.
\newblock \showarticletitle{Text from corners: a novel approach to detect text
  and caption in videos}.
\newblock \bibinfo{journal}{\emph{IEEE Transactions on Image Processing}}
  \bibinfo{volume}{20}, \bibinfo{number}{3} (\bibinfo{year}{2010}),
  \bibinfo{pages}{790--799}.
\newblock


\bibitem[\protect\citeauthoryear{Zhong, Zhang, and Jain}{Zhong
  et~al\mbox{.}}{2000}]%
        {zhong2000automatic}
\bibfield{author}{\bibinfo{person}{Yu Zhong}, \bibinfo{person}{Hongjiang
  Zhang}, {and} \bibinfo{person}{Anil~K Jain}.}
  \bibinfo{year}{2000}\natexlab{}.
\newblock \showarticletitle{Automatic caption localization in compressed
  video}.
\newblock \bibinfo{journal}{\emph{IEEE Trans. Pattern Anal. Mach. Intell}}
  \bibinfo{volume}{22}, \bibinfo{number}{4} (\bibinfo{year}{2000}),
  \bibinfo{pages}{385--392}.
\newblock


\bibitem[\protect\citeauthoryear{Zhou, Liu, Zhang, Wang, and Lin}{Zhou
  et~al\mbox{.}}{2014}]%
        {zhou2014perspective}
\bibfield{author}{\bibinfo{person}{Yu Zhou}, \bibinfo{person}{Shuang Liu},
  \bibinfo{person}{Yongzheng Zhang}, \bibinfo{person}{Yipeng Wang}, {and}
  \bibinfo{person}{Weiyao Lin}.} \bibinfo{year}{2014}\natexlab{}.
\newblock \showarticletitle{Perspective scene text recognition with feature
  compression and ranking}. In \bibinfo{booktitle}{\emph{Proceedings of ACCV}}.
  \bibinfo{pages}{181--195}.
\newblock


\bibitem[\protect\citeauthoryear{Zhu, Wang, Huang, and Chen}{Zhu
  et~al\mbox{.}}{2019}]%
        {zhu2019text}
\bibfield{author}{\bibinfo{person}{Yiwei Zhu}, \bibinfo{person}{Shilin Wang},
  \bibinfo{person}{Zheng Huang}, {and} \bibinfo{person}{Kai Chen}.}
  \bibinfo{year}{2019}\natexlab{}.
\newblock \showarticletitle{Text Recognition in Images Based on Transformer
  with Hierarchical Attention}. In \bibinfo{booktitle}{\emph{Proceedings of
  ICIP}}. \bibinfo{pages}{1945--1949}.
\newblock


\bibitem[\protect\citeauthoryear{Zhu, Yao, and Bai}{Zhu et~al\mbox{.}}{2016}]%
        {zhu2016scene}
\bibfield{author}{\bibinfo{person}{Yingying Zhu}, \bibinfo{person}{Cong Yao},
  {and} \bibinfo{person}{Xiang Bai}.} \bibinfo{year}{2016}\natexlab{}.
\newblock \showarticletitle{Scene text detection and recognition: Recent
  advances and future trends}.
\newblock \bibinfo{journal}{\emph{Frontiers of Computer Science}}
  \bibinfo{volume}{10}, \bibinfo{number}{1} (\bibinfo{year}{2016}),
  \bibinfo{pages}{19--36}.
\newblock


\end{thebibliography}

\appendix

\end{document}